%% file: main.tex
\documentclass[]{wan}

\usepackage{wrapfig}
\usepackage{tabularx}
\usepackage{textcomp}
\usepackage{stfloats}
\usepackage{url}
\usepackage{verbatim}
\usepackage{graphicx}
\usepackage{titlesec}
\usepackage{tocloft}
\usepackage{adjustbox}
\usepackage{multirow}
\usepackage{pifont}
\usepackage{tikz}
\usepackage{comment}
\usepackage{amsmath,amssymb} 
\usepackage{colortbl}  
\usepackage{color}
\usepackage{booktabs} 
\usepackage{hyperref}
\usepackage{graphicx}   
\usepackage{subcaption} 
\usepackage{booktabs}
\usepackage{amsmath} %
\usepackage{multirow} %
\usepackage{booktabs} %
\usepackage{subcaption} %
\usepackage{graphicx}   %
\usepackage{wrapfig}    %
\usepackage{caption}    %

\RequirePackage{xspace}
\makeatletter
\DeclareRobustCommand\onedot{\futurelet\@let@token\@onedot}
\def\@onedot{\ifx\@let@token.\else.\null\fi\xspace}

\def\eg{\emph{e.g}\onedot}

\makeatother

\usepackage{makecell}
\definecolor{Gray}{gray}{0.95}

\definecolor{adptorange}{RGB}{248, 205, 172}
\definecolor{cmpblue}{RGB}{189, 215, 238}
\definecolor{cmpblue}{RGB}{189, 215, 238}

\definecolor{our_red}{RGB}{232,157,160}
\definecolor{our_blue}{RGB}{136,206,230}
\definecolor{our_orange}{RGB}{246,200,168}
\definecolor{our_green}{RGB}{178,211,164}

\definecolor{attn_code0}{RGB}{247,215,200}
\definecolor{attn_code1}{RGB}{238,169,139}
\definecolor{mlp_code0}{RGB}{204,201,221}
\definecolor{mlp_code1}{RGB}{102,95,153}

\definecolor{token_blue}{RGB}{84, 120, 140}

\definecolor{myMagenta}{rgb}{0.9,0,0.4}

\usepackage{pifont}       
\usepackage{bbding}       
\usepackage{fontawesome}
\usepackage{xspace}

\usepackage{float}

\newlength\savewidth

\newcolumntype{x}[1]{>{\centering\arraybackslash}p{#1pt}}
\newcolumntype{y}[1]{>{\raggedright\arraybackslash}p{#1pt}}
\newcolumntype{z}[1]{>{\raggedleft\arraybackslash}p{#1pt}}

\usepackage{colortbl}
\usepackage{xcolor}
\usepackage{wrapfig}

\renewcommand{\paragraph}[1]{\vspace{1.25mm}\noindent\textbf{#1}}

\usepackage{algorithm}
\usepackage{algpseudocode}
\usepackage{listings}

\definecolor{codeblue}{rgb}{0.25, 0.5, 0.5}
\definecolor{codekw}{rgb}{0.35, 0.35, 0.75}
\lstdefinestyle{Pytorch}{
    language = Python,
    backgroundcolor = \color{white},
    basicstyle = \fontsize{9pt}{8pt}\selectfont\ttfamily\bfseries,
    columns = fullflexible,
    aboveskip=1pt,
    belowskip=1pt,
    breaklines = true,
    captionpos = b,
    commentstyle = \color{codeblue},
    keywordstyle = \color{codekw},
}

\definecolor{green}{HTML}{009000}
\definecolor{red}{HTML}{ea4335}

\newcommand{\name}{Wan-Animate-2\xspace}
\title{Wan-Animate-2: Pushing the Application Boundaries of Character Animation}

\author[*]{Guangyuan Wang}
\author[*,\dagger]{Li Hu}
\author[*]{Dechao Meng}
\author[*]{Zhongyi Zhang}
\author[*]{Peng Zhang}
\author[*]{Xindi Zhang}
\author{Mingyang Huang}
\author{Ruoshi Zhang}
\author{Ke Sun}
\author{Zhe Zhang}
\author{Xingjun Wang}
\author{Gang Cheng}
\author{Hai Xu}
\author[\ddagger]{Bang Zhang}
\affiliation{Tongyi Lab, Alibaba Group}
\contribution[*]{Core Contribution}
\contribution[\dagger]{Project Lead}
\contribution[\ddagger]{Sponsor}

\input{sec/0_abstract}

\begin{document}
\thispagestyle{firstheader}
\maketitle
\pagestyle{empty}

\input{sec/1_intro}
\input{sec/2_data}
\input{sec/3_Wan-Animate-2-Base}
\input{sec/4_Wan-Animate-2-Lite}

\input{sec/5_eval}
\input{sec/6_conclusion}

\clearpage

\bibliographystyle{assets/plainnat}
\bibliography{main}


\end{document}

%% file: sec/0_abstract.tex
 \abstract{

Character image animation remains a foundational yet challenging task in computer vision. Existing approaches can be broadly categorized into three paradigms: methods based on explicit motion representations suffer from extraction errors and identity drift; methods based on implicit motion features lose fine-grained dynamics through compression; and in-context learning approaches avoid intermediate representations but incur prohibitive computational costs. Furthermore, all current systems are designed for offline synthesis, unable to meet the real-time requirements of interactive applications such as digital avatars and live-streaming hosts. To address these limitations, we present \name, an end-to-end character animation framework that directly consumes the driving video within a redesigned Diffusion Transformer. Our architecture achieves superior motion fidelity and identity preservation by eliminating intermediate motion extractors entirely. We further introduce text-driven viewpoint control that decouples the output camera perspective from the driving video—a capability rarely supported by prior character animation methods that rely on explicit motion representations. Beyond generation quality, we present \name-Lite, an efficient variant that reduces inference latency to real-time thresholds through a three-stage training paradigm: teacher forcing pretraining with error buffer mechanism, and Self-Forcing distillation with chunk-wise backpropagation. This enables streaming character animation for interactive applications, opening new deployment scenarios that were previously infeasible. Qualitative evaluations and user studies demonstrate that \name achieves high-fidelity animation results across diverse characters and motion patterns. To foster further research and community development, we will release the \name-Base model weights to the public. Project page: \href{https://humanaigc.github.io/wan-animate-2/}{\textcolor{blue}{https://humanaigc.github.io/wan-animate-2/}}

}

%% file: sec/1_intro.tex
\section{Introduction}
Character image animation aims to transfer the spatiotemporal motion features from a driving video to a reference image containing a specific subject to generate an animated video. This technology shows broad application potential in film production, digital avatar creation, and animation production. Its core objective is to accurately model the motion dynamics in the driving sequence and seamlessly transfer them to a new character, thereby achieving precise motion control while faithfully preserving the visual appearance of the image. Meanwhile, leading closed-source video generation platforms now increasingly incorporate character animation as a built-in capability, while the open-source community has yet to produce systems of comparable quality, widening the gap between proprietary and publicly available solutions. Recognizing character animation as a fundamental and widely demanded video generation task, we aim to advance this direction in the open-source domain.

With the recent adoption of large-scale Diffusion Transformers (DiTs)~\cite{peebles2023scalable} as generative backbones, effectively capturing the underlying motion dynamics of a reference video and injecting them into the generative process remains a significant challenge; current character animation frameworks can be broadly categorized into three distinct paradigms: (a) methods based on explicit motion representations~\cite{chang2023magicdance,hu2023animateanyone,mimicmotion2024,ma2023follow,xu2023magicanimate,wang2023disco,zhu2024champ,karras2023dreamposefashionimagetovideosynthesis,yoon2025tpctesttimeprocrustescalibration,kim2024tcananimatinghumanimages,wang2024unianimate,tan2024animatexuniversalcharacterimage}, (b) methods based on implicit motion features~\cite{wang2024lia,Siarohin_2019_NeurIPS,song2025x,ding2025mtvcrafter,wang2022latent}, and (c) methods based on in-context learning~\cite{luo2026dreamactor,yan2026scail2}. Each paradigm addresses motion injection differently, yet each introduces distinct limitations in representation fidelity, generalizability, or computational cost.

The first two paradigms share a common reliance on intermediate motion representations extracted from the driving video. Explicit-motion methods~\cite{zhu2024champ,hu2023animateanyone,mimicmotion2024} employ auxiliary networks to derive 2D skeletons~\cite{yang2023effective,cao2017realtime,xu2022vitpose,xu2023vitpose++} or 3D SMPL parameters~\cite{loper2023smpl,pavlakos2019expressive}, which are then fused with the denoising features via convolutional encoders. While spatial alignment between the motion cues and the reference video provides a natural structural prior, this pipeline is vulnerable to extraction errors and prone to identity drift during cross-identity transfer, particularly when significant appearance or body-shape discrepancies exist between the source and the driver. Implicit-motion methods~\cite{wang2022latent,wang2024lia,song2025x} circumvent explicit cues by compressing the driving video into a learned latent space through a dedicated encoder, improving generalization at the cost of information loss: the compression bottleneck discards fine-grained dynamics essential for subtle expressions, intricate hand movements, and complex non-rigid motions.

In contrast, in-context learning (ICL) approaches~\cite{luo2026dreamactor,yan2026scail2} bypass intermediate representations entirely, conditioning generation on the raw driving video through self-attention between tokenized reference sequences and denoising latents. By operating directly on the pixel-level motion signal, ICL methods avoid both the extraction errors of explicit approaches and the information loss of implicit ones, achieving strong representational capacity. However, this comes at a steep computational price: full-sequence self-attention over all reference and target tokens incurs quadratic complexity, severely limiting scalability and rendering inference impractical for long sequences or high-resolution outputs.

Our key insight is that the reference video itself, when processed natively within the DiT, already constitutes a robust and information-complete motion prior—no intermediate representation is needed. Building on this, we propose \textbf{\name-Base}, an end-to-end framework that directly consumes the patchified latents of the reference video within the DiT architecture, complemented by a series of architectural refinements. (1) \textit{Dual-Branch Design}. Rather than concatenating all tokens into a single sequence for full self-attention, we employ a dual-branch DiT in which the reference branch operates independently with dedicated timestep and prompt inputs, propagating its key and value features to the latent branch. This preserves structural independence while ensuring efficient condition injection and reducing computational complexity. (2) \textit{Time-Align RoPE}. To resolve the positional encoding ambiguity between heterogeneous branches, we synchronize positional embeddings by performing frame-wise token concatenation before applying rotary position encoding, ensuring coherent spatio-temporal alignment across branches regardless of resolution discrepancies. (3) \textit{Sparse-Ref Attention}. Exploiting the inherent temporal correspondence between reference and target sequences, we restrict each latent token to attend only to its temporally aligned reference counterpart, substantially reducing attention costs while preserving high-fidelity motion guidance. Beyond motion transfer, we further address an orthogonal limitation in prior paradigms based on explicit or implicit motion representations: the rigid coupling between the output camera viewpoint and the driving video. To decouple these, we introduce an optional \textit{Viewpoint LoRA} that maps discretized azimuth and elevation angles into a text-controlled space via low-rank adaptation of the cross-attention layers, enabling flexible camera manipulation through simple text prompts without requiring explicit camera parameters.

While \name substantially advances generation quality through its architectural design, the aforementioned methods—spanning explicit, implicit, and ICL-based paradigms—share a fundamental limitation: they are all designed for offline video synthesis. Multi-step diffusion sampling, while effective for quality, requires tens of denoising iterations, resulting in per-frame latencies far exceeding real-time thresholds. This poses a critical barrier for the most impactful applications of character animation, such as interactive digital avatars, live-streaming hosts, and real-time virtual environments, where low-latency, streaming video generation is a prerequisite rather than a convenience. The gap between offline generation quality and online deployment feasibility thus constitutes an orthogonal yet equally pressing challenge.

To bridge this gap, we further present \textbf{\name-Lite}, a lightweight variant that achieves significant inference acceleration through a principled three-stage training paradigm: (1) \textit{Teacher Forcing Pretraining}, which reformulates the diffusion model into a causal generator, enabling chunk-wise autoregressive synthesis; (2) \textit{Error Buffer Training}, which injects realistic prediction residuals into the training context to mitigate the exposure bias between teacher-forced training and autoregressive inference; and (3) \textit{Self-Forcing Distillation}~\cite{huang2026self}, which compresses the multi-step denoising process into fewer iterations. To make this tractable at the 14B-parameter scale, we design a chunk-wise backpropagation strategy that decouples the forward rollout from gradient computation, reducing peak memory from sequence-proportional to chunk-proportional while preserving the theoretical guarantees of Distribution Matching Distillation~\cite{yin2024one}.

Together, \name pushes the application boundaries of character animation in three complementary directions: higher-fidelity generation through a redesigned end-to-end architecture, flexible viewpoint control as a new capability, and a real-time streaming variant that unlocks interactive deployment scenarios. To facilitate further research and practical applications, we will release the \name-Base model to the public. Our key contributions are summarized as follows:

\begin{itemize}
    \item We propose an end-to-end character animation framework that eliminates reliance on auxiliary motion extractors by directly consuming the driving video within a redesigned Diffusion Transformer, achieving superior motion fidelity and identity preservation across diverse characters and motion patterns.
    \item We introduce text-driven viewpoint control that decouples the output camera perspective from the driving video, enabling flexible camera manipulation via simple text prompts—a capability rarely supported by prior character animation methods that rely on explicit motion representations.
    \item We present \name-Lite, an efficient variant that reduces inference latency to real-time thresholds through a principled three-stage training paradigm, enabling streaming character animation for interactive applications such as digital avatars and live-streaming hosts.
\end{itemize}

%% file: sec/2_data.tex
\section{Data}

The quality and diversity of training data are fundamental to building a robust character animation framework. We construct two complementary datasets tailored to distinct training objectives: (i) a large-scale collection of paired video data synthesized by Wan-Animate models, which serves as the primary training corpus for both the \name-Base architecture and the \name-Lite acceleration variant; and (ii) a high-fidelity synthetic multi-view dataset rendered via Unreal Engine, which provides ground-truth viewpoint variations used exclusively to train our text-driven Viewpoint LoRA. This separation allows us to optimize each component independently while maintaining data quality through rigorous filtering protocols. We detail the construction and curation of each dataset below.

\subsection{Paired Video Data}

Training the \name-Base model requires strictly aligned video pairs in which different characters exhibit identical motion dynamics and facial expressions. Acquiring such paired data from real-world sources is inherently challenging, as natural videos rarely contain the controlled variations needed for supervised learning. To address this bottleneck, we establish an automated synthesis pipeline built upon Wan-Animate that generates high-quality paired video data at scale.

\paragraph{Data Synthesis Pipeline.} Our pipeline begins with a curated collection of reference videos covering diverse character types and motion patterns. For each training sample, we generate a reference image using one of two complementary strategies: (1) extracting the initial frame from collected reference videos and editing both background and character elements using Qwen-Image-Edit~\cite{wu2025qwen} to produce high-quality reference images; or (2) generating diverse character assets—including humans and anthropomorphic cartoon animals—using Qwen-Image~\cite{wu2025qwen} and Z-Image~\cite{cai2025z}, spanning various shot scales (full-body, half-body, and close-ups) and aspect ratios. We then randomly pair selected reference videos with generated reference images and synthesize new video clips using Wan-Animate, creating the aligned video pairs needed for training.

\paragraph{Quality Filtering.} To ensure training data quality, we implement a rigorous multi-dimensional filtering pipeline that evaluates each synthesized sample across three key dimensions: (1) \textit{Overall video quality}, which removes samples with artifacts, blur, or low visual fidelity; (2) \textit{Motion characteristics}, including amplitude and smoothness metrics that filter out static or jittery motions; and (3) \textit{Subject consistency}, which measures the degree to which the target character maintains identity throughout the generated sequence. Only samples meeting stringent thresholds across all criteria are retained for training.

\subsection{Synthetic Multi-View Data}

While paired video data enables learning of motion transfer capabilities, training the Viewpoint LoRA requires explicit supervision of camera viewpoint variations—something that is difficult to obtain from real-world videos. We utilize Unreal Engine~\cite{unrealengine} to construct a synthetic multi-view dataset with precise camera control.

\paragraph{Rendering Protocol.} We discretize the camera viewpoint space into 12 azimuthal orientations and 4 elevation angles, yielding 48 distinct viewpoints. For each action sequence and scene configuration, we randomly sample combinations from this discrete viewpoint space to synthesize multi-view renderings. This provides ground-truth supervision for learning viewpoint-controlled generation across the full range of camera positions.

\paragraph{Text-Based Annotation.} Rather than encoding camera parameters as numerical values (e.g., rotation matrices or Euler angles), we adopt a simplified text-based annotation scheme. Each rendered frame is labeled with descriptive spatial regions such as ``right 60-degree view'' or ``top angle''. This design choice aligns with our Viewpoint LoRA's text-driven interface and offers two key advantages: (1) it eliminates the burden on users to provide standardized camera parameters, and (2) it enables intuitive viewpoint control through natural language prompts during inference.

%% file: sec/3_Wan-Animate-2-Base.tex
\section{Wan-Animate-2-Base}

\subsection{Overview}

\begin{figure*}[t]
    \centering
    \includegraphics[width=\linewidth]{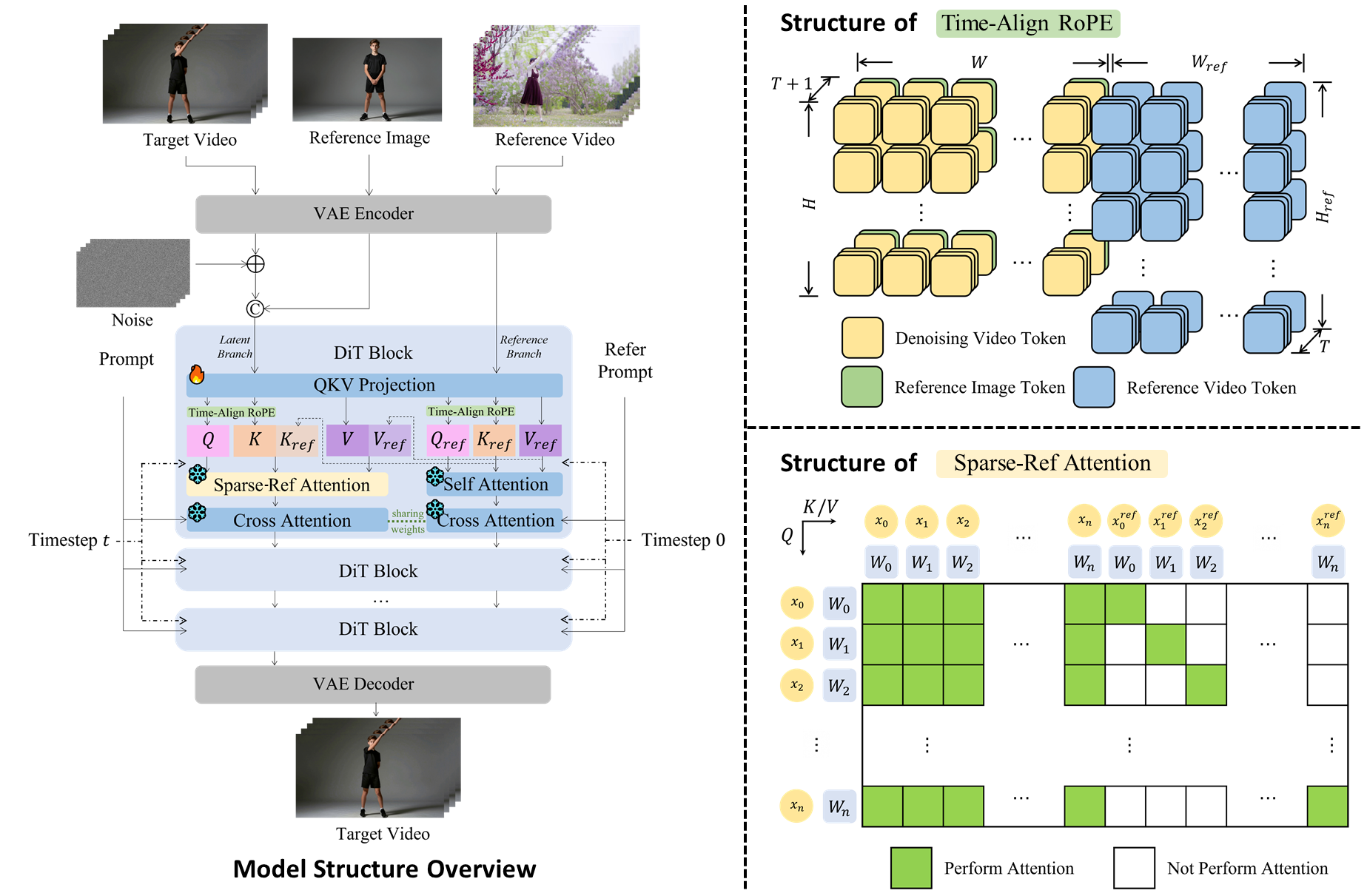}
    \caption{Overview of our framework. Our method conditions video generation on a reference image and a reference video through a dual-branch DiT architecture. To effectively fuse static appearance and dynamic motion cues, we introduce Time-Align RoPE for temporal alignment between denoising video tokens and reference tokens, and Sparse-Ref Attention to selectively attend to informative reference features.}
    \label{fig:pipeline-base}
\end{figure*}

A fundamental challenge in character image animation lies in effectively integrating motion dynamics from a reference video into the generative process without relying on error-prone intermediate representations or incurring prohibitive computational costs. \name addresses this by reformulating animation as a direct video-conditioned generation task within a Diffusion Transformer (DiT). Our architecture consists of four key components: a \textit{Dual-Branch DiT} that decouples reference and latent streams for efficient condition injection; a \textit{Time-Align RoPE} that ensures precise spatio-temporal alignment across branches; a \textit{Sparse-Ref Attention} that exploits temporal correspondence to reduce computational overhead; and an optional \textit{Viewpoint LoRA} that enables text-driven camera control. The overall pipeline is illustrated in Figure~\ref{fig:pipeline-base}.

\subsection{Dual-Branch DiT Architecture}

In-context learning (ICL) leverages the capacity of self-attention mechanisms to propagate structural priors from input sequences to target latent representations. Building on this, we propose a memory-efficient conditioning strategy that integrates motion priors into the latent space while avoiding the prohibitive computational overhead of full-sequence self-attention. Specifically, we employ a shared query-key-value (QKV) projection layer to derive projections for both the noisy latent and reference latent streams, supplemented by a time-align Rotary Positional Embedding (RoPE). As shown in Figure~\ref{fig:pipeline-base}, we adopt a decoupled attention strategy: while both branches share the same projection parameters, the latent branch operates at the diffusion timestep $t$, whereas the reference branch is anchored to $t=0$ to ensure a clean, noise-free motion prior. We concatenate the key and value tensors from the reference branch with those of the latent branch but execute the Sparse-Ref attention and self-attention independently across these two streams. Subsequently, these streams are processed through separate cross-attention blocks conditioned on distinct text prompts.

This design achieves two primary objectives: first, it mitigates the quadratic complexity of global self-attention while ensuring the conditional branch remains shielded from the noise-induced interference of the denoising process, thereby preserving high-fidelity guidance; second, by fixing the reference branch at $t=0$, we effectively anchor the generative process to a clean latent manifold, reinforcing robust temporal alignment across all denoising stages. To facilitate stable learning, we restrict the training process for these attention modules to only the shared QKV projection layers, keeping the core transformer weights frozen.

\subsection{Time-Align RoPE}

Having established the dual-branch architecture, we next address the positional encoding challenge introduced by the reference branch. Rotary Positional Embeddings (RoPE) are critical for encoding relative spatiotemporal dependencies within attention mechanisms. In our architecture, the introduction of a reference branch alongside the denoising latent branch necessitates a dedicated positional strategy. Simply applying identical RoPE configurations to both branches introduces ambiguity that can destabilize training. To resolve this, we introduce Time-Align RoPE, a specialized encoding scheme designed to explicitly synchronize the latent and reference branches.

As illustrated in Figure~\ref{fig:pipeline-base}, we implement this by prepending reference video tokens to the denoised video sequence, forming a unified temporal manifold. Given that both sequences share an identical temporal duration, we perform frame-wise token concatenation prior to applying the RoPE transformation. To accommodate resolution discrepancies between the reference and target outputs, the spatial offset of the RoPE is dynamically computed: specifically, if the reference and target have spatial dimensions $(H_r, W_r)$ and $(H_t, W_t)$, the reference tokens are assigned a spatial offset of $H_t \times W_t$ so that their positional indices do not overlap with those of the target tokens. This construction ensures consistent positional indexing across variable resolutions, allowing the DiT to maintain precise spatiotemporal correspondence between the reference motion and the evolving target frames.

\subsection{Sparse Reference Attention}

While the dual-branch architecture reduces the overall attention scope, full cross-branch interaction between all reference and latent tokens still incurs substantial overhead. To further optimize computational throughput, we introduce the Sparse-Ref Attention mechanism. In standard in-context learning, full-sequence attention across concatenated reference and latent tokens has complexity $\mathcal{O}((N_r + N_l)^2)$, where $N_r$ and $N_l$ denote the number of reference and latent tokens respectively. To mitigate this, we employ a temporally-constrained attention mask during the interaction between the latent and reference branches. Specifically, we restrict each query token in the latent branch to attend only to its temporally corresponding key and value tokens from the reference branch, while maintaining full self-attention within the latent stream. This design is predicated on the inherent frame-wise alignment between the reference and target sequences in character animation. By pruning non-essential cross-sequence interactions, our Sparse-Ref attention reduces the cross-branch attention complexity from $\mathcal{O}(N_r \times N_l)$ to $\mathcal{O}(N_l)$, significantly lowering computational and memory overhead while preserving high-fidelity motion guidance through targeted, frame-aligned feature interaction.

\subsection{Viewpoint LoRA}

Beyond motion transfer, existing character animation frameworks suffer from a rigid coupling between camera viewpoints and motion dynamics, effectively tethering the output viewpoint to the driving video. To decouple these elements, we formulate viewpoint adjustment as a text-conditioned task and integrate a lightweight Low-Rank Adaptation (LoRA)~\cite{hu2022lora} into the cross-attention layers of the DiT. We define a discrete viewpoint space comprising 12 azimuthal orientations and 4 elevation angles, and augment the training prompts with textual descriptions of these camera states (e.g., ``right 60-degree view'', ``top angle''). The LoRA modules are applied exclusively to the cross-attention projection matrices, leaving the base model weights unchanged. This text-based formulation maps viewpoint control into the semantic space of the diffusion model, enabling flexible camera manipulation through natural language prompts without requiring users to specify explicit camera parameters.

%% file: sec/4_Wan-Animate-2-Lite.tex
\section{Wan-Animate-2-Lite}

\subsection{Overview}

\begin{figure*}[t]
    \centering
    \includegraphics[width=\linewidth]{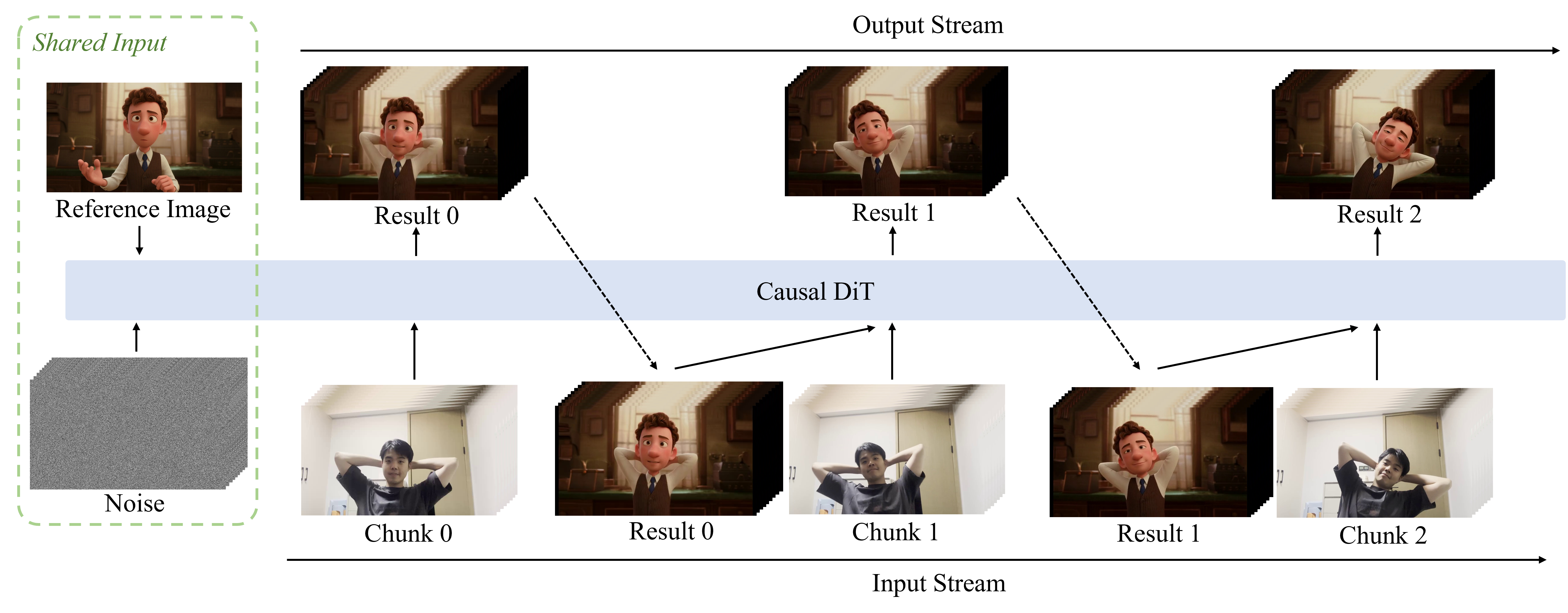}
    \caption{Autoregressive inference pipeline of \name-Lite. Leveraging a Causal Diffusion Transformer (Causal DiT), the framework processes character animations in a chunk-wise manner. During inference, the input driving stream is partitioned into temporal segments of 8 frames. The generated latents of the preceding chunk, along with the static reference image and noise, are fed into the Causal DiT to autoregressively synthesize each successive chunk, ensuring seamless temporal continuity across the output stream.}
    \label{fig:lite_inference}
\end{figure*}

While \name-Base achieves high-fidelity character animation through its multi-step diffusion process, the iterative denoising procedure inherently demands substantial computational resources, limiting its applicability in latency-sensitive scenarios. To address this, we introduce \name-Lite, a lightweight variant designed to significantly reduce inference cost while preserving generation quality. The corresponding inference pipeline, which leverages a causal formulation to synthesize video sequences in an autoregressive, chunk-by-chunk manner, is illustrated in Figure~\ref{fig:lite_inference}. Our approach is grounded in a three-stage training paradigm: (1) \textit{Teacher Forcing Pretraining}, which reformulates the diffusion model into a causal generation framework by conditioning on previously generated latents; (2) \textit{Error Buffer Training}, which injects realistic prediction residuals into the training context to mitigate the exposure bias between teacher-forced training and autoregressive inference; and (3) \textit{Self-Forcing Distillation}, which distills multi-step denoising into fewer steps through a novel chunk-wise backpropagation strategy tailored for large-scale models.

\subsection{Teacher Forcing Pretraining}

The first stage of \name-Lite reformulates the standard diffusion model into a causal generation framework that enables sequential, chunk-wise video synthesis. In conventional diffusion-based video generation, the model denoises the entire latent sequence simultaneously, requiring all frames to be processed in a single forward pass. This global denoising paradigm precludes streaming generation and imposes prohibitive memory requirements for long sequences.

To overcome this limitation, we adopt a teacher forcing strategy that transforms the model into an autoregressive generator operating over temporal chunks. Specifically, during training, we partition the video latent sequence into non-overlapping temporal chunks. For each chunk, we concatenate the corresponding clean latents from preceding chunks with the noisy latents of the current chunk along the temporal dimension. A causal attention mask is then applied to enforce an autoregressive dependency structure: noisy latents within the current chunk are permitted to attend to all preceding clean latents and to each other, while clean context tokens are masked from attending to future noisy tokens. This design effectively simulates the inference regime, where the model sequentially generates video chunks conditioned on its own prior outputs.

By anchoring the conditioning context to ground-truth clean latents during this stage, teacher forcing provides a stable training signal that enables the model to learn the causal generation pattern.

\paragraph{Error Buffer Mechanism.} A critical discrepancy persists between training and inference: during training, the model conditions on ground-truth clean latents, whereas at inference time, it must condition on its own imperfect predictions. This exposure bias leads to error accumulation across chunks, progressively degrading generation quality over extended sequences. To bridge this train-inference gap, we incorporate an error buffer mechanism~\cite{longlive_2.0,li2025stable} directly into the teacher forcing pretraining process. Concretely, at each training step, we first execute a forward pass for the current chunk and compute the residual between the model's one-step prediction and the ground-truth latent. This residual is recorded as the \textit{error buffer}, representing the characteristic prediction error of the model. In subsequent training iterations, we corrupt the clean context latents by adding the recorded error buffer before feeding them as conditioning input. The error buffer is maintained as a running estimate that captures the distribution of prediction errors across the training set. By exposing the model to realistic, imperfect conditioning signals during teacher forcing, we enable it to develop robustness against cascading errors without requiring full autoregressive rollouts, thereby maintaining computational efficiency while substantially improving temporal consistency in long-sequence generation.

\subsection{Self-Forcing Distillation}

The final stage of \name-Lite distills the multi-step denoising process into fewer sampling steps to further accelerate inference. We adopt the Self-Forcing paradigm~\cite{huang2026self}, which leverages the model's own generation trajectory as a training signal for step reduction. However, directly applying Self-Forcing to large-scale models (\eg, 14B parameters) presents a fundamental memory challenge: the algorithm requires full autoregressive rollouts through the model to compute distribution matching scores, making naive gradient-based optimization infeasible.

To enable Self-Forcing training at the 14B scale, we propose a chunk-wise backpropagation strategy that decouples the forward rollout from the gradient computation. Our approach proceeds in two phases:

\paragraph{Phase 1: Autoregressive Rollout and Score Computation.} We first perform a complete autoregressive inference pass through the student model \textit{without} gradient tracking. We randomly sample a number of denoising steps $T$ and, for each temporal chunk, execute $T$ denoising steps with the student model. Upon completion, we record two quantities per chunk: the noisy input at the final denoising step and the student's predicted clean output after all $T$ steps. We then add noise to all recorded outputs and feed the entire noisy sequence into the pre-trained real and fake score models. Both score models execute full bidirectional attention over the entire sequence and produce per-chunk distributional scores following the Distribution Matching Distillation (DMD) framework~\cite{yin2024one,yin2024improved}, all in a single batch pass without gradient tracking.

\paragraph{Phase 2: Chunk-wise Gradient Accumulation.} With the pre-computed real and fake scores, we re-feed the recorded noisy inputs into the student model chunk by chunk, this time with gradient tracking enabled. For each chunk, we compute the DMD loss using the corresponding pre-computed scores, backpropagate, and accumulate the gradients. Crucially, we strictly sever the computational graph between consecutive chunks, ensuring that gradients do not flow across chunk boundaries. After all chunks have been processed, a single parameter update is applied using the accumulated gradients (Algorithm~\ref{alg:chunkwise-gradient}). This design reduces the peak memory footprint from being proportional to the full sequence length to being proportional to a single chunk, thereby enabling Self-Forcing training on the 14B-parameter model with practical hardware constraints. 

\begin{algorithm}[t]
\caption{Chunk-wise Gradient Update for Self-Forcing Distillation}
\label{alg:chunkwise-gradient}
\begin{algorithmic}[1]
\Require Student model $\theta_s$, real score model $\phi^{\text{real}}$, fake score model $\phi^{\text{fake}}$, number of chunks $N$
\Ensure Updated student model parameters $\theta_s$
\State \textbf{// Phase 1: Autoregressive Rollout (no gradient)}
\State Randomly sample number of denoising steps $T$
\For{each chunk $i = 1, \ldots, N$}
    \State Initialize $z_i^{(T)}$ from noise schedule
    \For{each denoising step $t = T, T\!-\!1, \ldots, 1$}
        \State $z_i^{(t-1)} \leftarrow \theta_s(z_i^{(t)})$ \Comment{Student forward, no gradient}
    \EndFor
    \State Record $\hat{z}_i \leftarrow z_i^{(1)}$ \Comment{Noisy input to the last step}
    \State Record $\hat{y}_i \leftarrow z_i^{(0)}$ \Comment{Final denoised output}
\EndFor
\State \textbf{// Score Computation (bidirectional, all chunks at once)}
\State Add noise to all $\{\hat{y}_i\}_{i=1}^{N}$ to obtain $\{\tilde{y}_i\}_{i=1}^{N}$
\State $\{s^{\text{real}}_i\}_{i=1}^{N} \leftarrow \phi^{\text{real}}(\{\tilde{y}_i\}_{i=1}^{N})$ \Comment{Full-sequence bidirectional pass, no gradient}
\State $\{s^{\text{fake}}_i\}_{i=1}^{N} \leftarrow \phi^{\text{fake}}(\{\tilde{y}_i\}_{i=1}^{N})$ \Comment{Full-sequence bidirectional pass, no gradient}
\State \textbf{// Phase 2: Chunk-wise Gradient Accumulation}
\State Initialize accumulated gradient $g \leftarrow 0$
\For{each chunk $i = 1, \ldots, N$}
    \State Feed $\hat{z}_i$ into student model $\theta_s$ \textit{with} gradient tracking
    \State Compute DMD loss: $\mathcal{L}_i \leftarrow \text{DMD}(\theta_s(\hat{z}_i),\; s^{\text{real}}_i,\; s^{\text{fake}}_i)$
    \State Backpropagate $\nabla_{\theta_s} \mathcal{L}_i$ and accumulate: $g \leftarrow g + \nabla_{\theta_s} \mathcal{L}_i$
    \State Detach computational graph \Comment{Sever gradient flow between chunks}
\EndFor
\State Update $\theta_s \leftarrow \theta_s - \eta \cdot g$ \Comment{Single parameter update after all chunks}
\end{algorithmic}
\end{algorithm}

This chunk-wise strategy preserves the theoretical guarantees of DMD-based distillation—since the scores are computed from complete, untruncated rollouts—while making the gradient computation tractable through temporal decomposition. The resulting distilled model achieves comparable generation quality to the full multi-step teacher while requiring significantly fewer denoising iterations at inference time.

%% file: sec/5_eval.tex
\section{Results}

\subsection{Qualitative Results}

\begin{figure*}
    \begin{center}
    \setlength{\tabcolsep}{0.5pt}
    \begin{tabular}{m{1.85cm}<{\centering}m{1.85cm}<{\centering}m{1.85cm}<{\centering}m{1.85cm}<{\centering}m{0.5cm}<{\centering}m{1.85cm}<{\centering}m{1.85cm}<{\centering}m{1.85cm}<{\centering}m{1.85cm}<{\centering}}

    & \multicolumn{3}{c}{\scriptsize Reference Video} & & & \multicolumn{3}{c}{\scriptsize Reference Video} \\
    & \includegraphics[width=1.8cm]{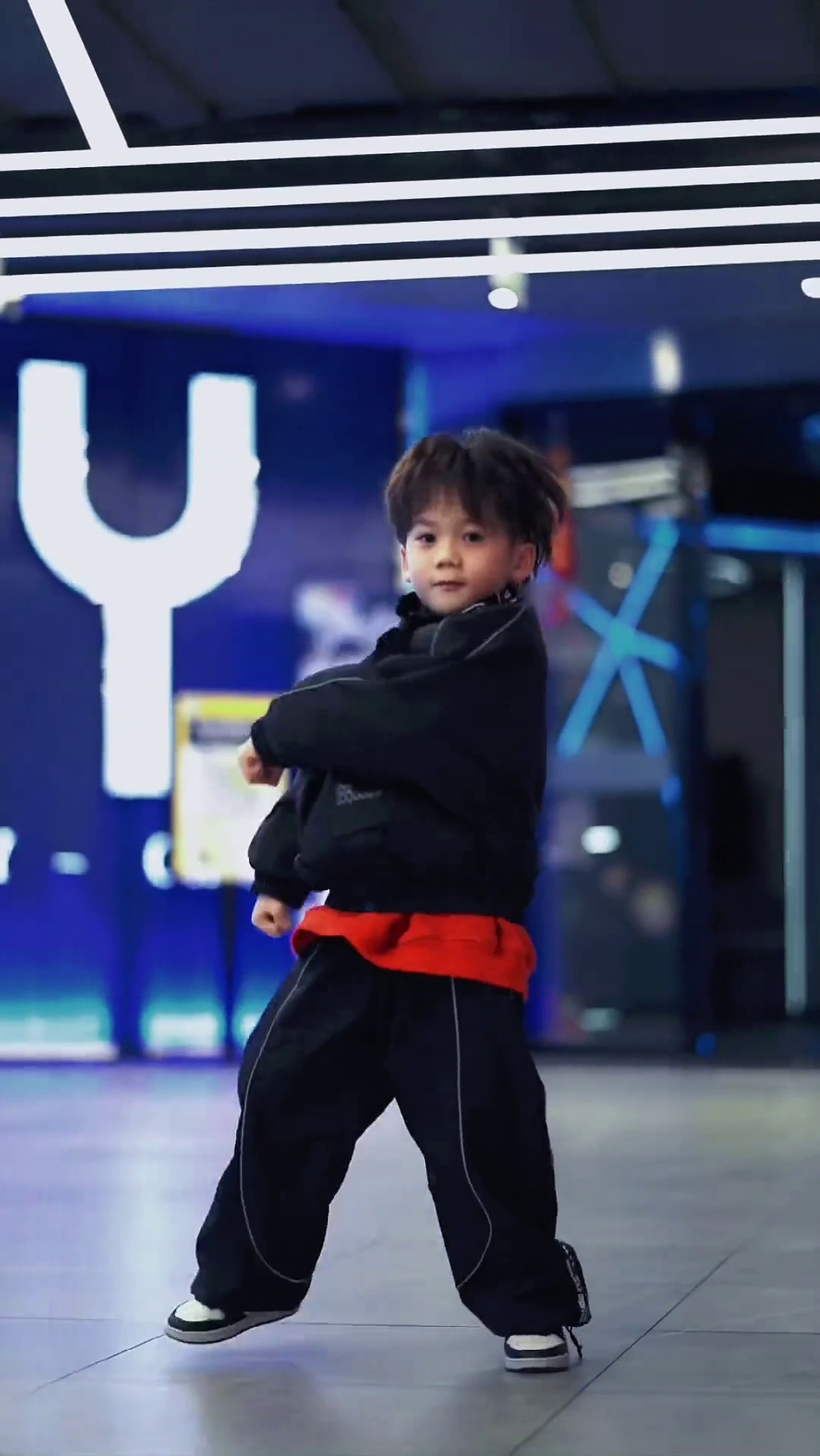} & \includegraphics[width=1.8cm]{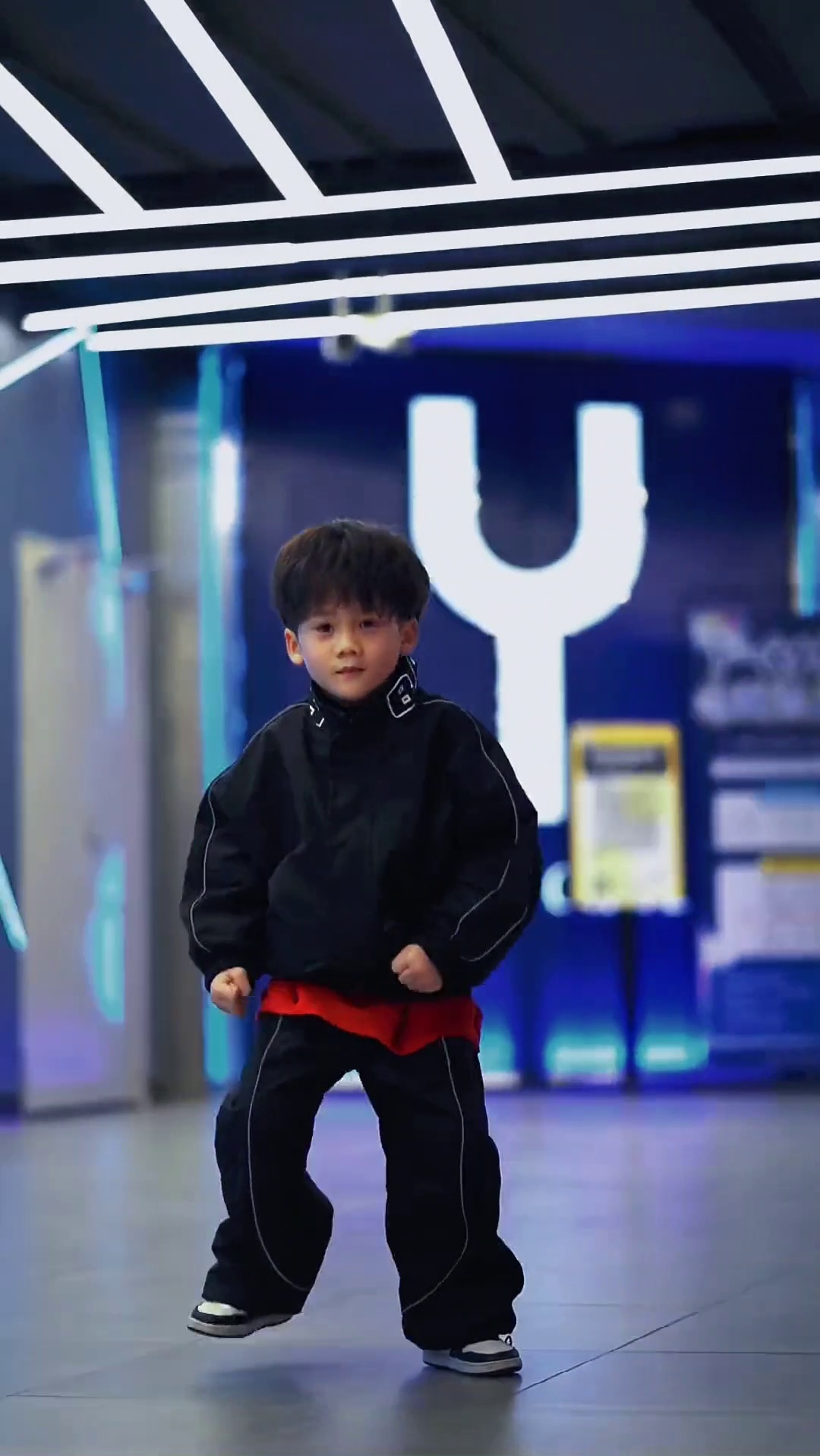} & \includegraphics[width=1.8cm]{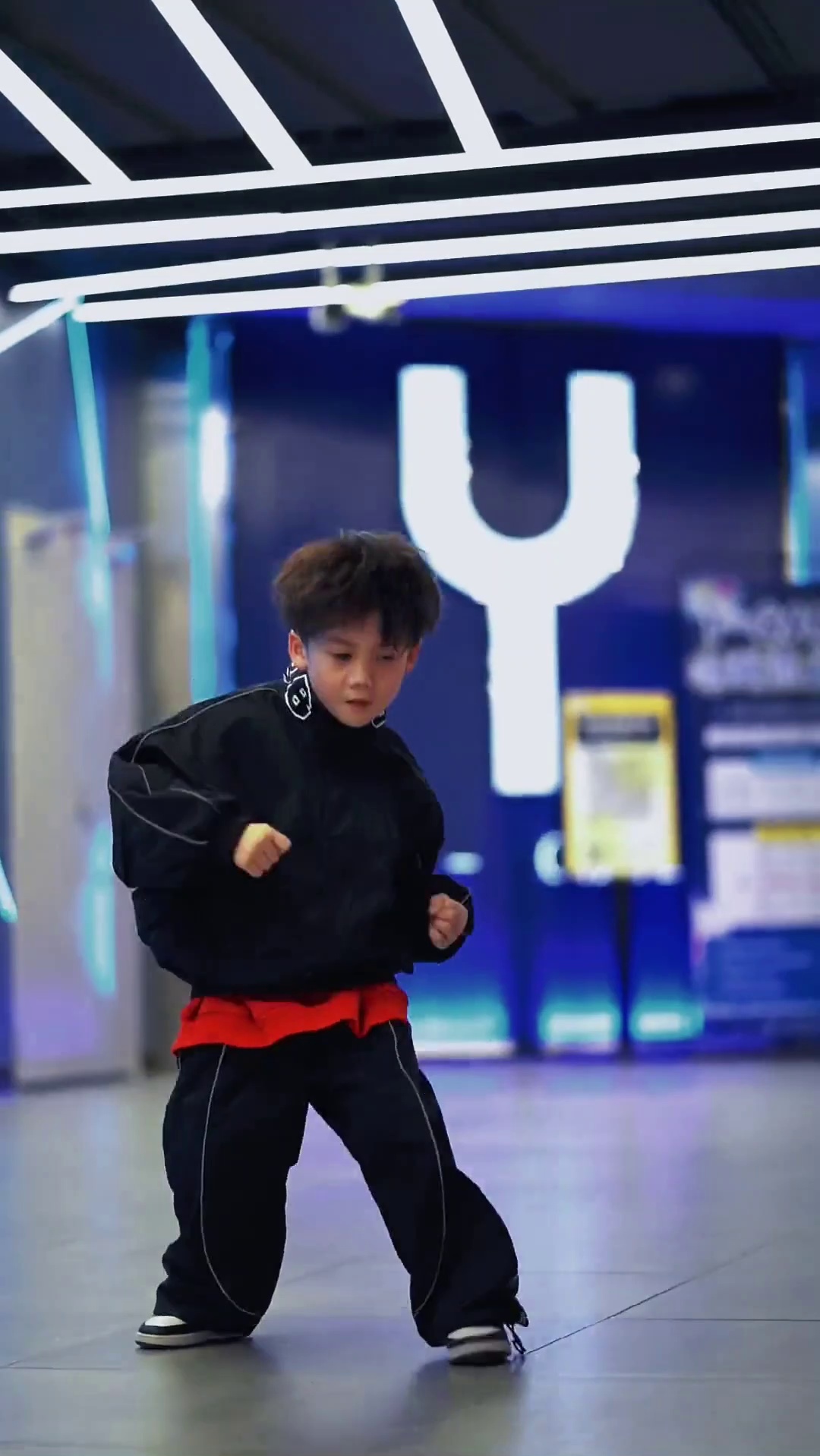} & & & \includegraphics[width=1.8cm]{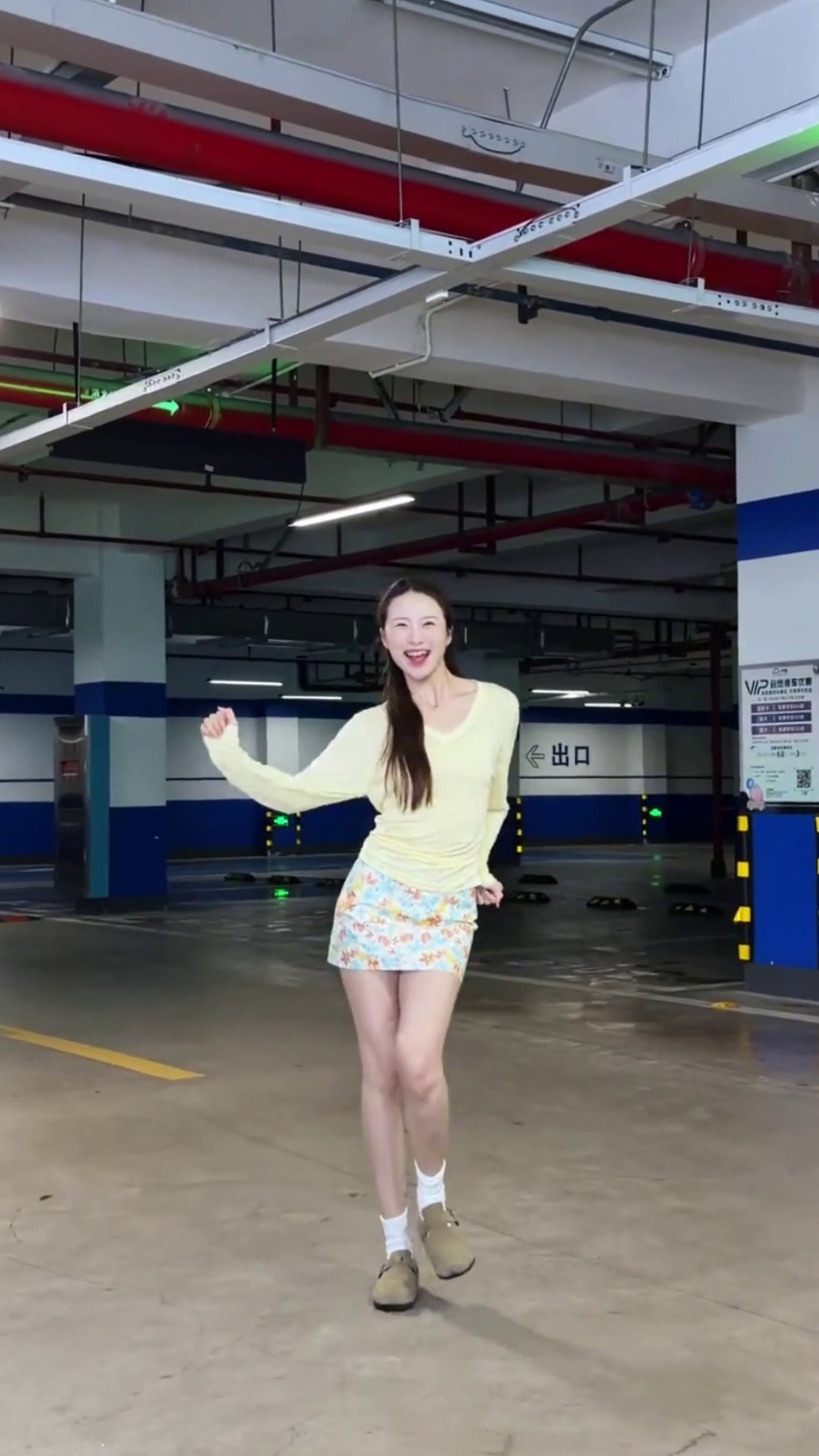} & \includegraphics[width=1.8cm]{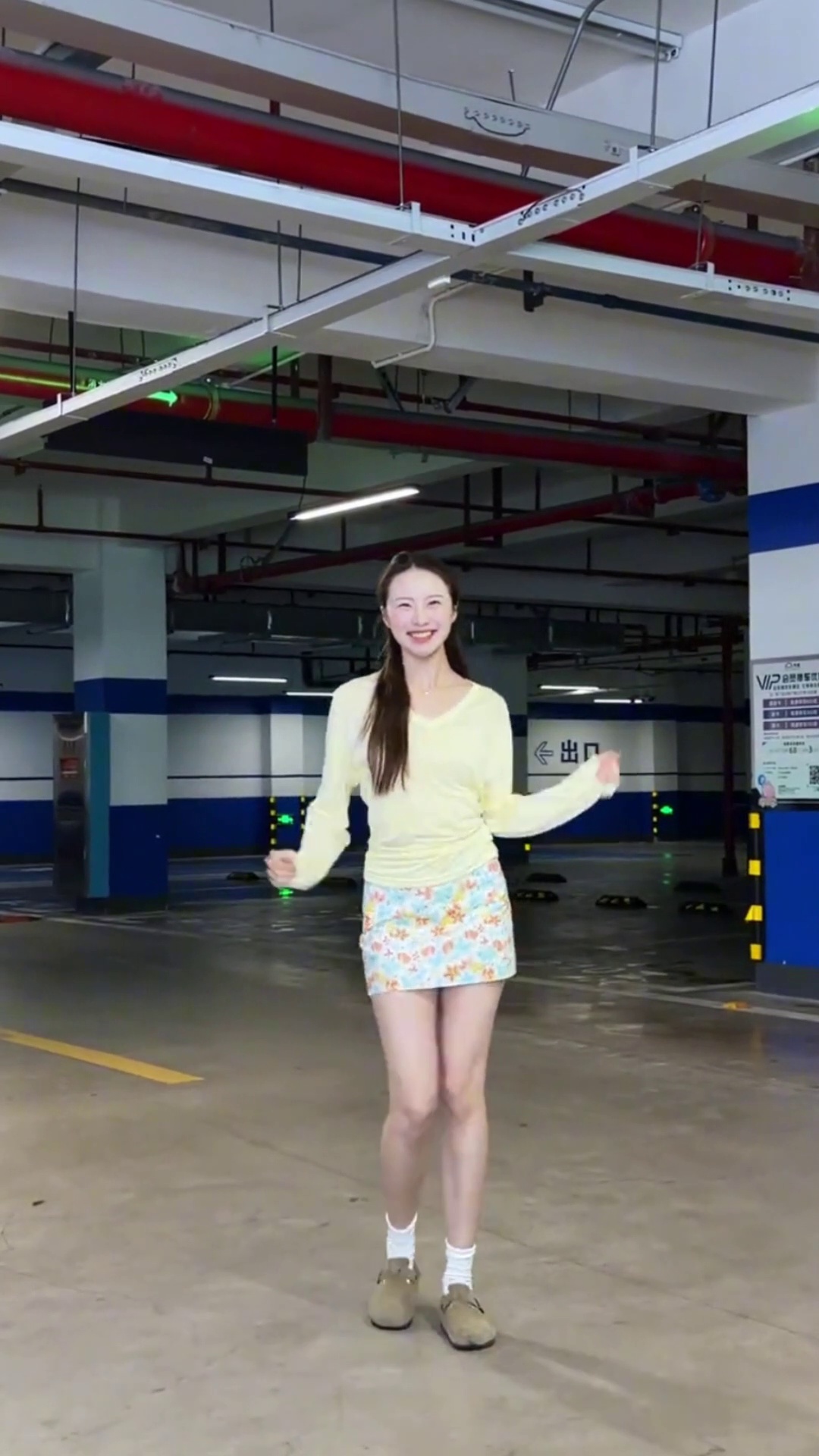} & \includegraphics[width=1.8cm]{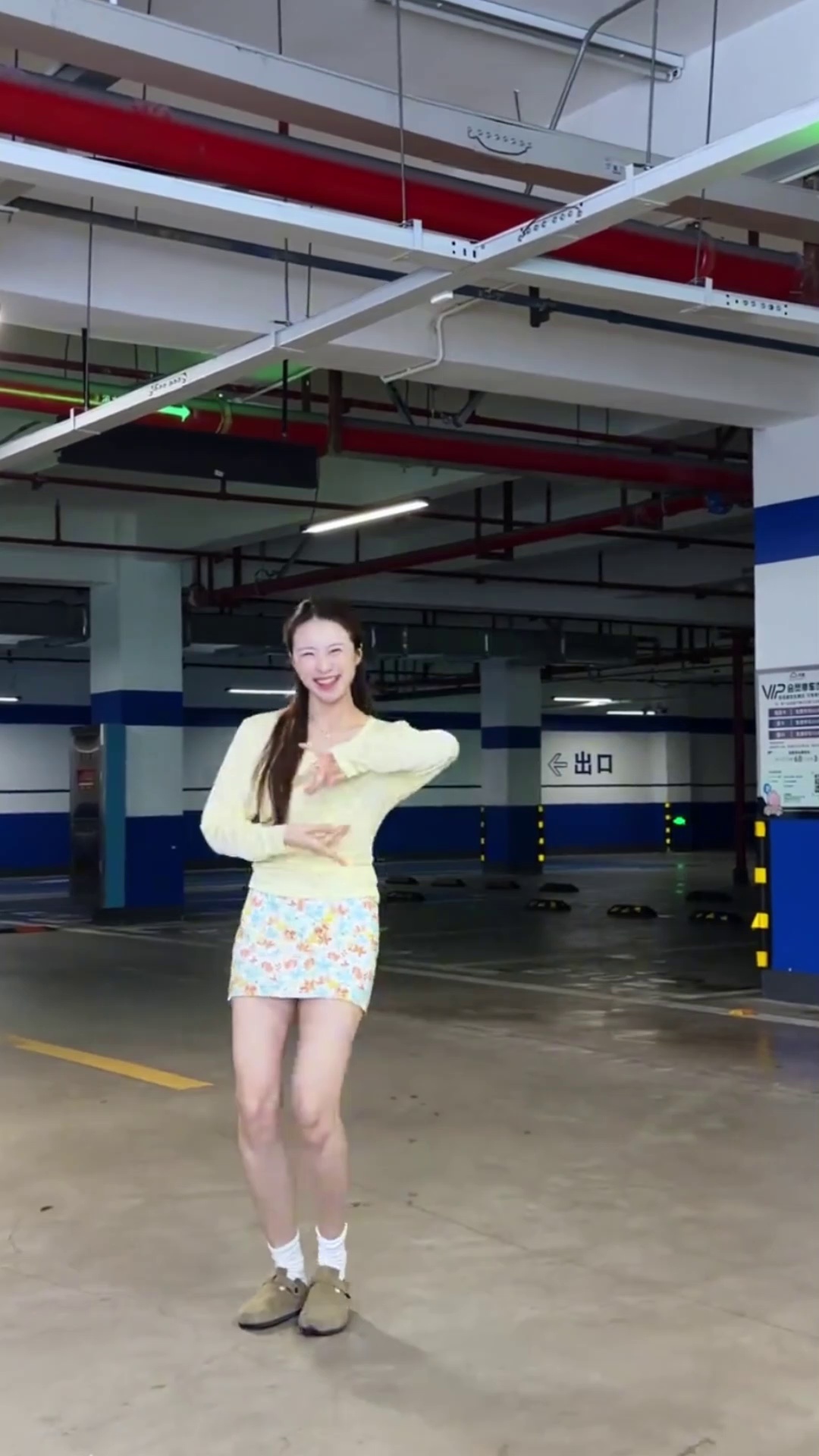} \\

    \scriptsize Reference Image & \multicolumn{3}{c}{\scriptsize Animation Result} & & \scriptsize Reference Image & \multicolumn{3}{c}{\scriptsize Animation Result} \\
    \includegraphics[width=1.8cm]{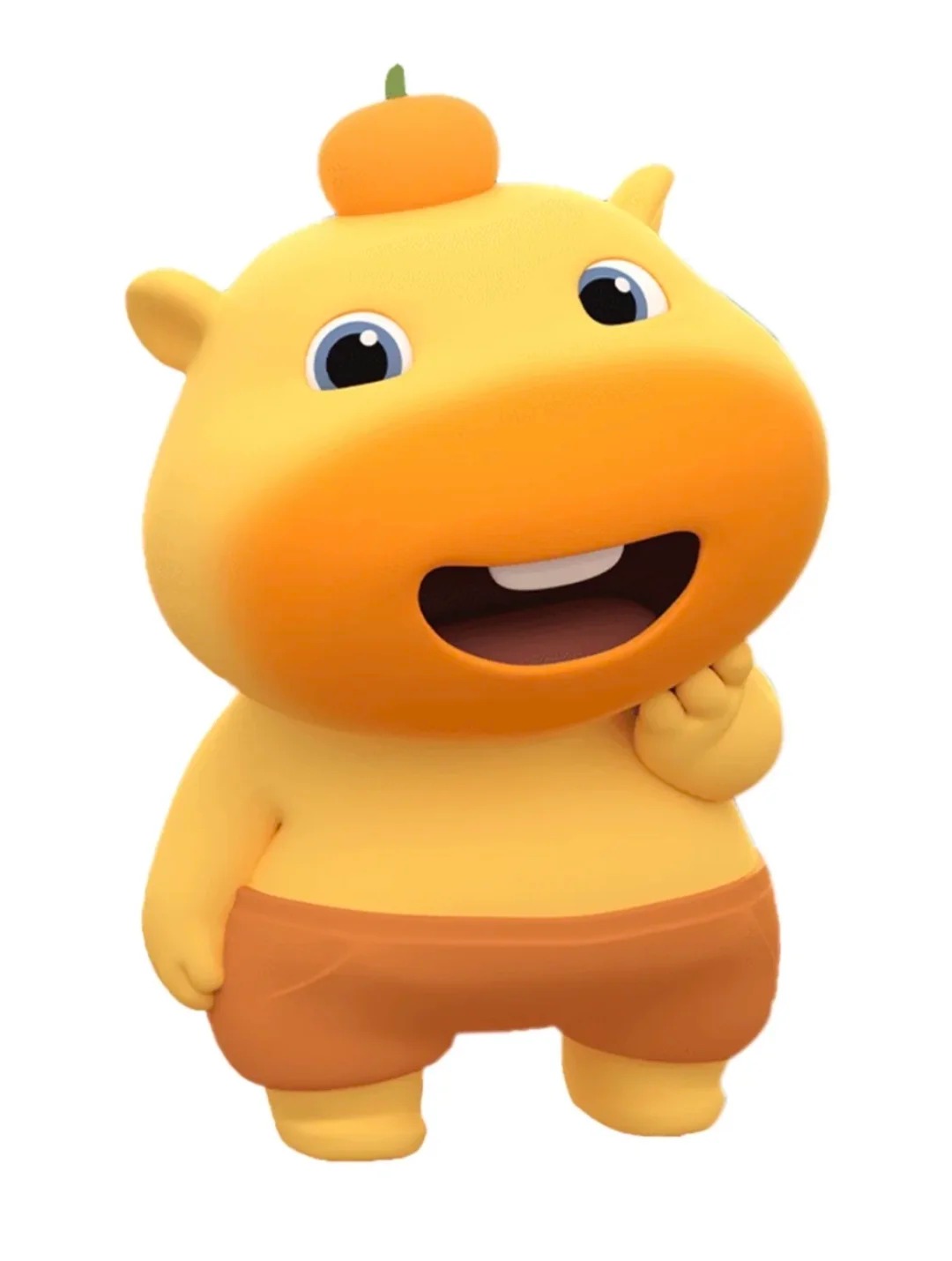} & \includegraphics[width=1.8cm]{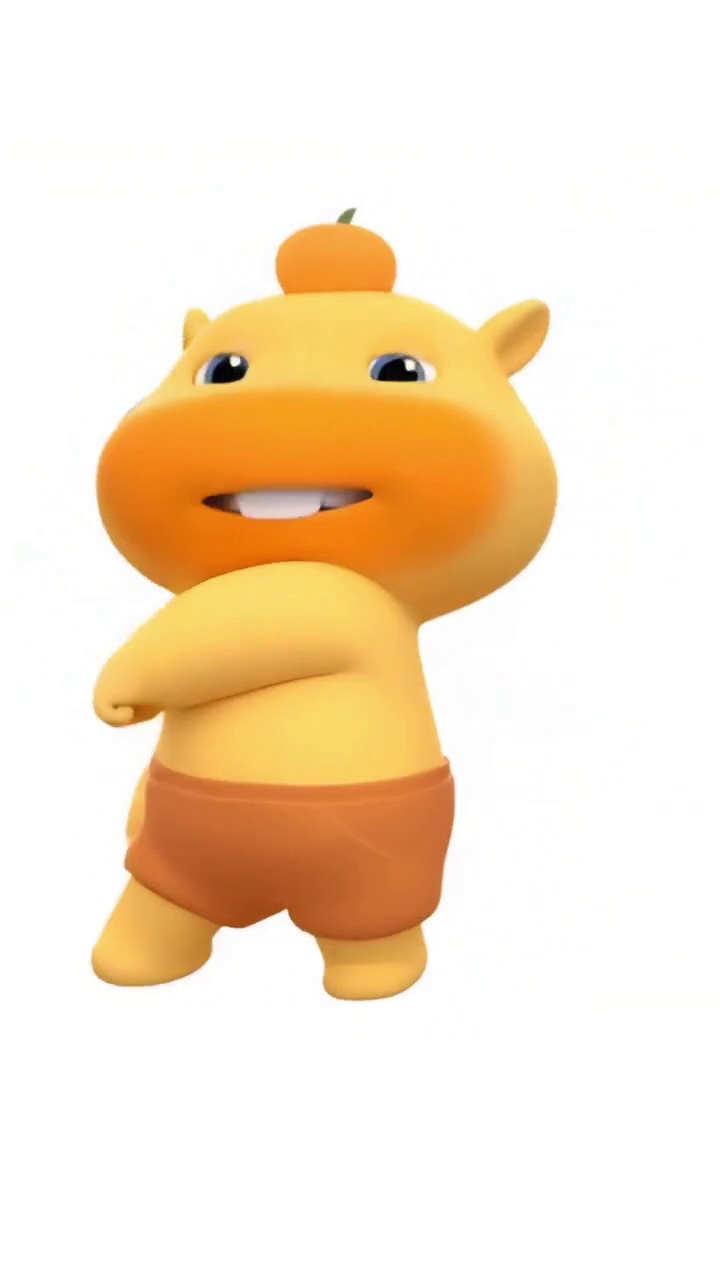} & \includegraphics[width=1.8cm]{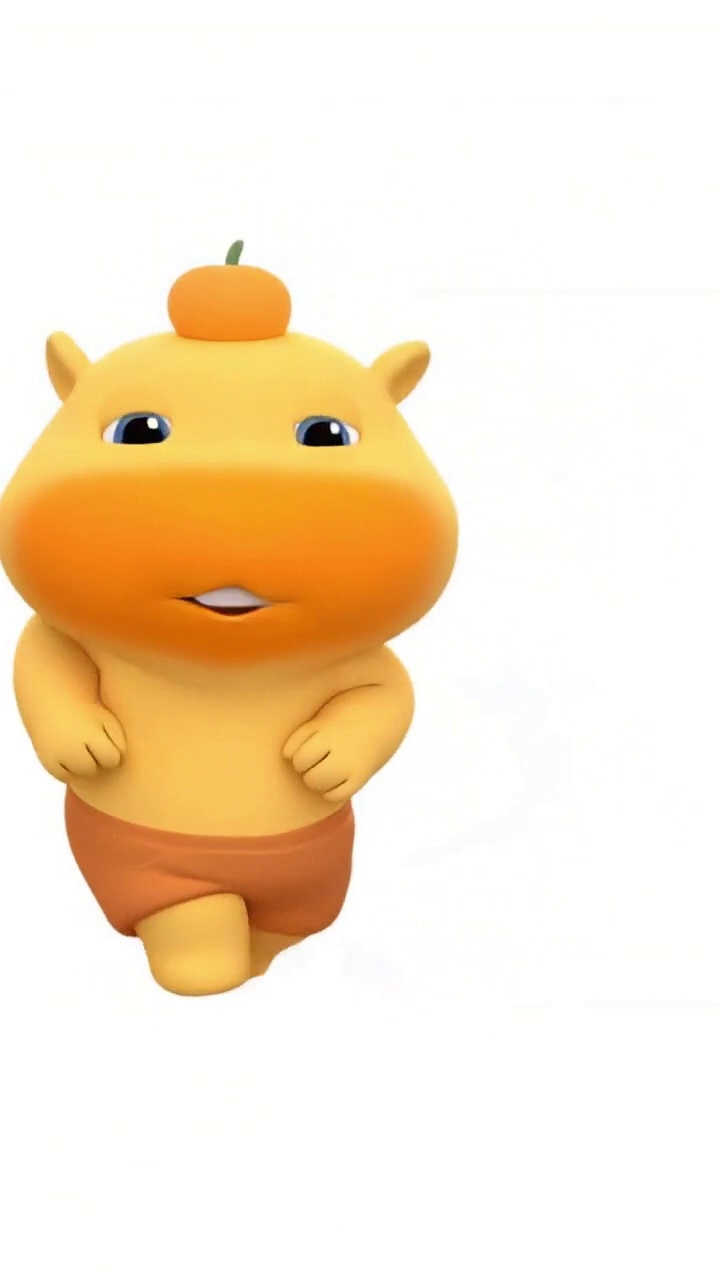} & \includegraphics[width=1.8cm]{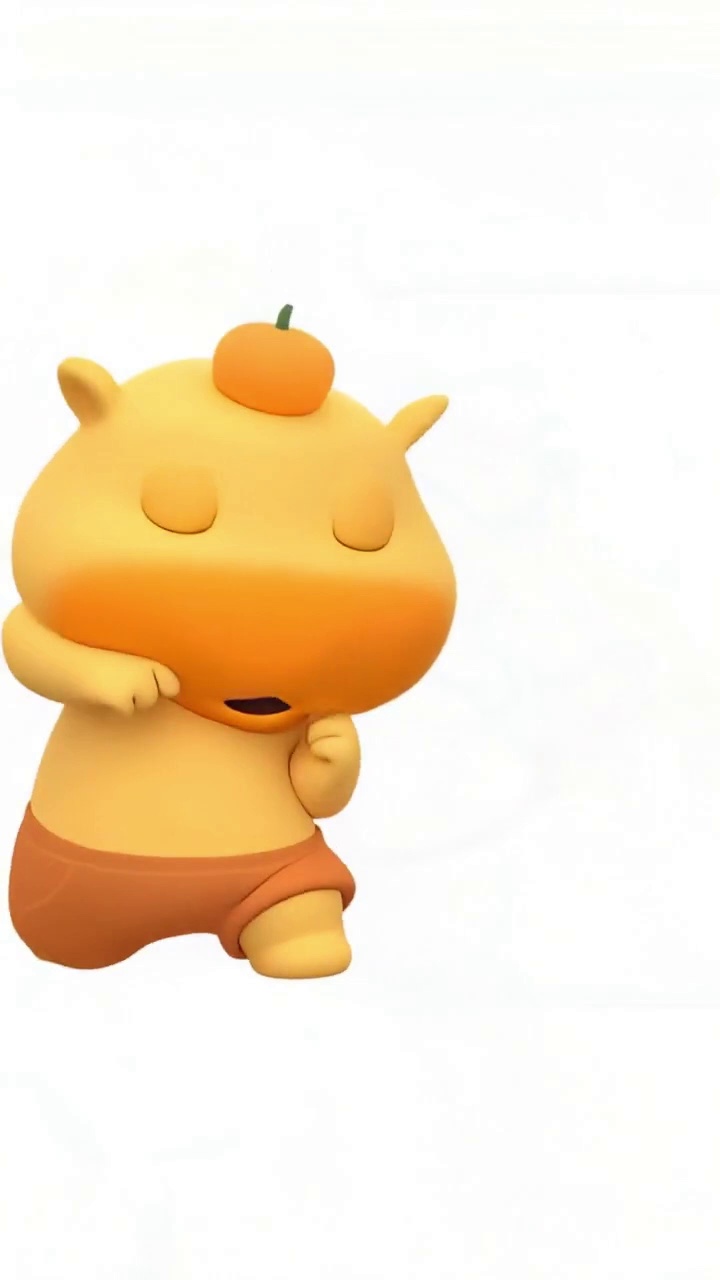} & & \includegraphics[width=1.8cm]{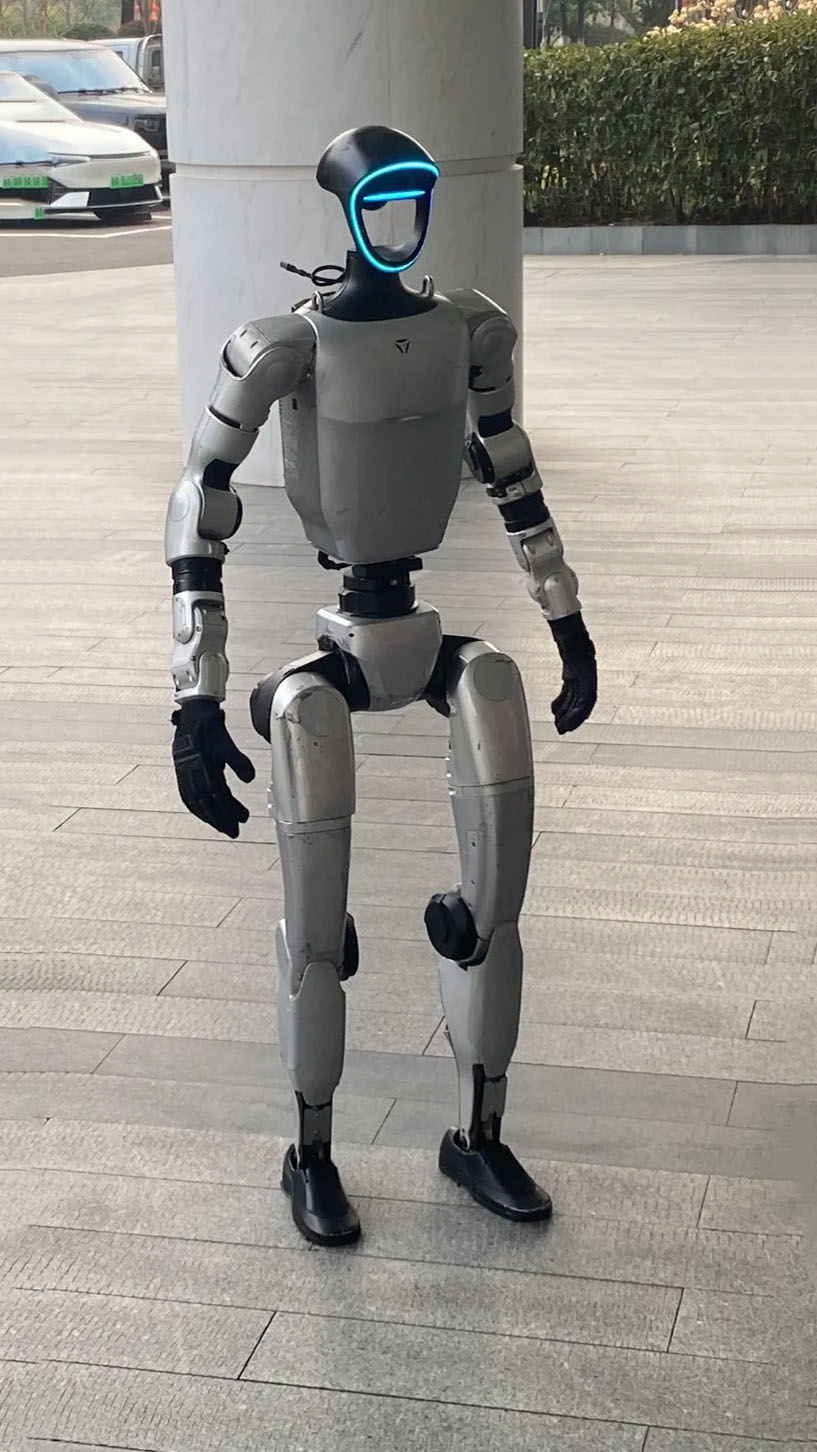} & \includegraphics[width=1.8cm]{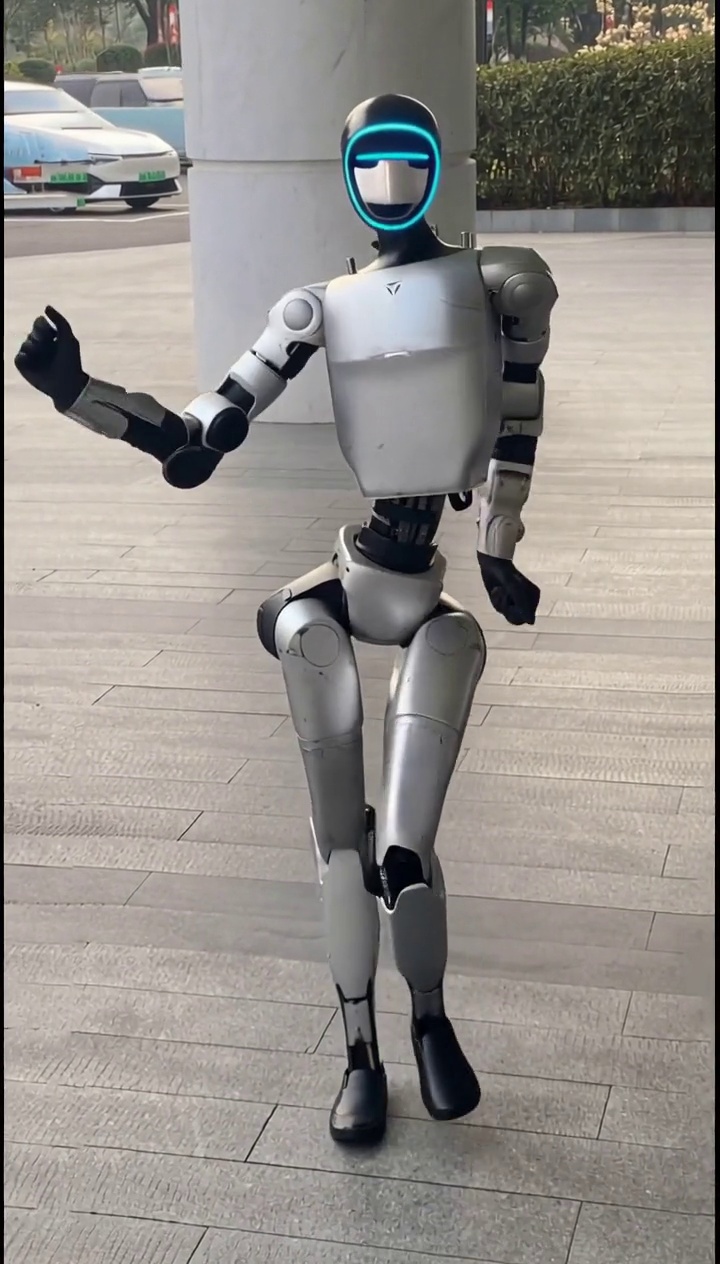} & \includegraphics[width=1.8cm]{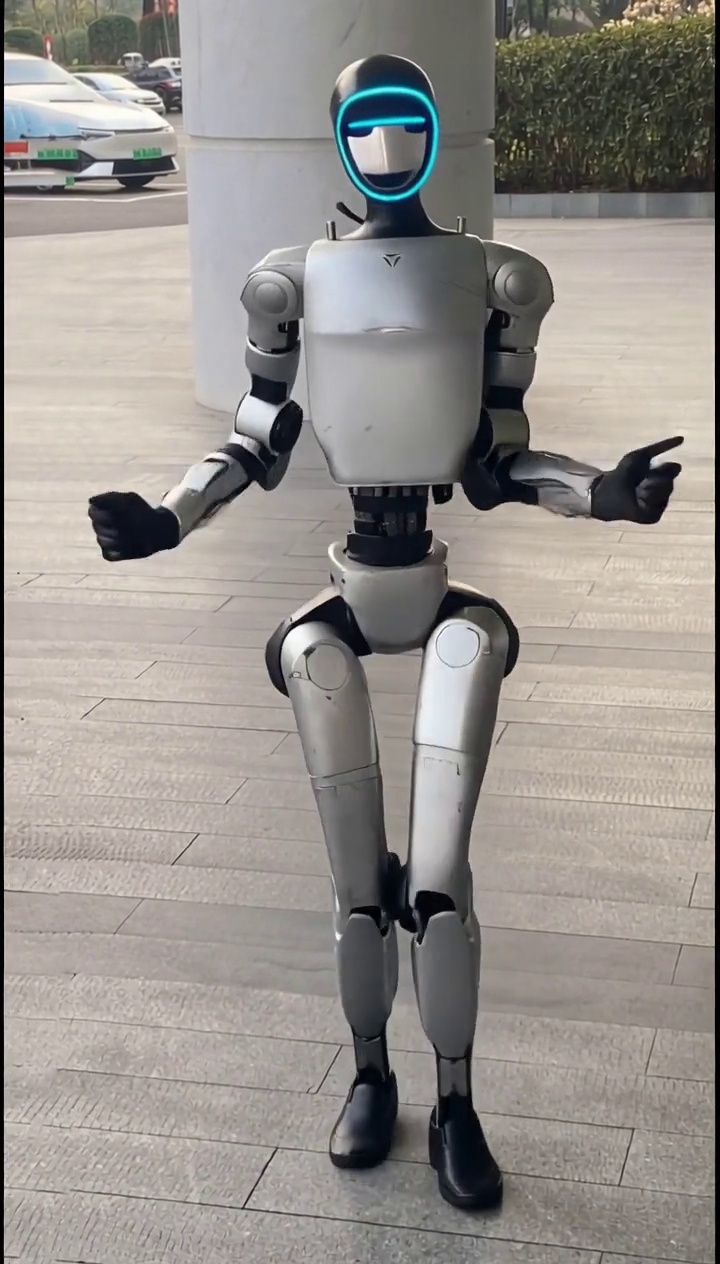} & \includegraphics[width=1.8cm]{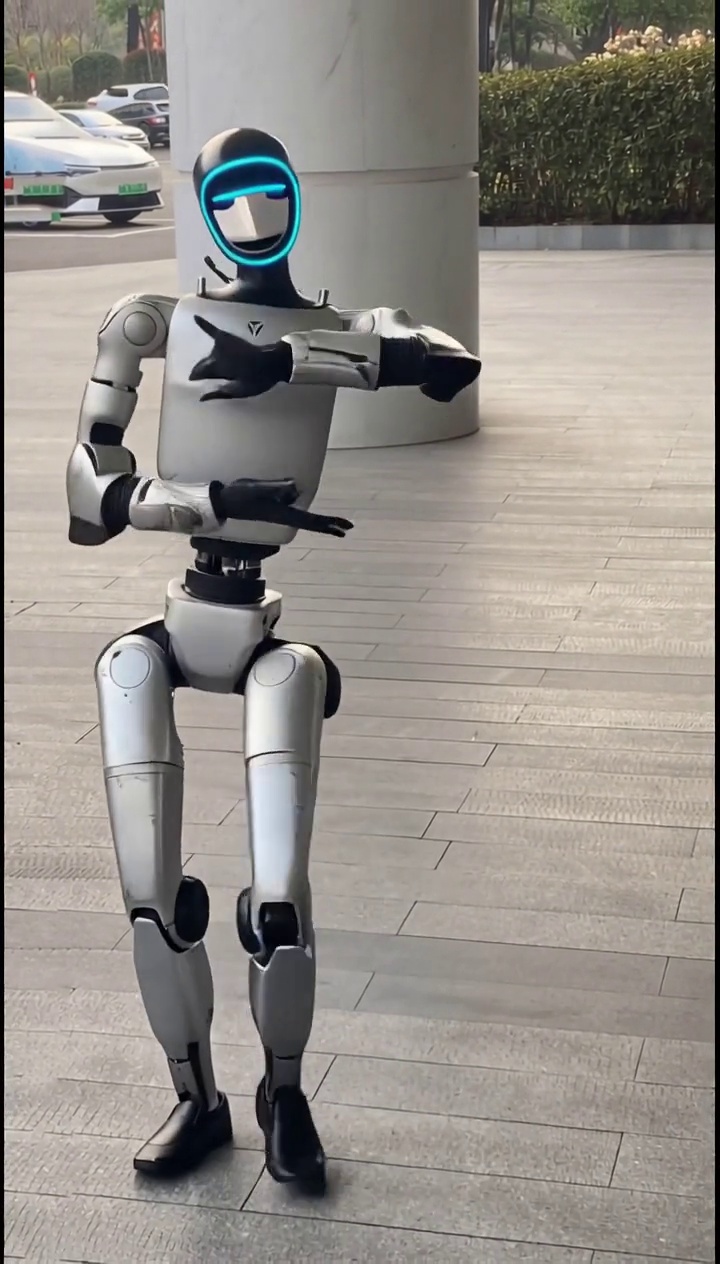} \\

    & \multicolumn{3}{c}{\scriptsize Reference Video} & & & \multicolumn{3}{c}{\scriptsize Reference Video} \\
    & \includegraphics[width=1.8cm]{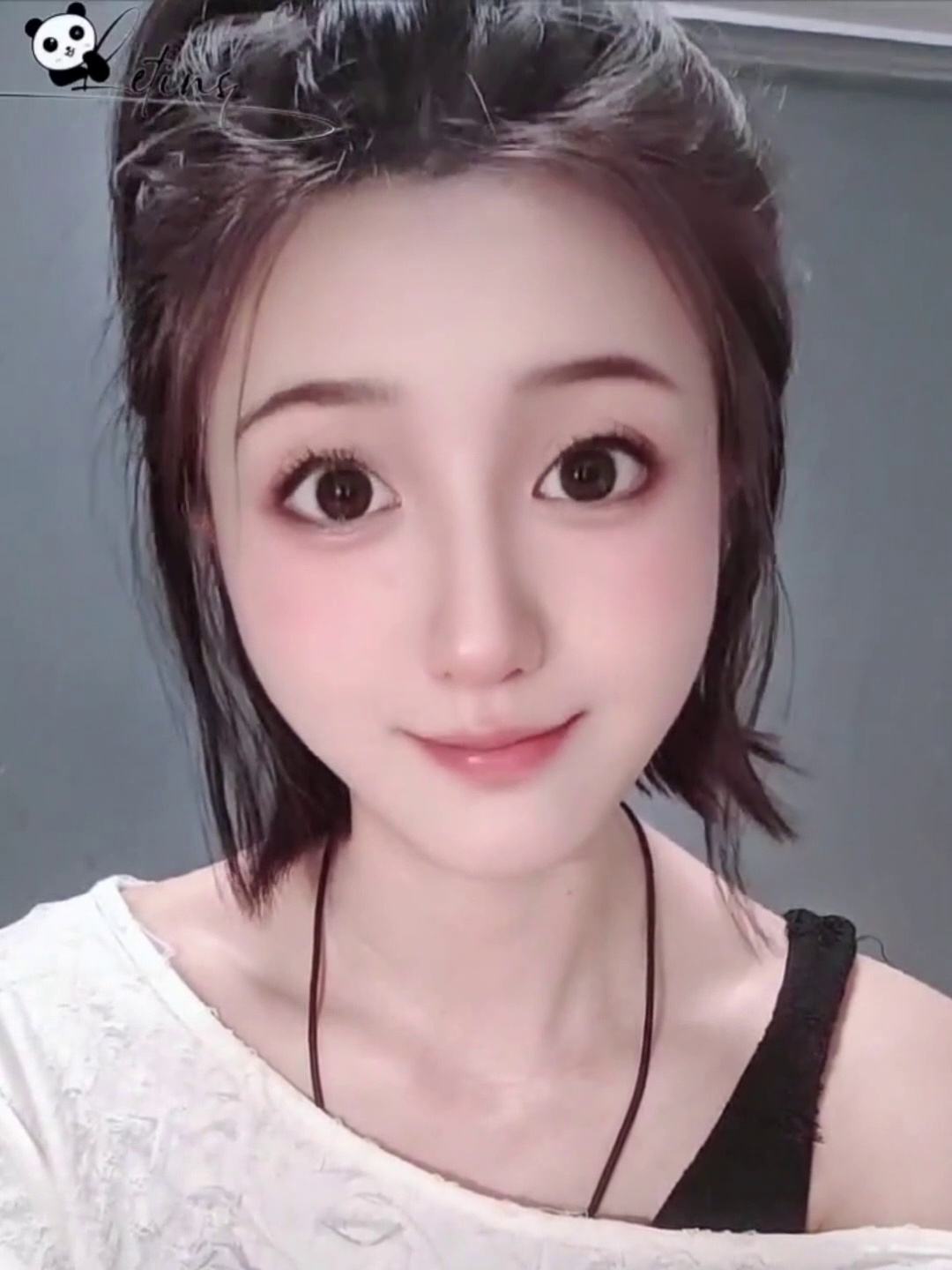} & \includegraphics[width=1.8cm]{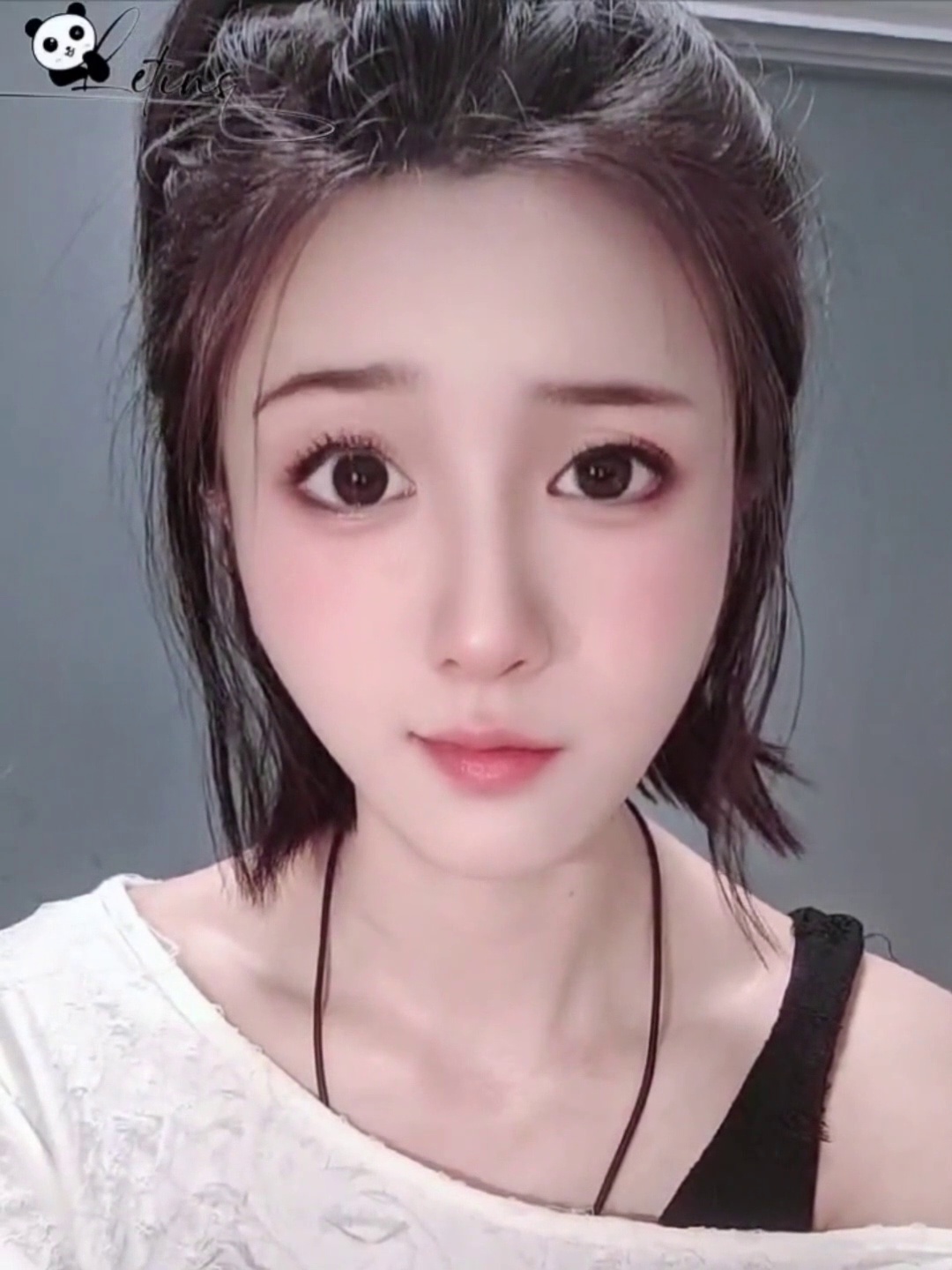} & \includegraphics[width=1.8cm]{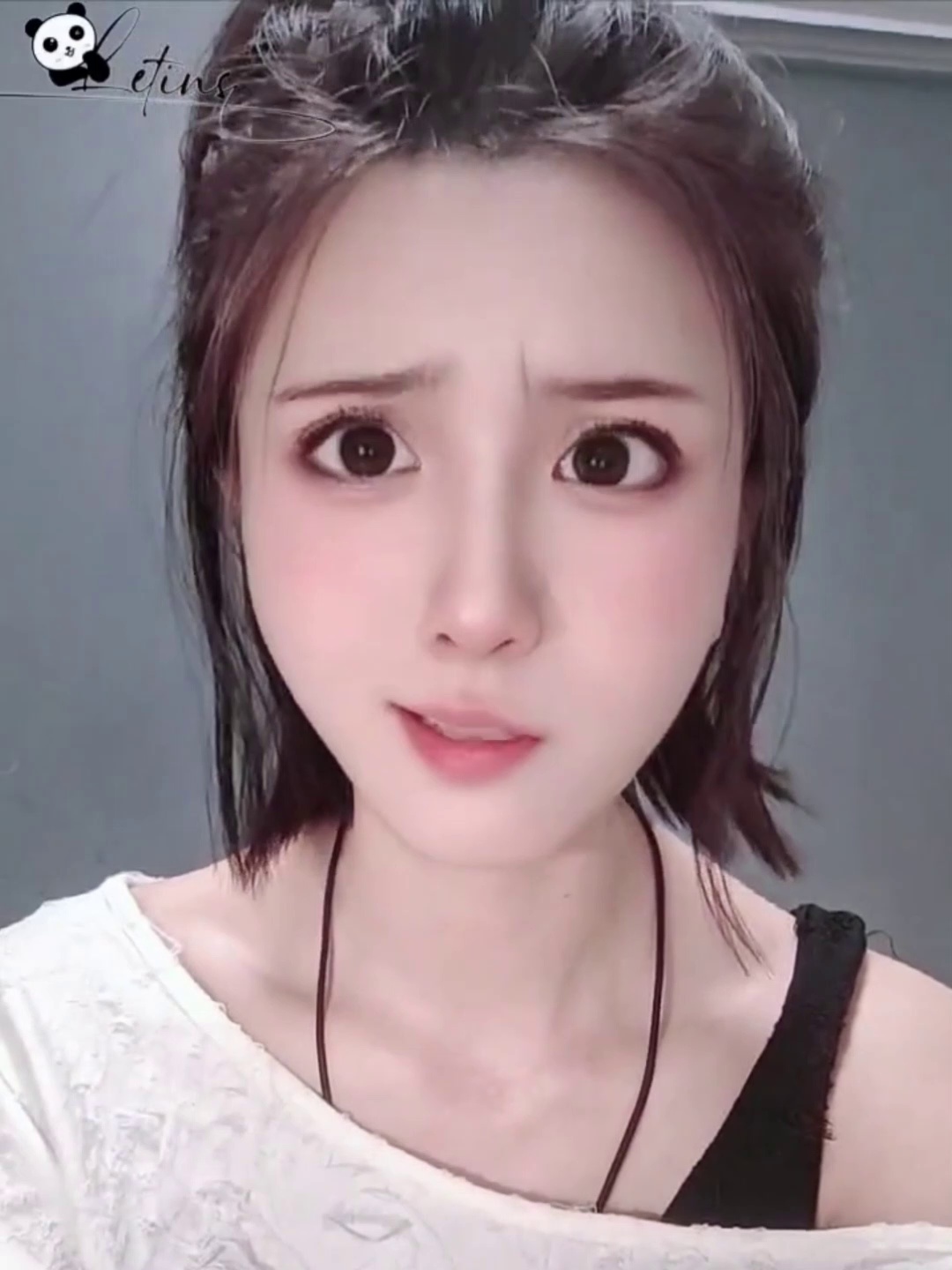} & & & \includegraphics[width=1.8cm]{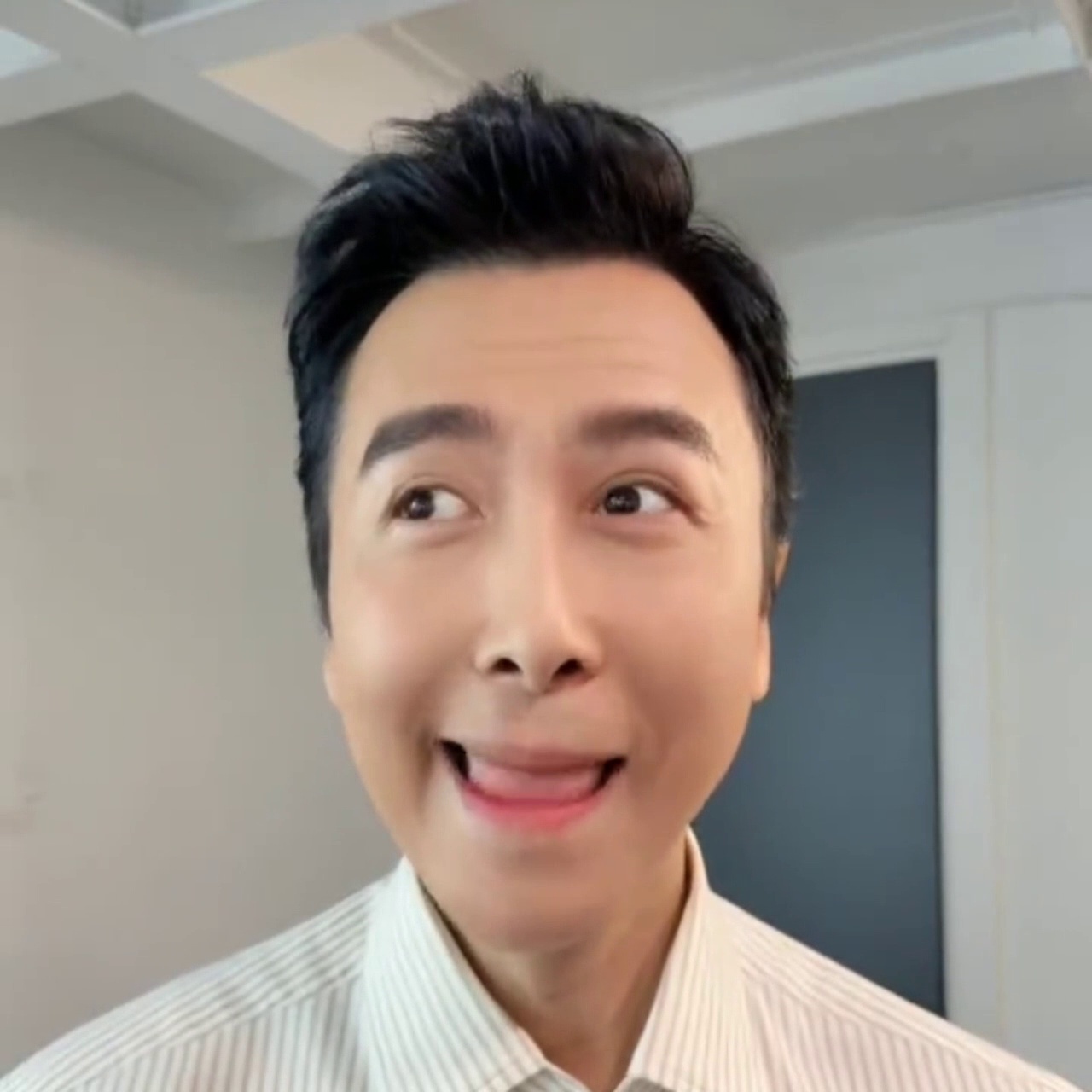} & \includegraphics[width=1.8cm]{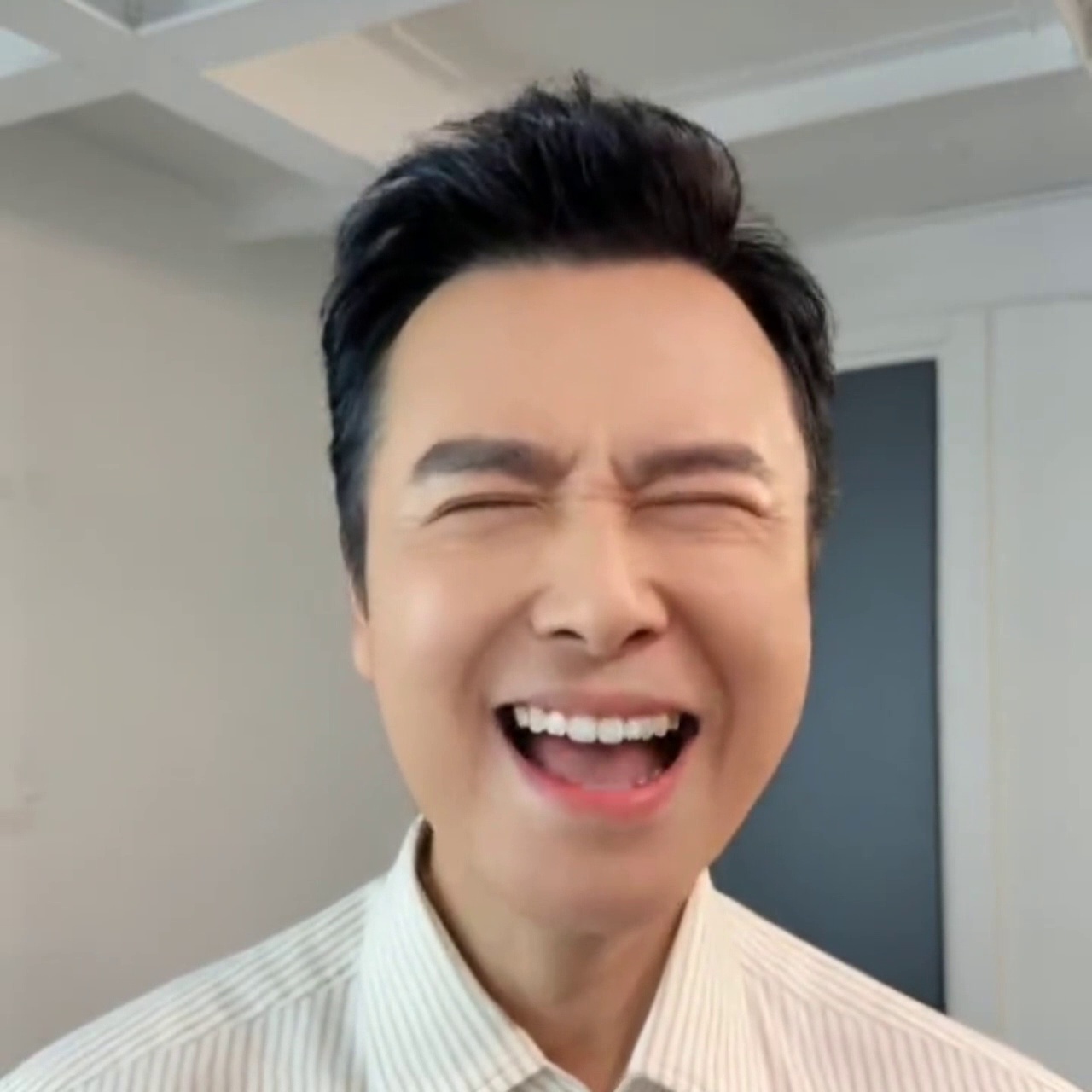} & \includegraphics[width=1.8cm]{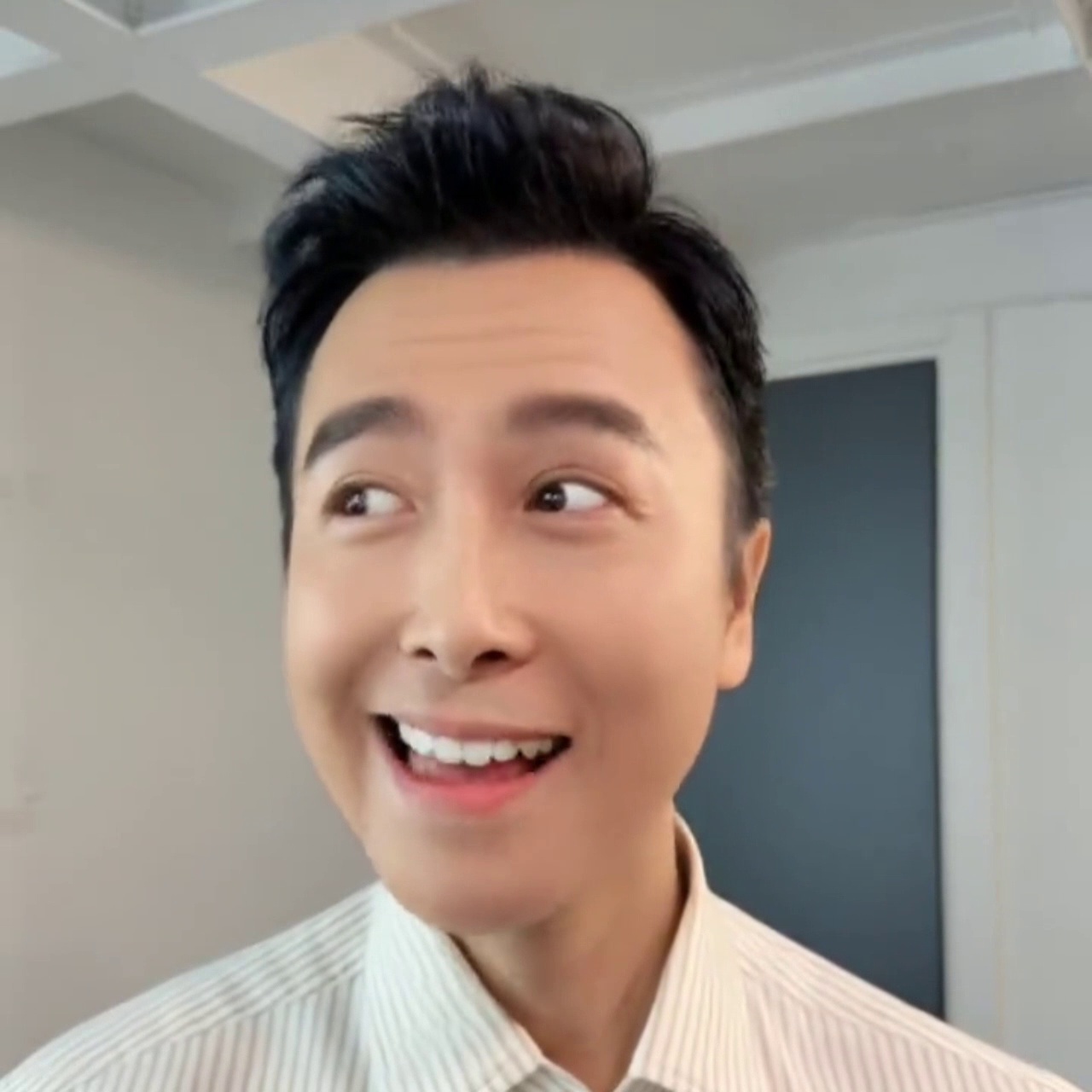} \\

    \scriptsize Reference Image & \multicolumn{3}{c}{\scriptsize Animation Result} & & \scriptsize Reference Image & \multicolumn{3}{c}{\scriptsize Animation Result} \\
    \includegraphics[width=1.8cm]{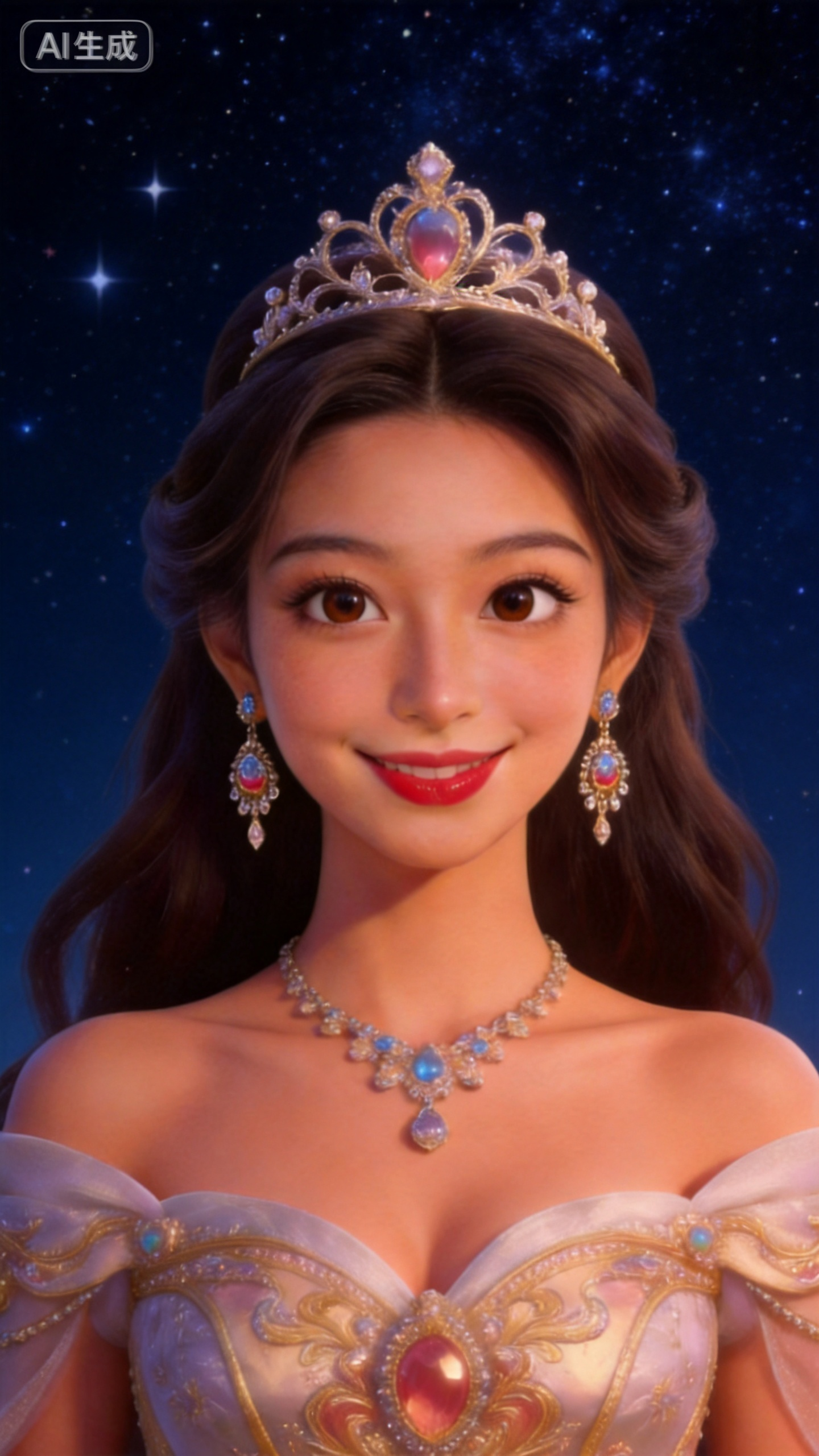} & \includegraphics[width=1.8cm]{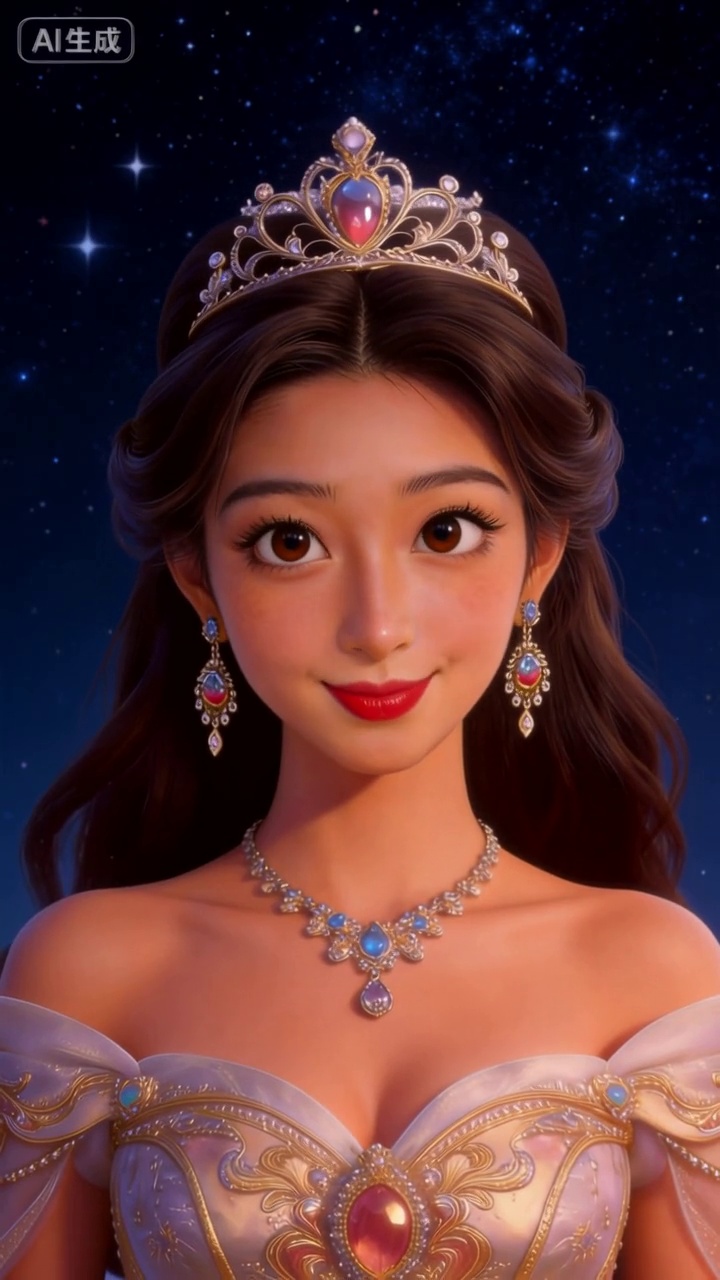} & \includegraphics[width=1.8cm]{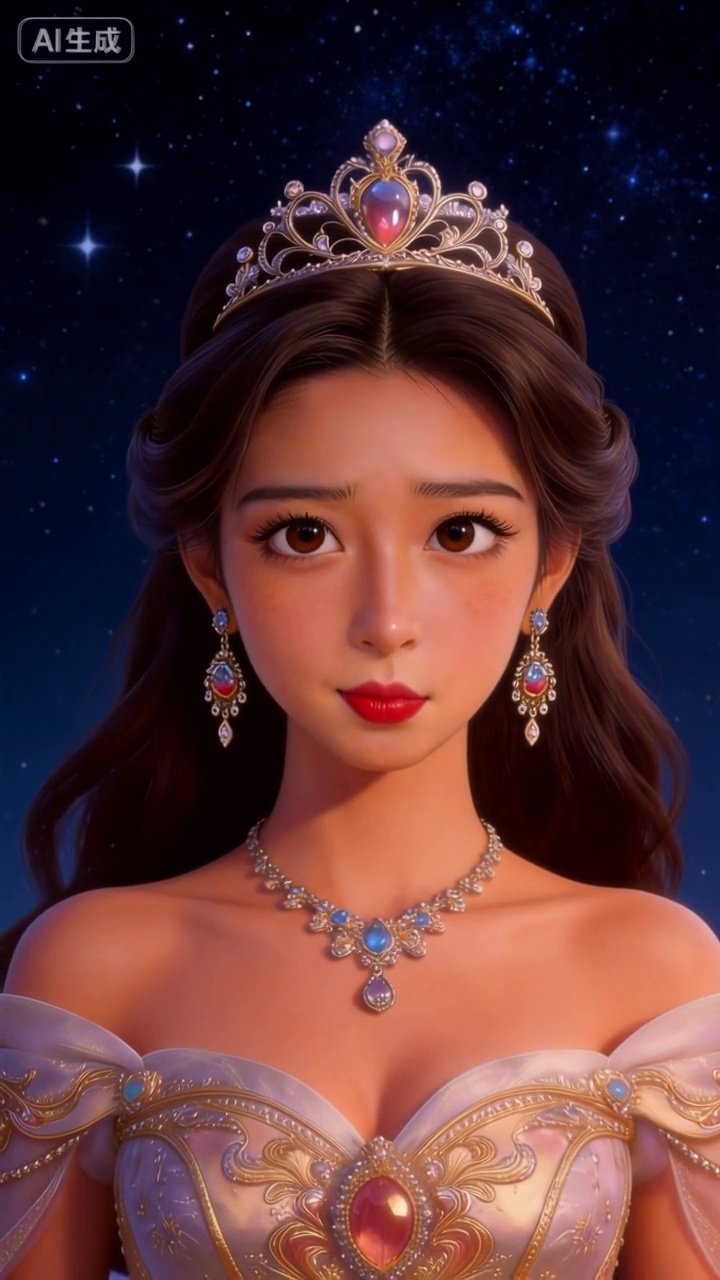} & \includegraphics[width=1.8cm]{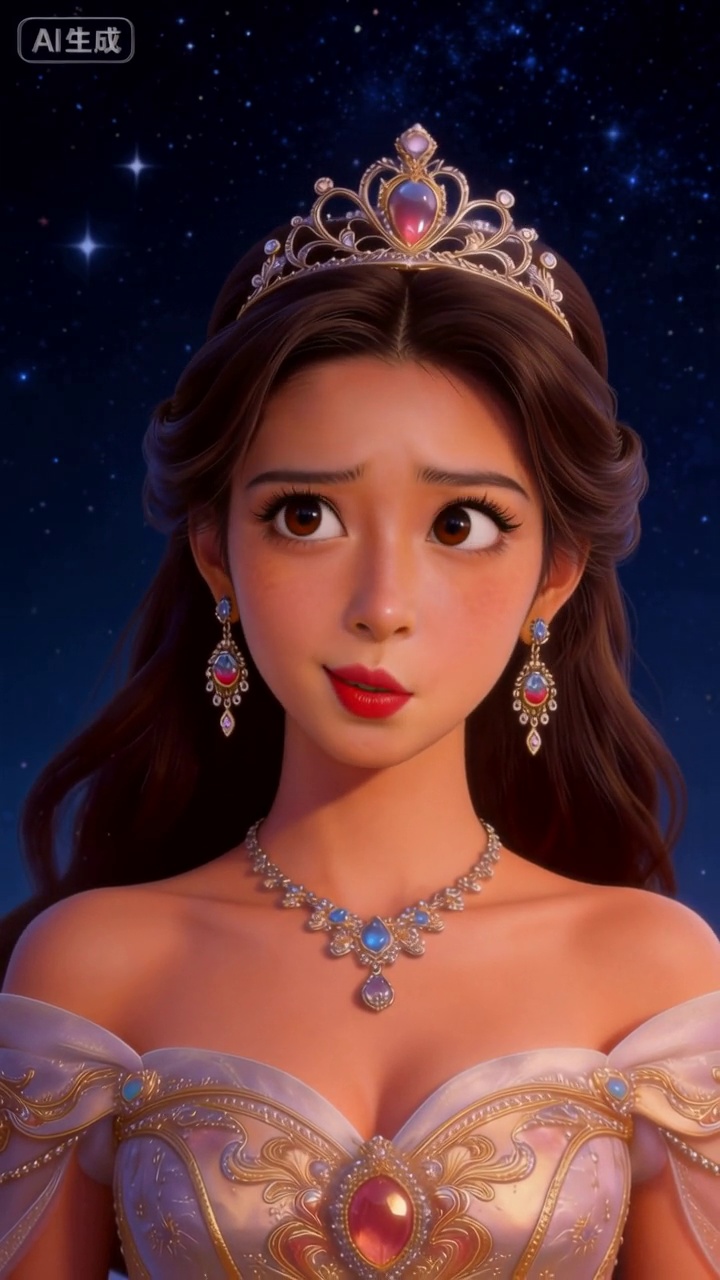} & & \includegraphics[width=1.8cm]{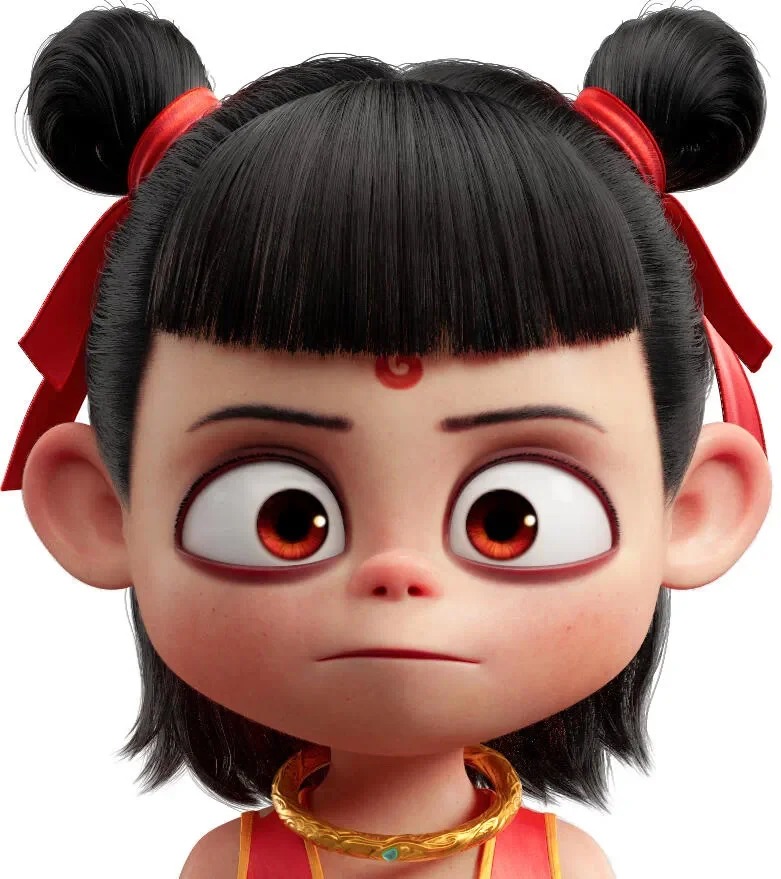} & \includegraphics[width=1.8cm]{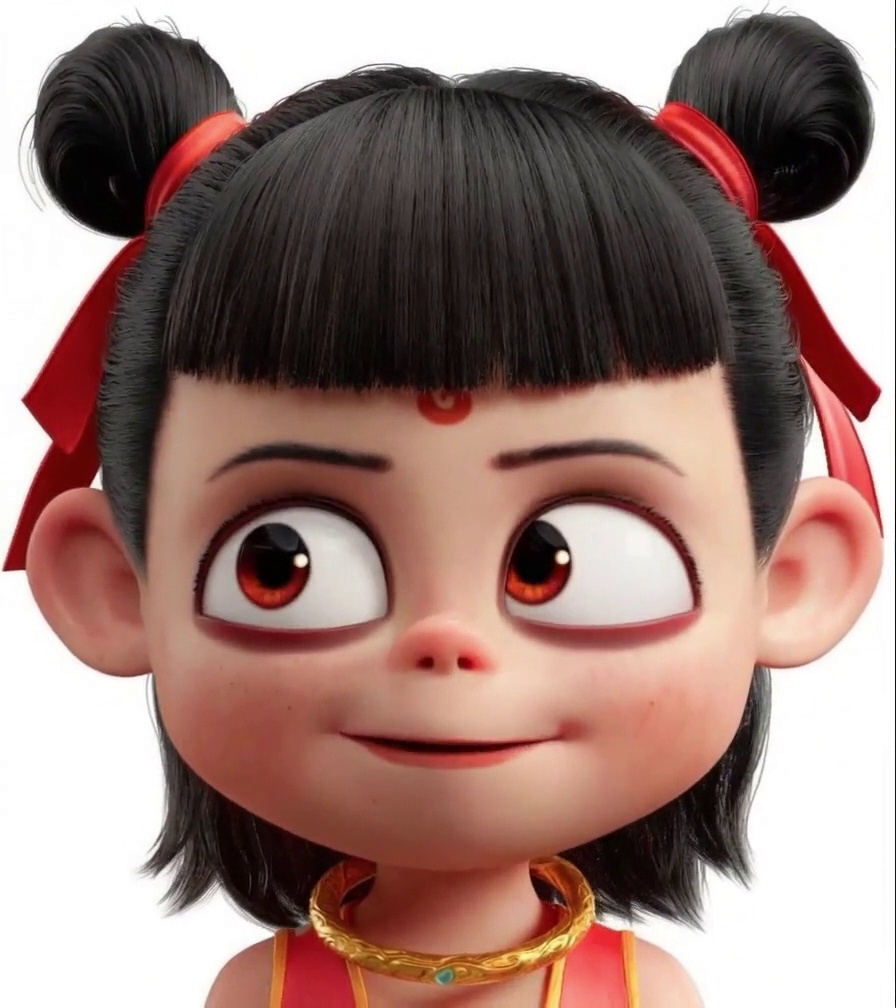} & \includegraphics[width=1.8cm]{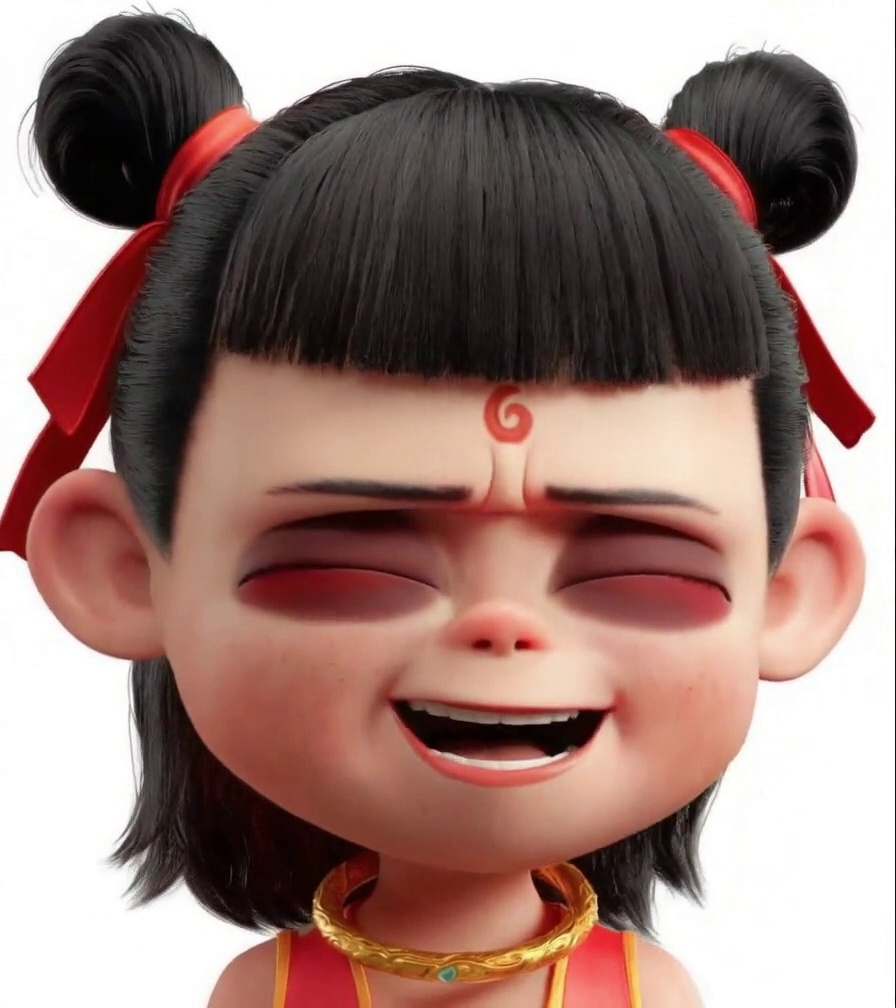} & \includegraphics[width=1.8cm]{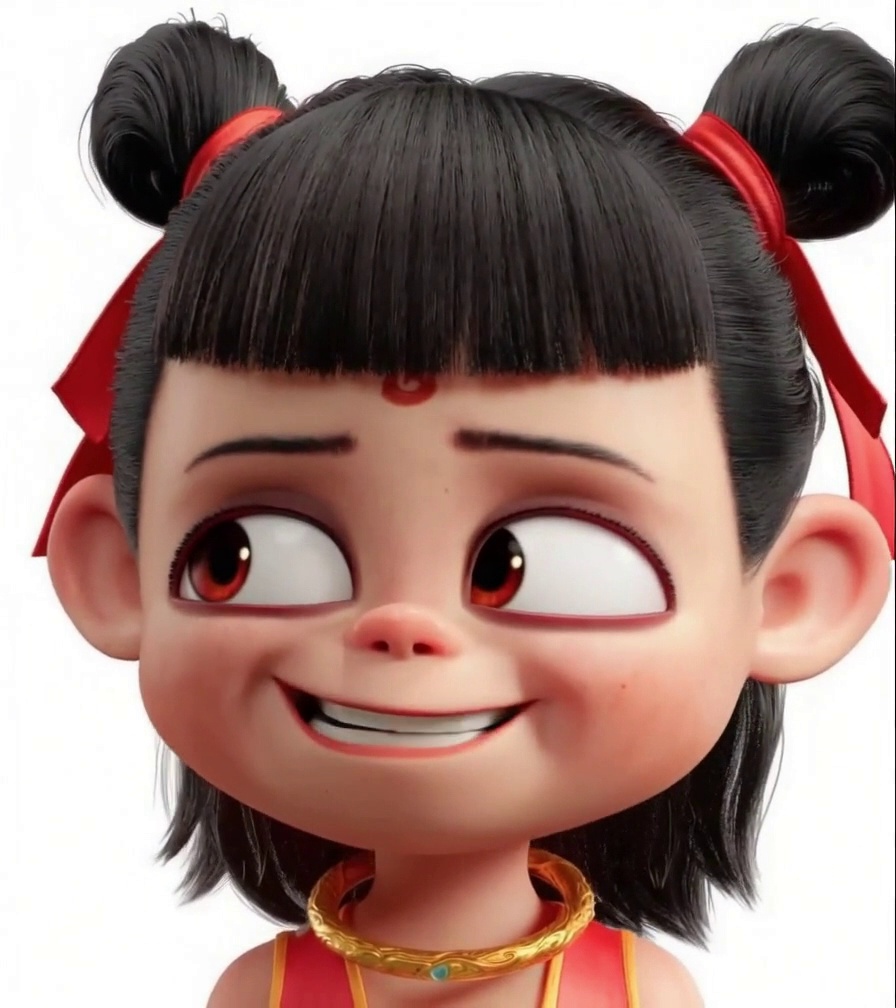} \\

    & \multicolumn{3}{c}{\scriptsize Reference Video} & & & \multicolumn{3}{c}{\scriptsize Reference Video} \\
    & \includegraphics[width=1.8cm]{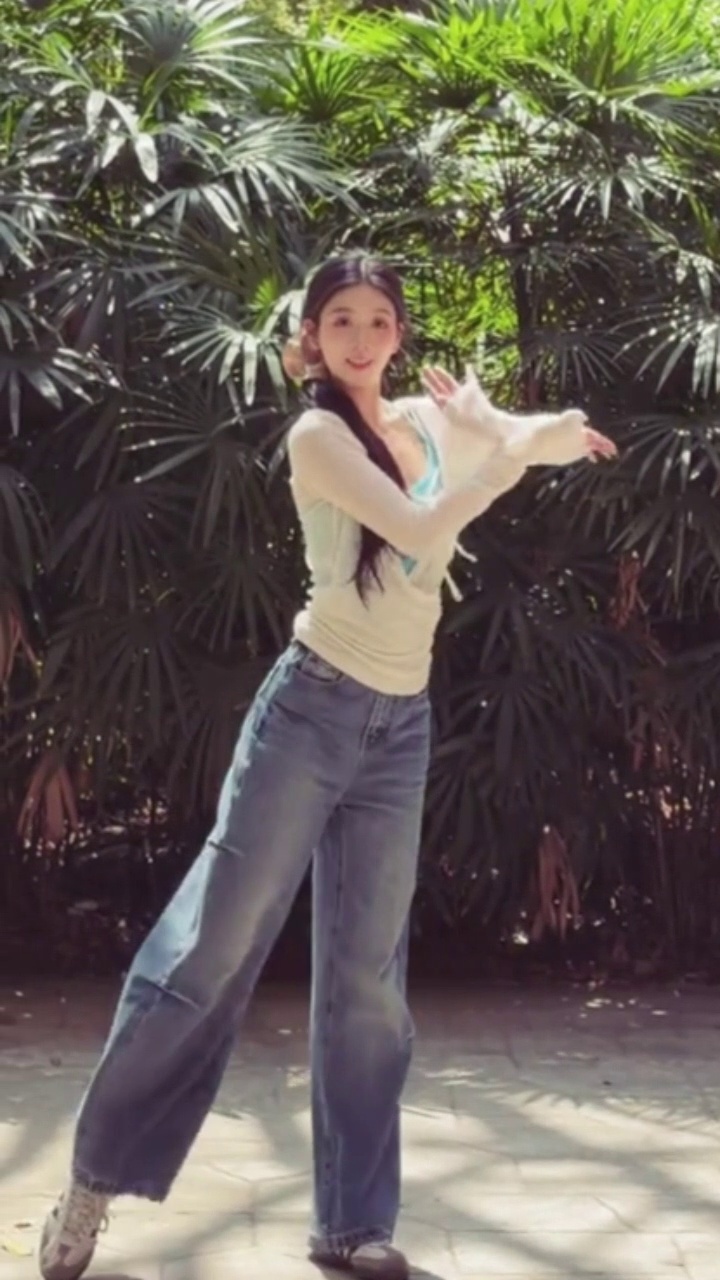} & \includegraphics[width=1.8cm]{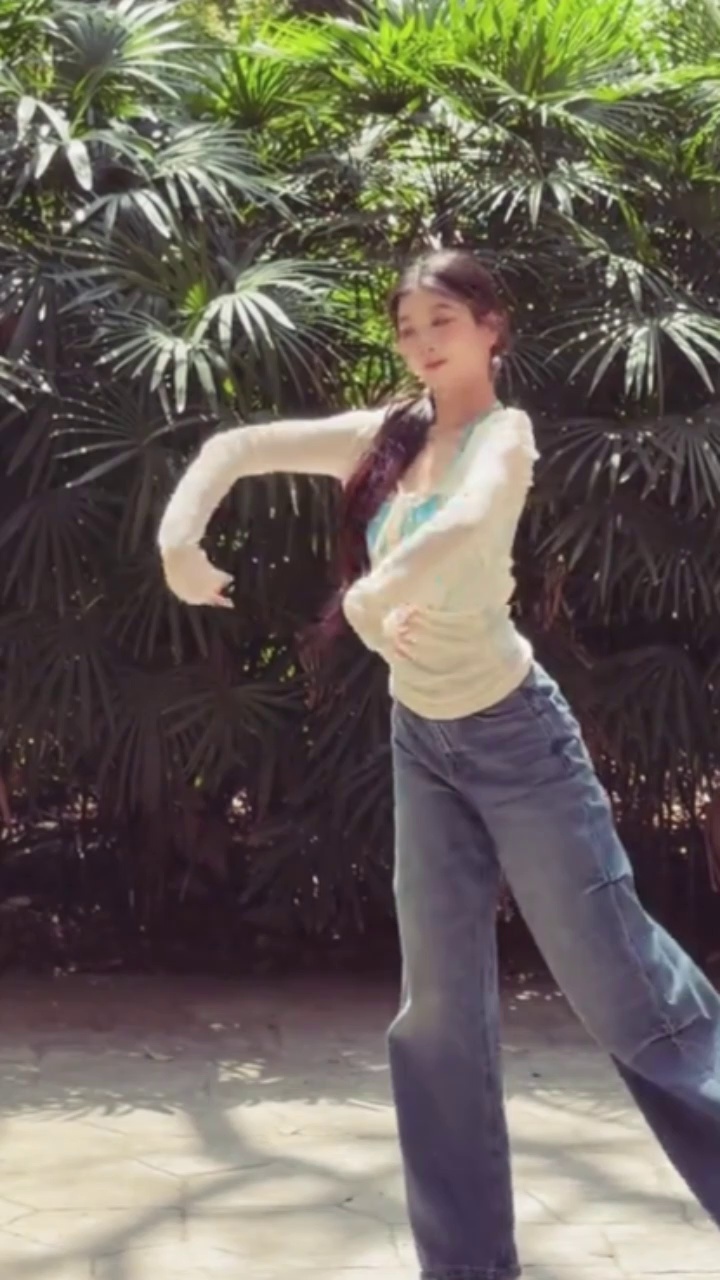} & \includegraphics[width=1.8cm]{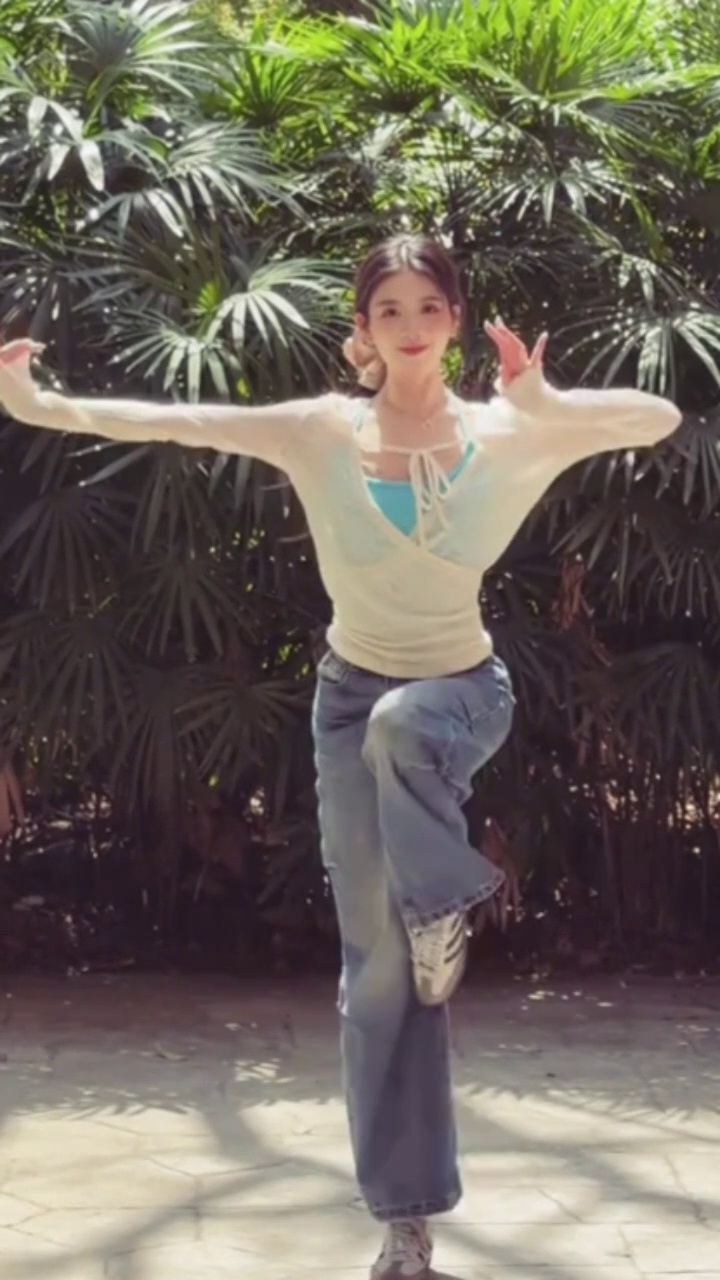} & & & \includegraphics[width=1.8cm]{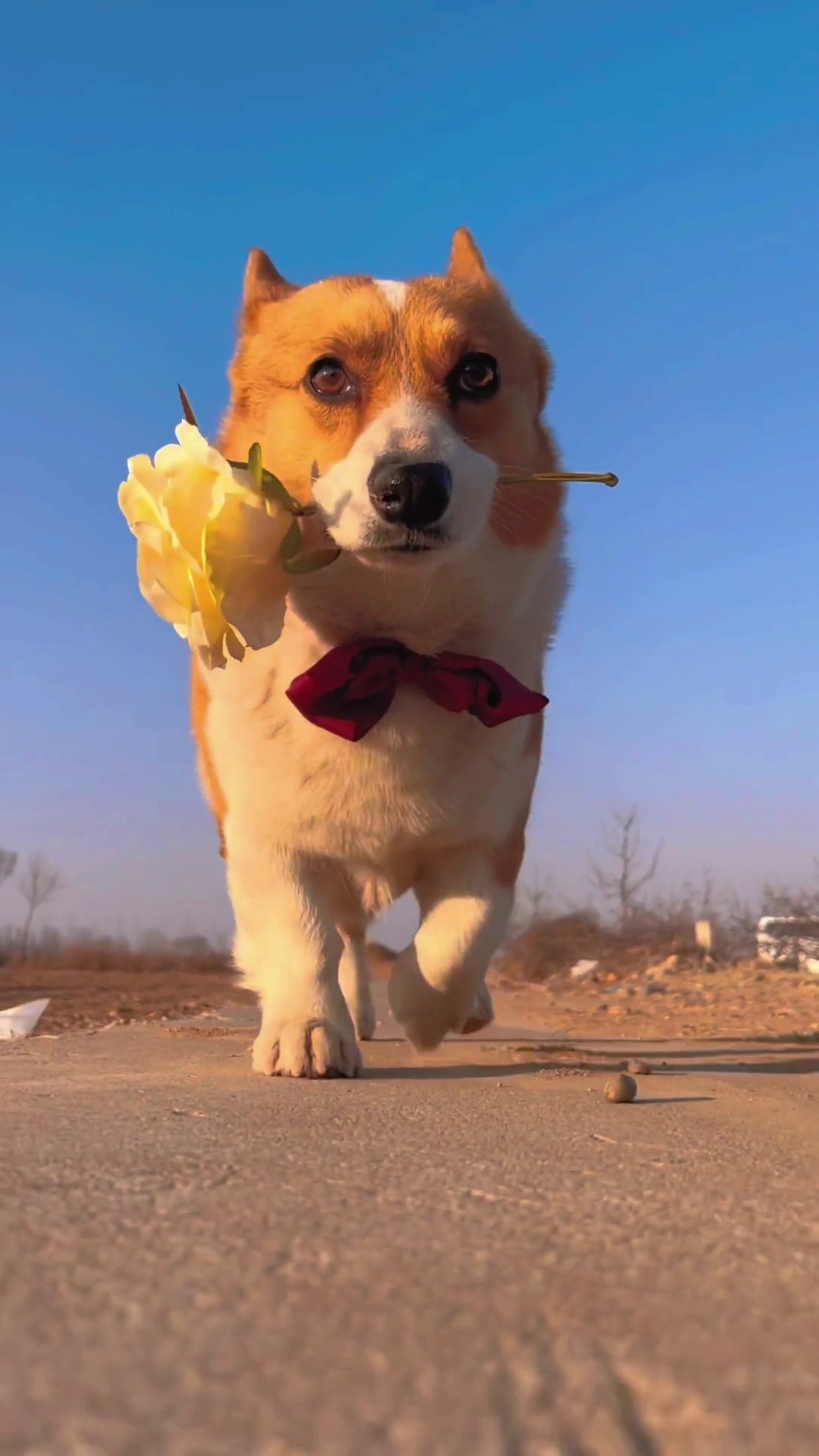} & \includegraphics[width=1.8cm]{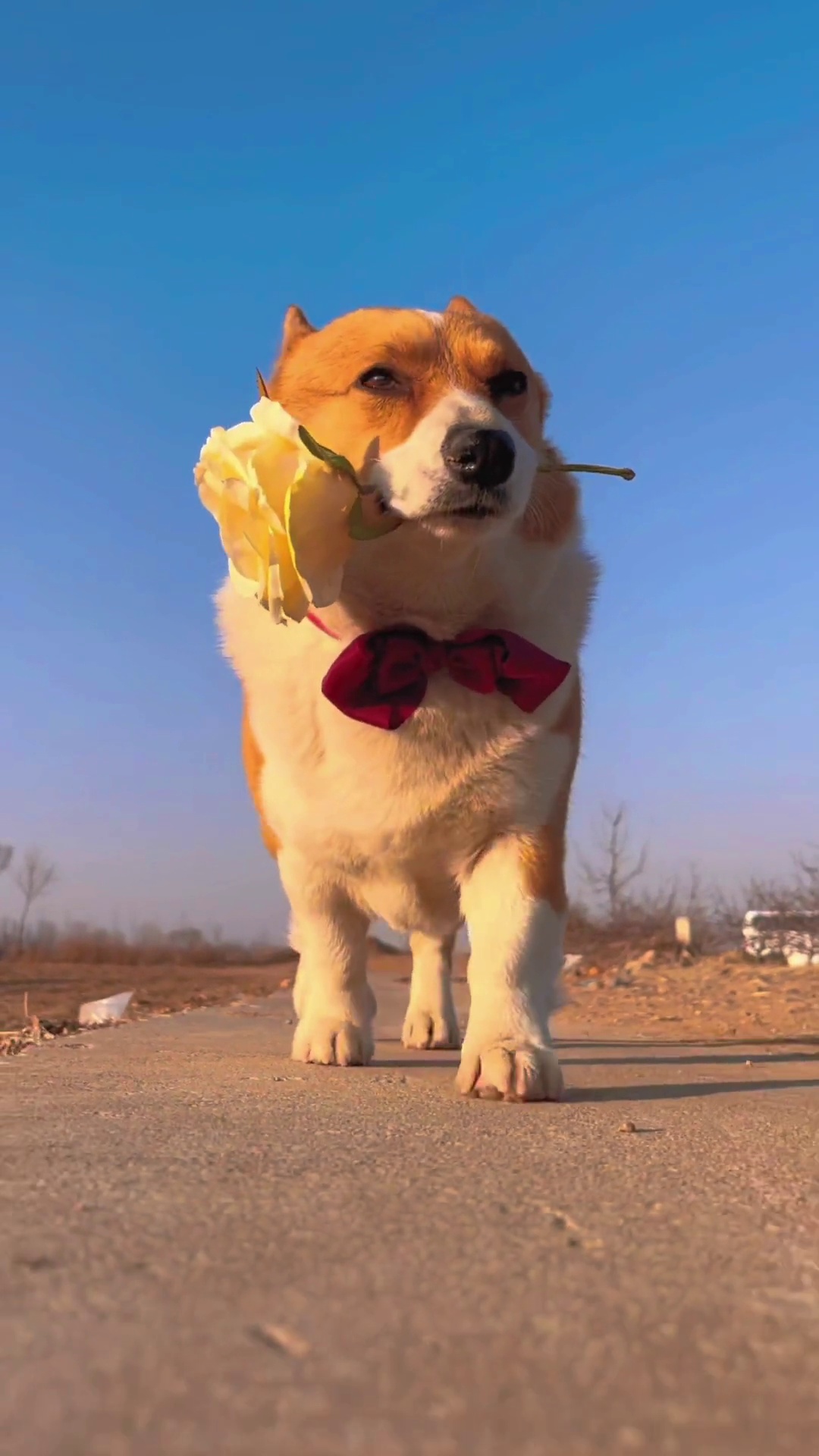} & \includegraphics[width=1.8cm]{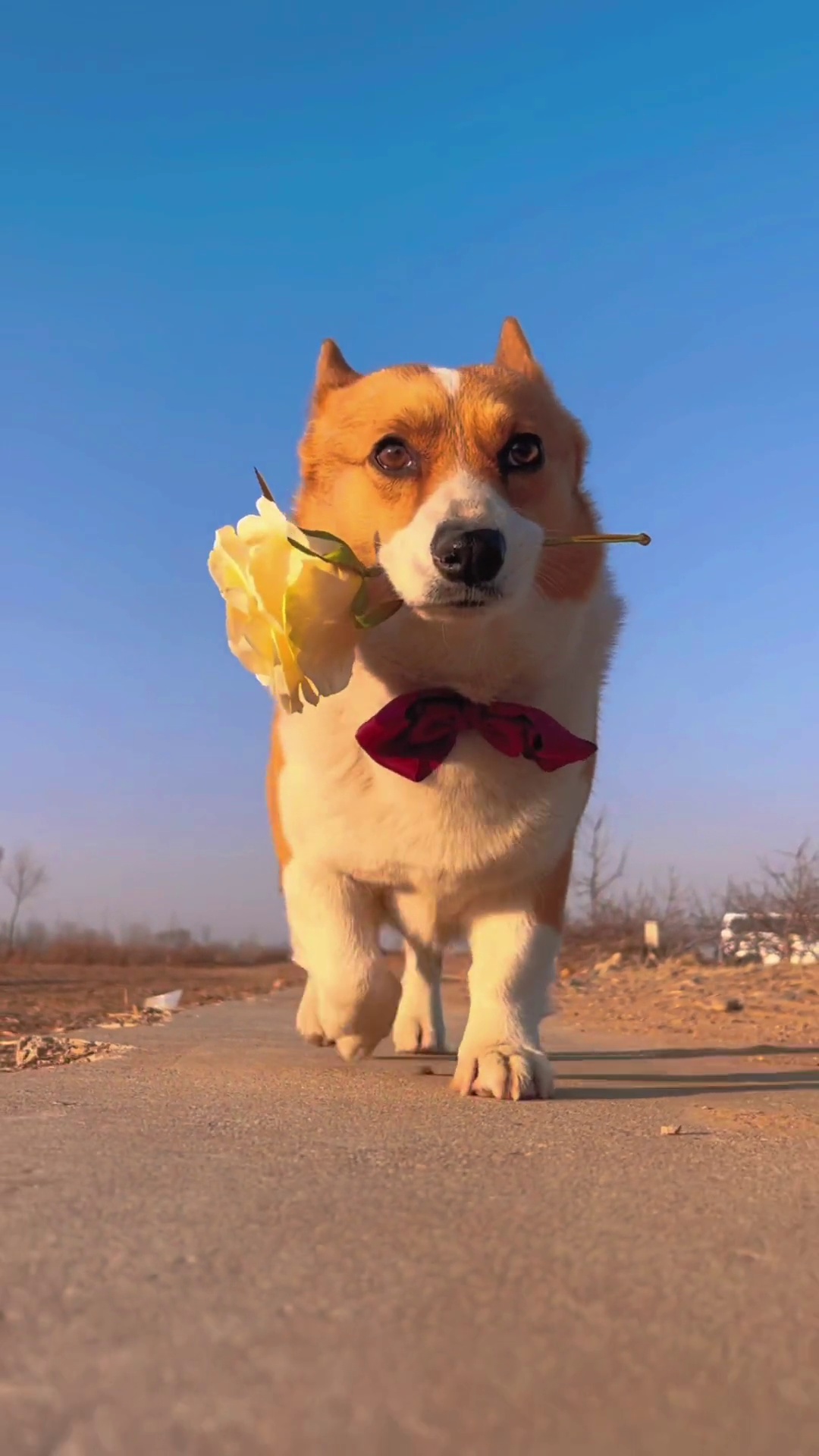} \\

    \scriptsize Reference Image & \multicolumn{3}{c}{\scriptsize Animation Result} & & \scriptsize Reference Image & \multicolumn{3}{c}{\scriptsize Animation Result} \\
    \includegraphics[width=1.8cm]{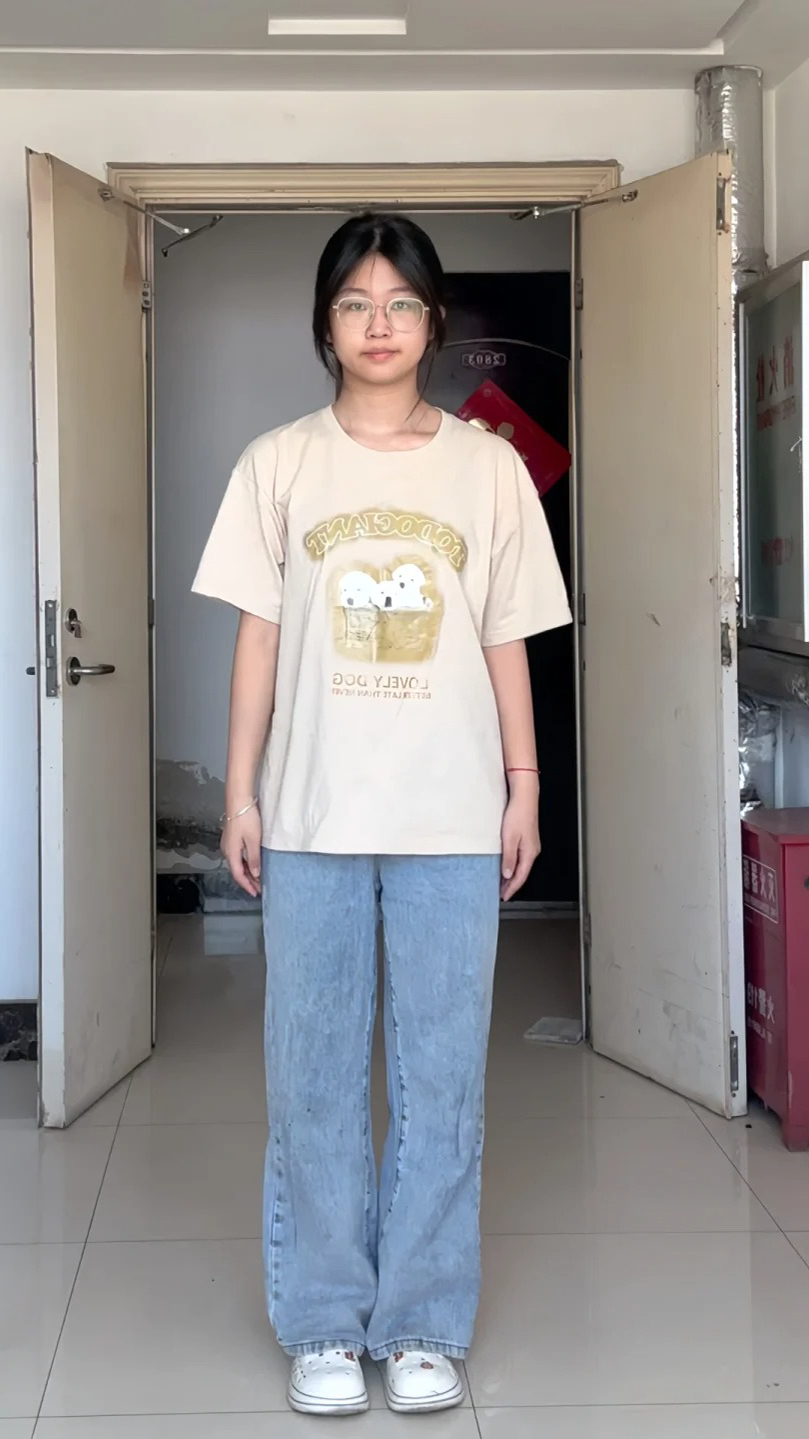} & \includegraphics[width=1.8cm]{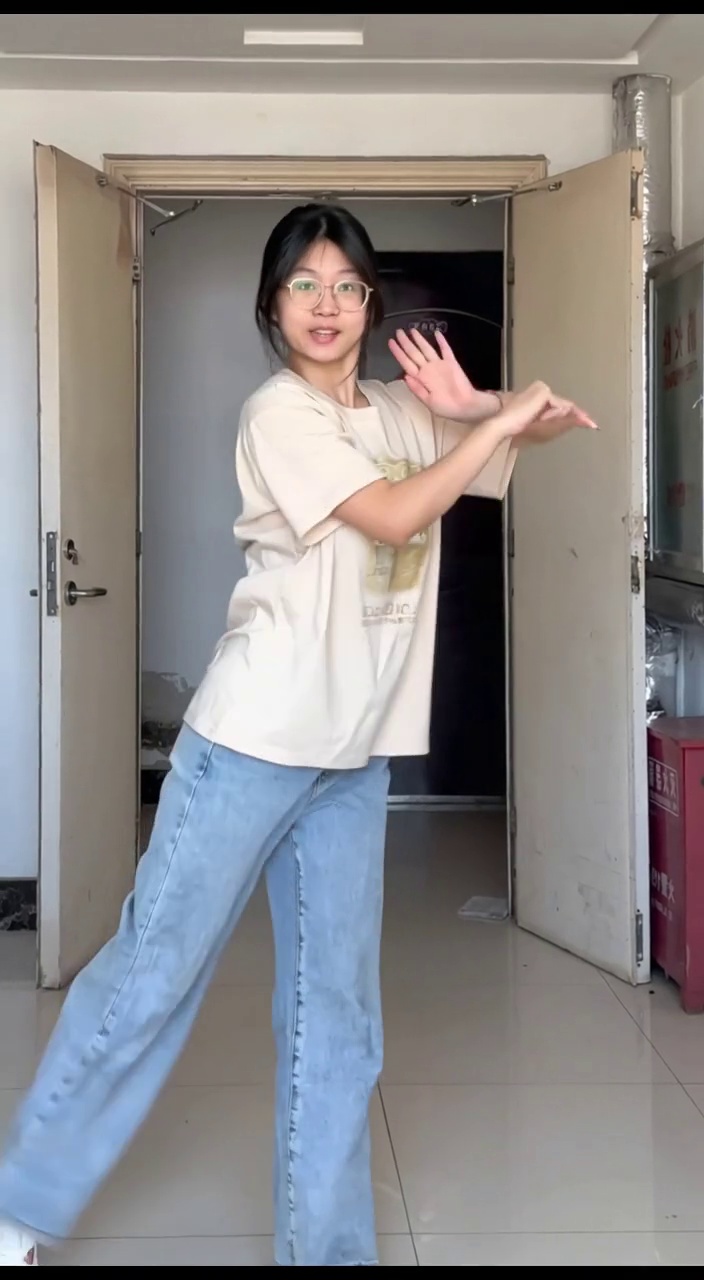} & \includegraphics[width=1.8cm]{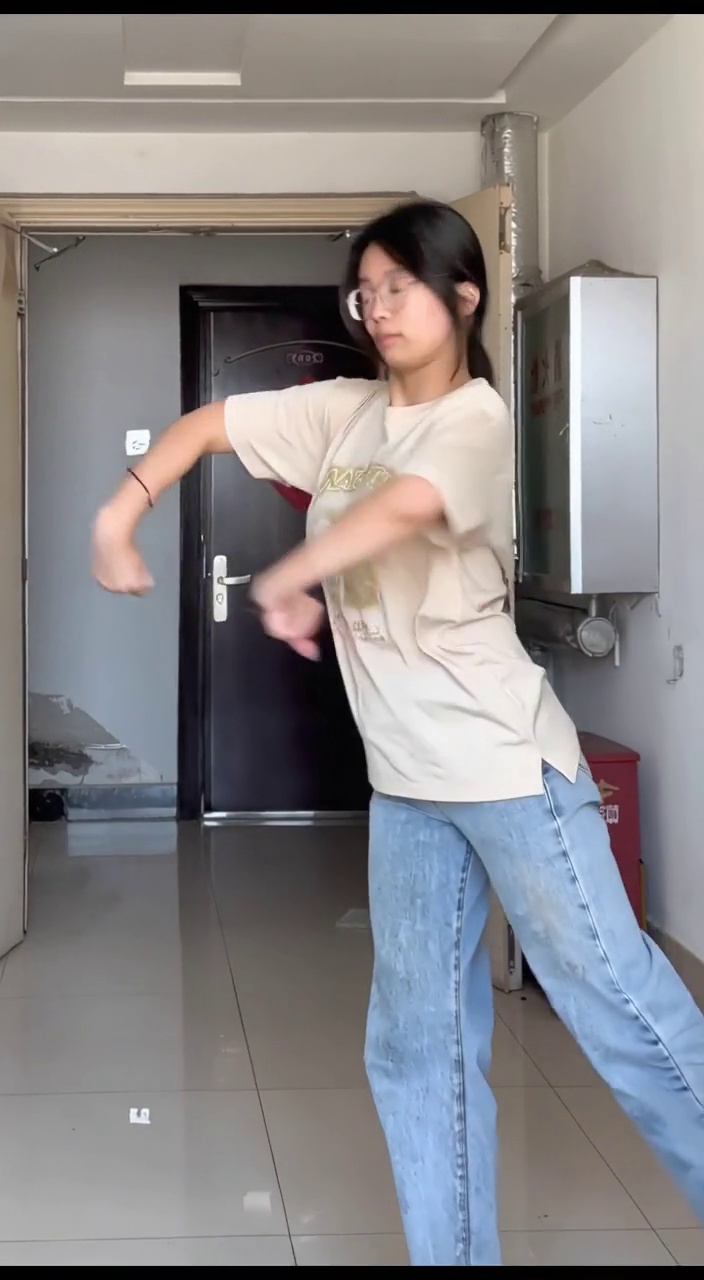} & \includegraphics[width=1.8cm]{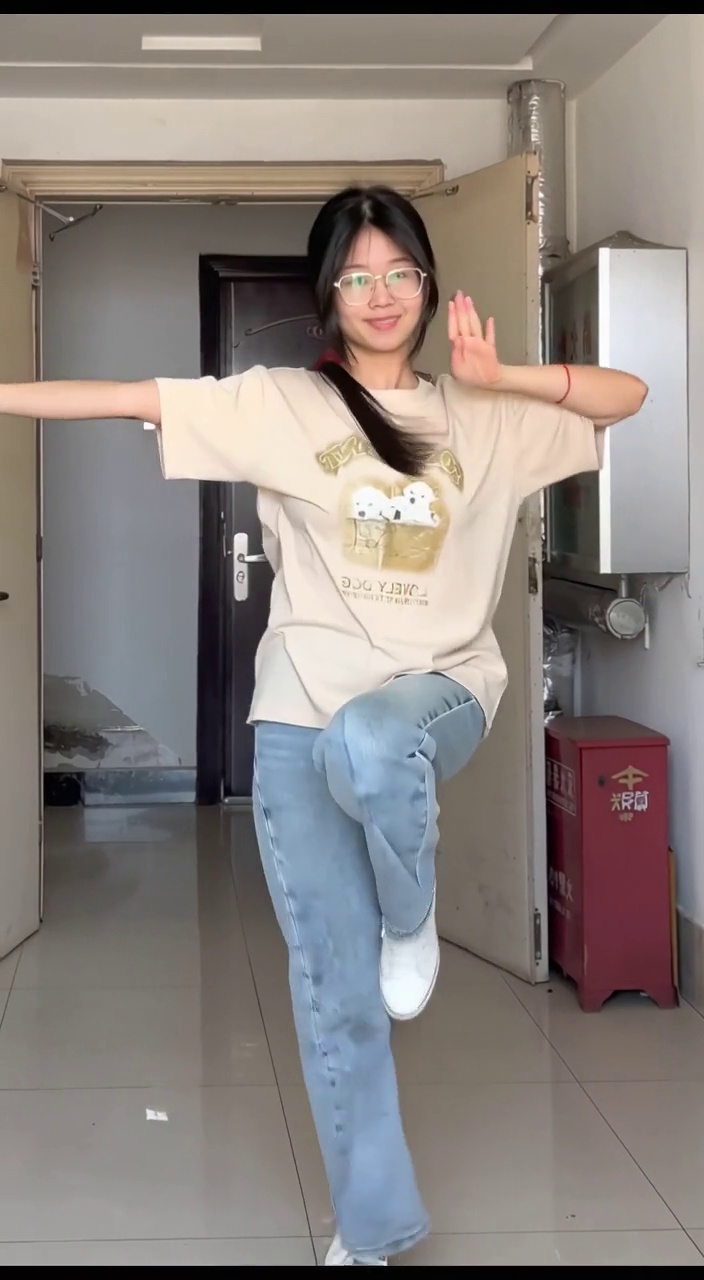} & & \includegraphics[width=1.8cm]{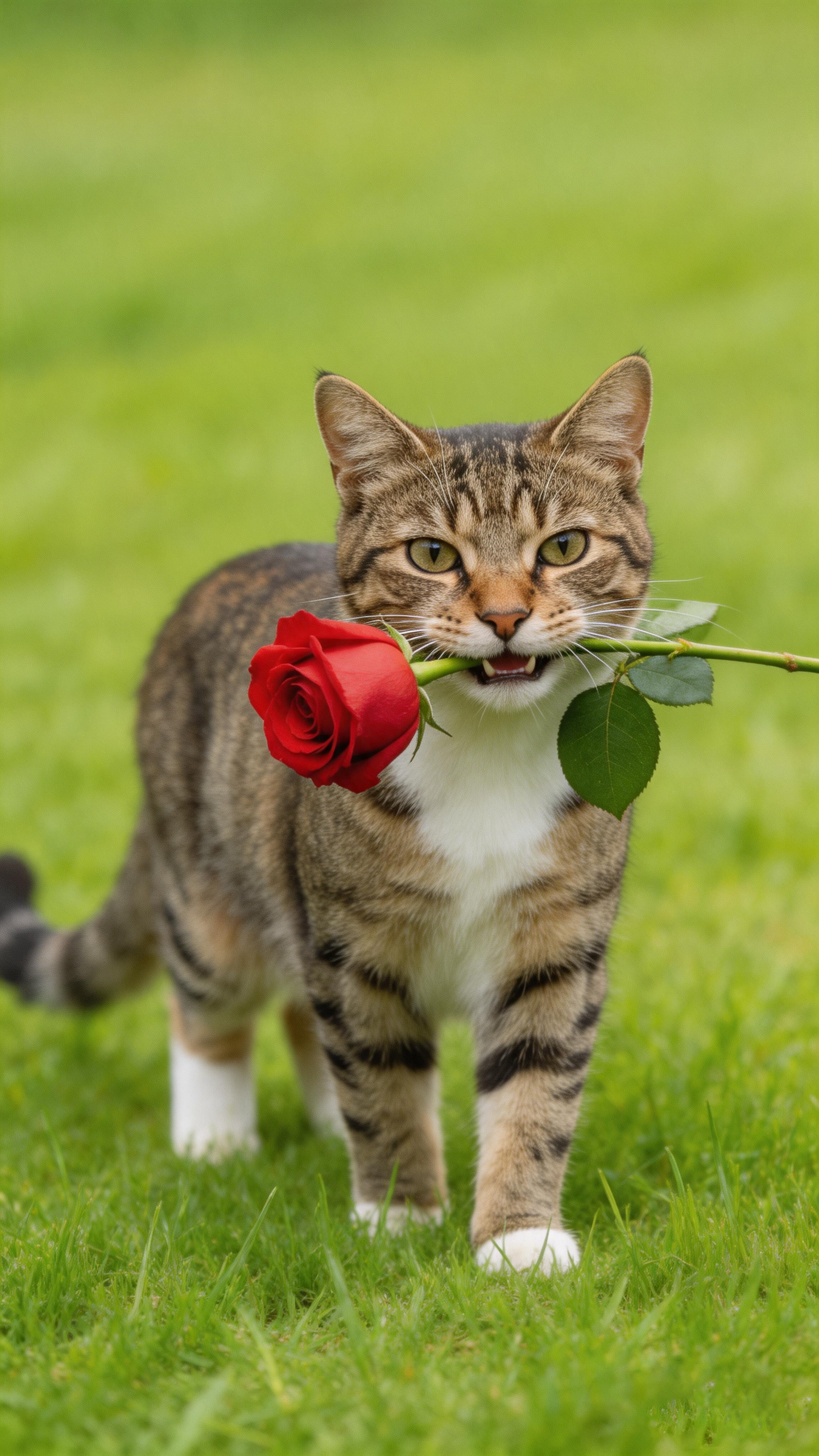} & \includegraphics[width=1.8cm]{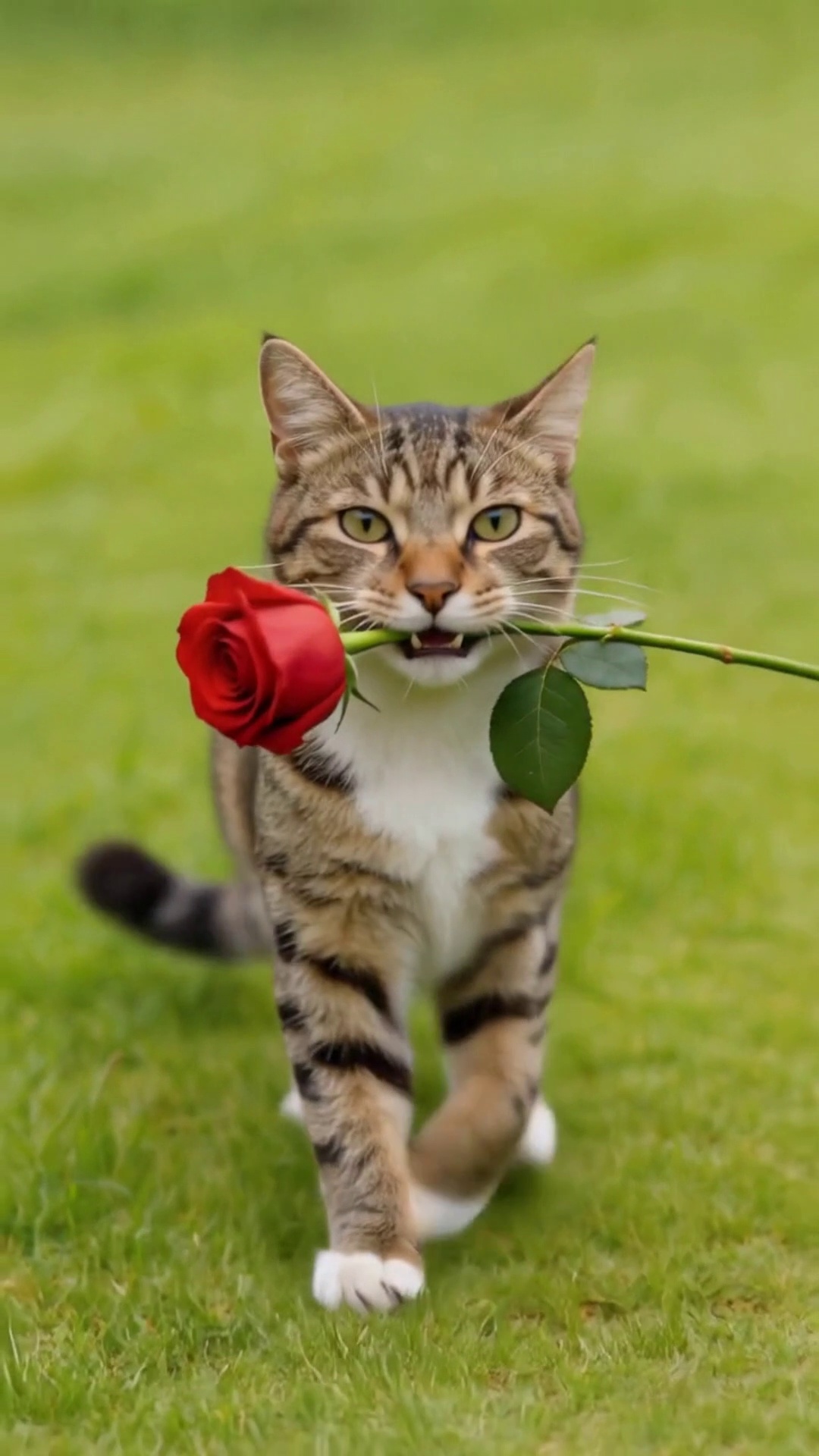} & \includegraphics[width=1.8cm]{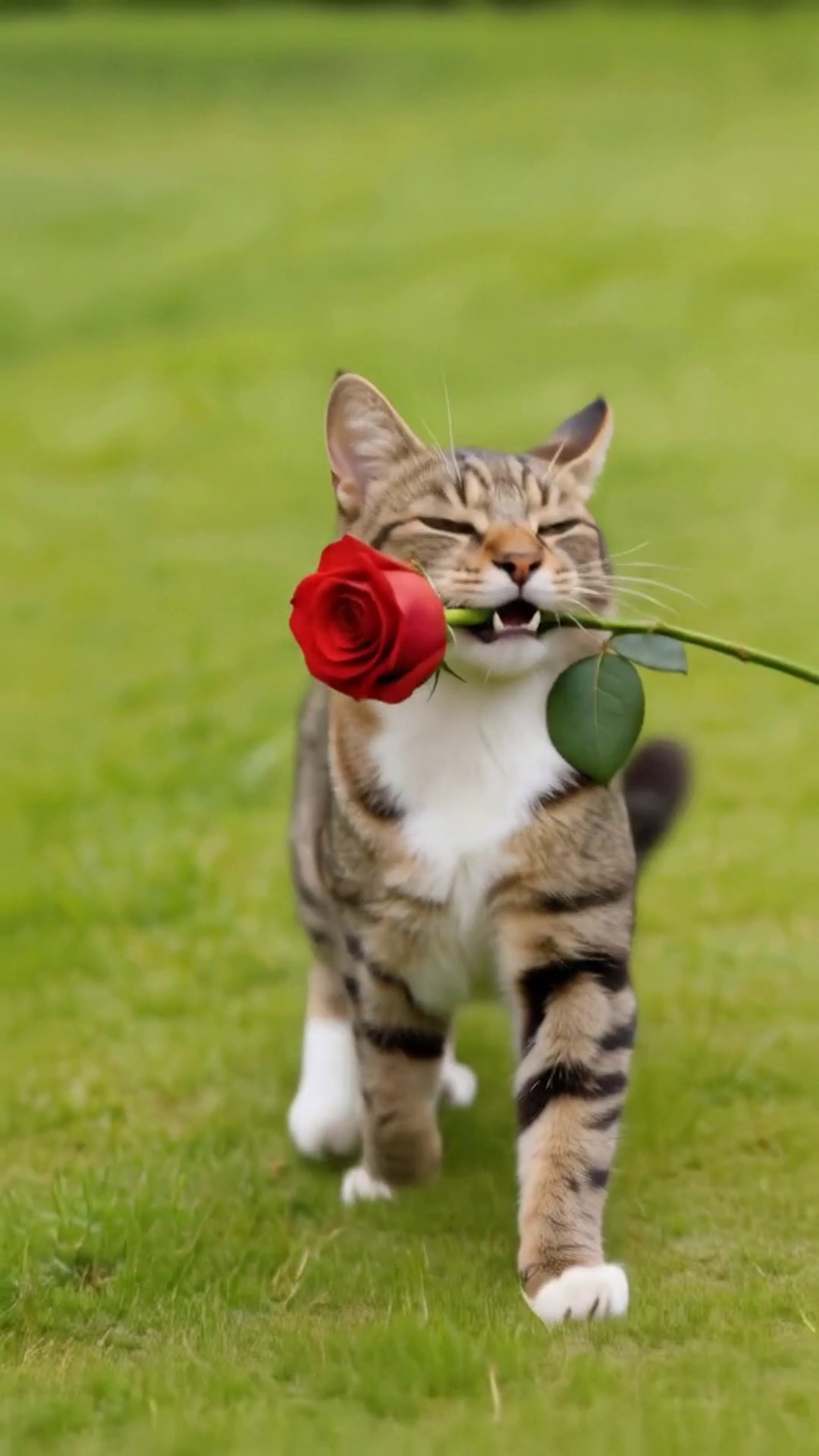} & \includegraphics[width=1.8cm]{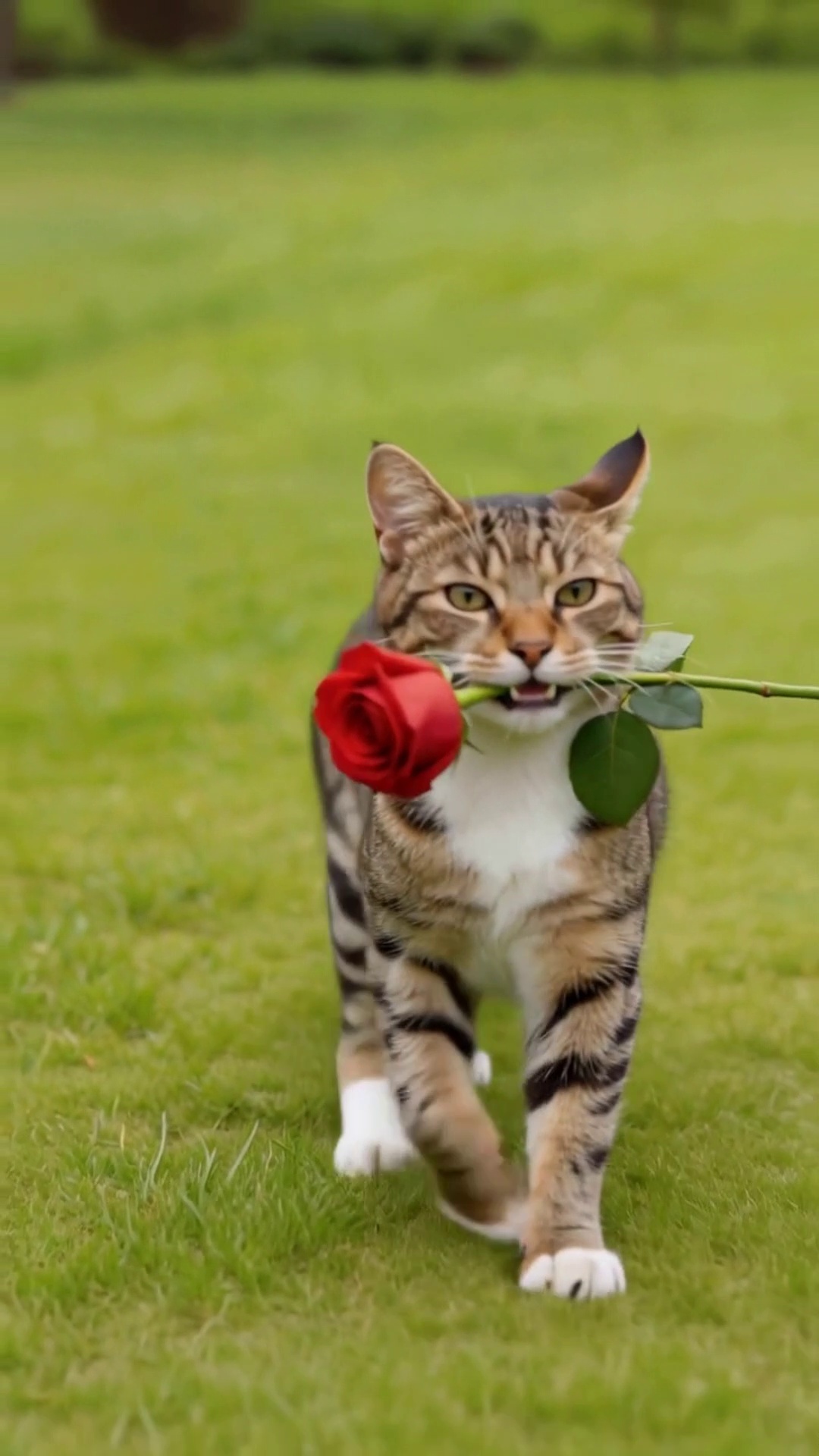} \\

    \end{tabular}
    \end{center}
    \caption{Animation results of \name. Our approach enables accurate cross-identity animation by faithfully transferring motion from the reference video to targets with drastically different body shapes and appearances, including humans, cartoon characters, robots, and animals.} 
    \label{fig:qual}
\end{figure*}

This section presents qualitative evaluations of \name across diverse and challenging character animation scenarios. As shown in Figure~\ref{fig:qual}, our framework robustly handles cross-identity transfer between subjects with drastically different body shapes and appearances, including humans, cartoon characters, robots, and animals. In particular, the results demonstrate faithful preservation of fine-grained dynamics such as subtle facial expressions, intricate hand movements, and complex non-rigid motions, while maintaining consistent character identity throughout the generated sequences.

\subsection{Viewpoint Control}

\begin{figure*}[t]
    \begin{center}
    \setlength{\tabcolsep}{0.5pt}
    \begin{tabular}{m{2cm}<{\centering}m{2cm}<{\centering}m{2cm}<{\centering}m{2cm}<{\centering}m{0.5cm}<{\centering}m{2cm}<{\centering}m{2cm}<{\centering}m{2cm}<{\centering}m{2cm}<{\centering}}

    \multicolumn{4}{c}{\scriptsize{Reference Video}} & & \multicolumn{4}{c}{\scriptsize{Reference Video}} \\
    \includegraphics[width=1.95cm]{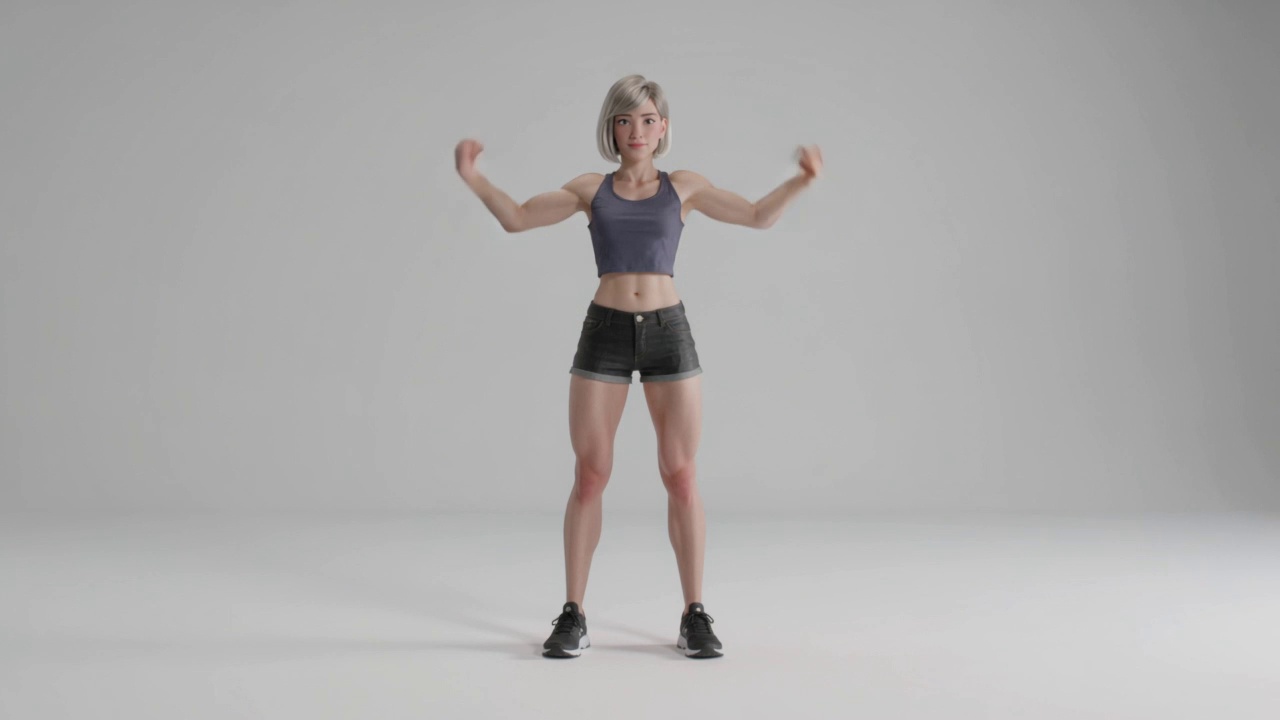} & \includegraphics[width=1.95cm]{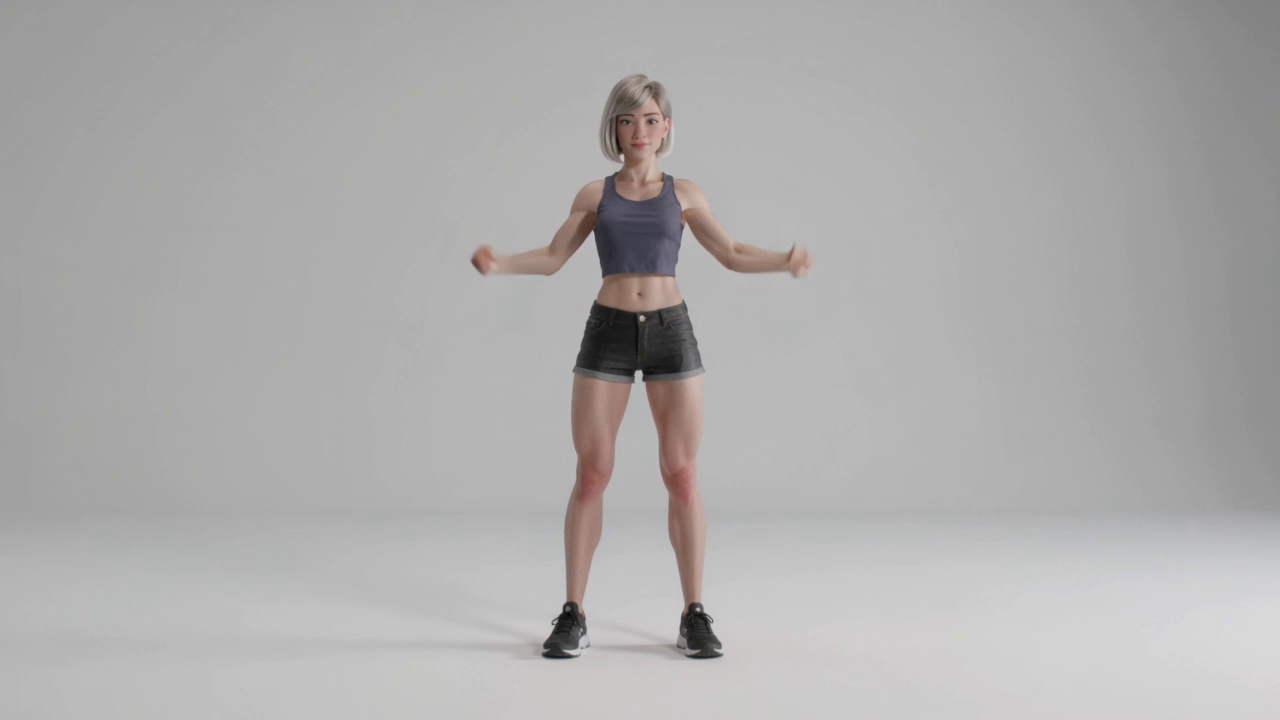} & \includegraphics[width=1.95cm]{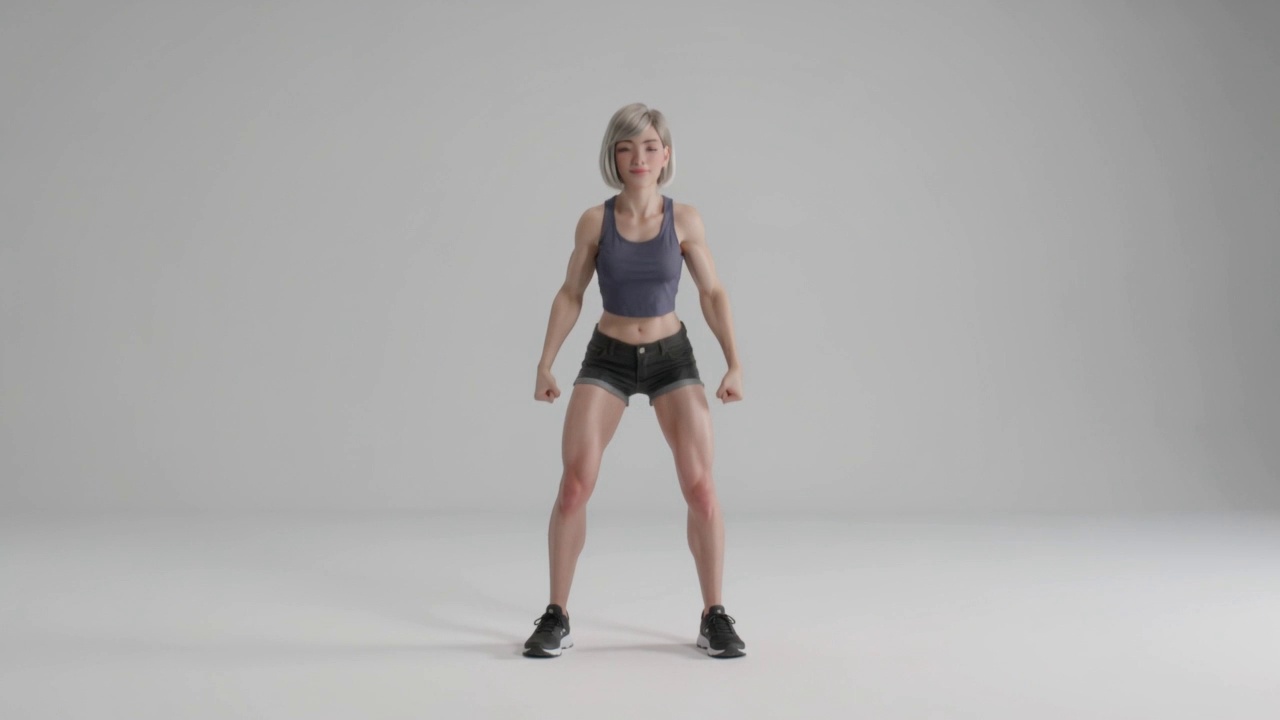} & \includegraphics[width=1.95cm]{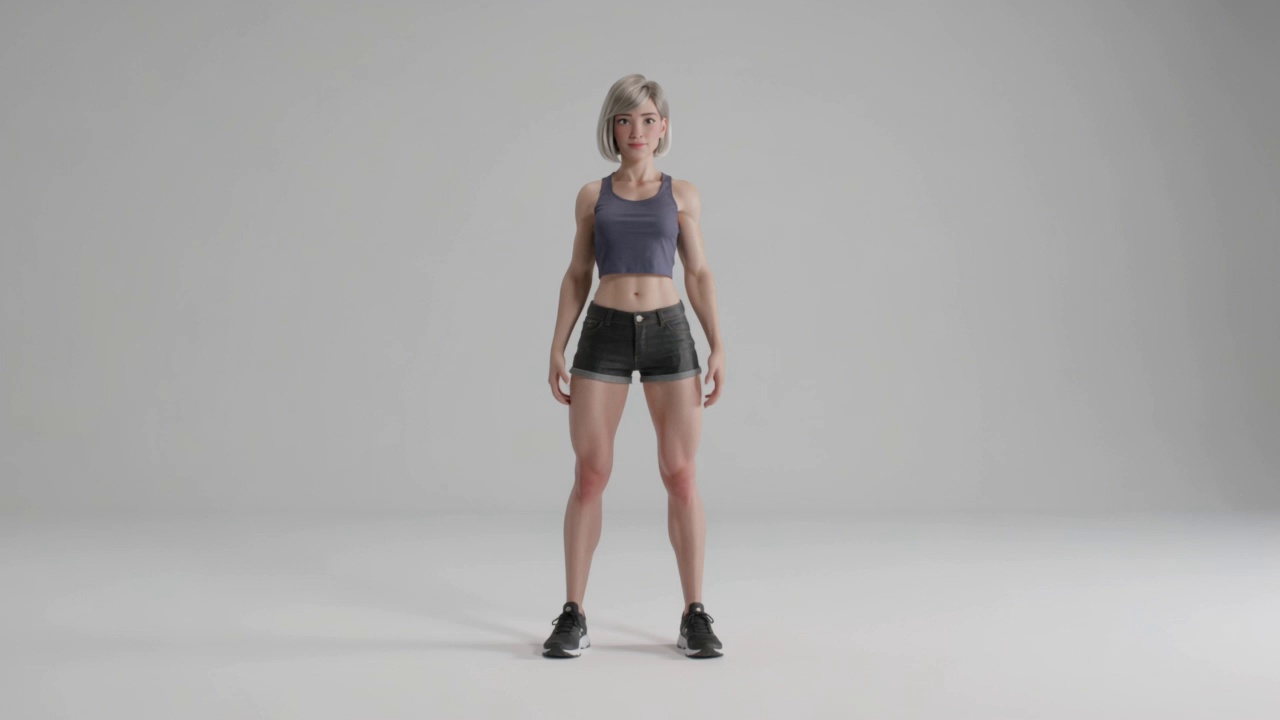} & & \includegraphics[width=1.95cm]{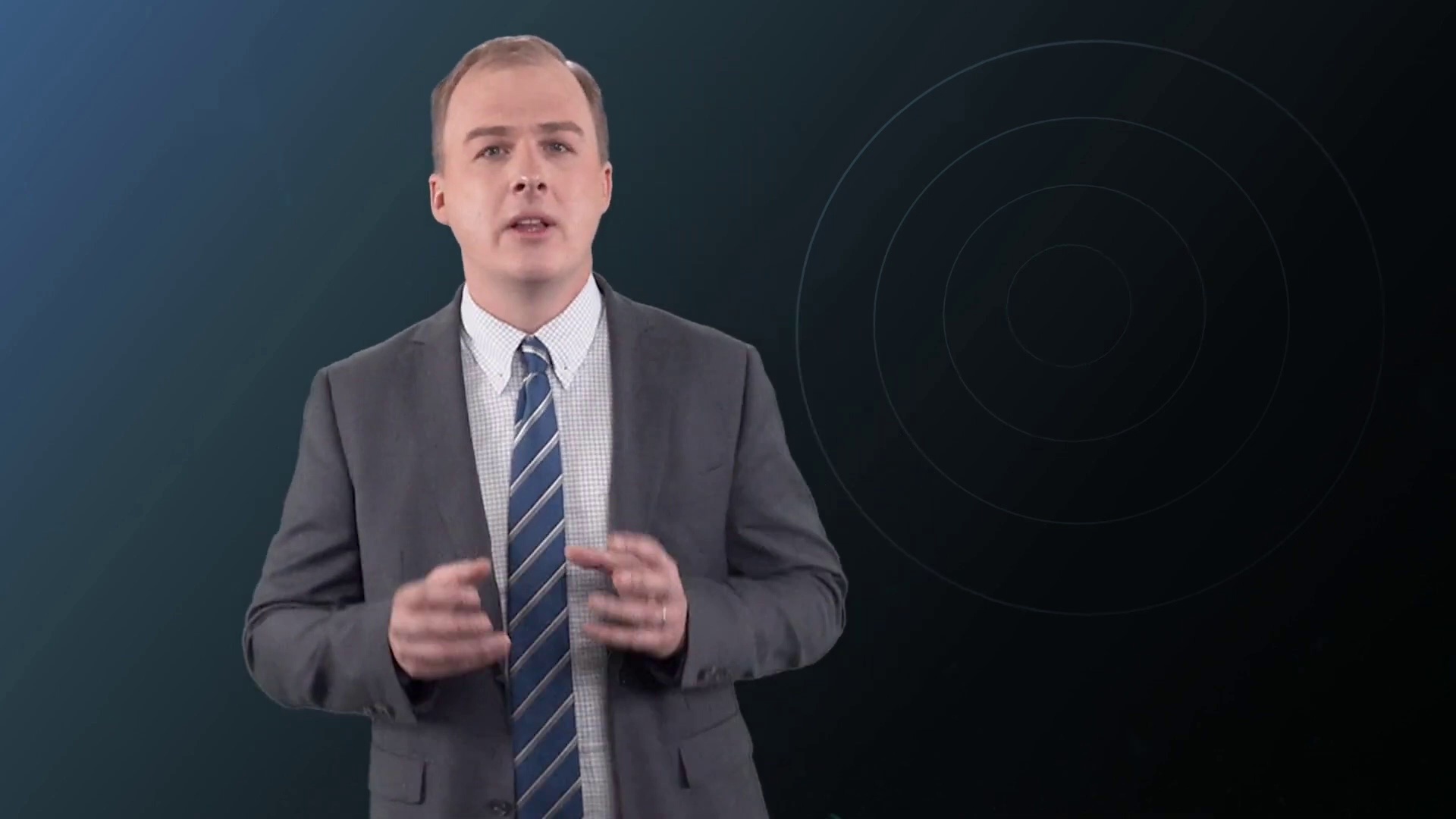} & \includegraphics[width=1.95cm]{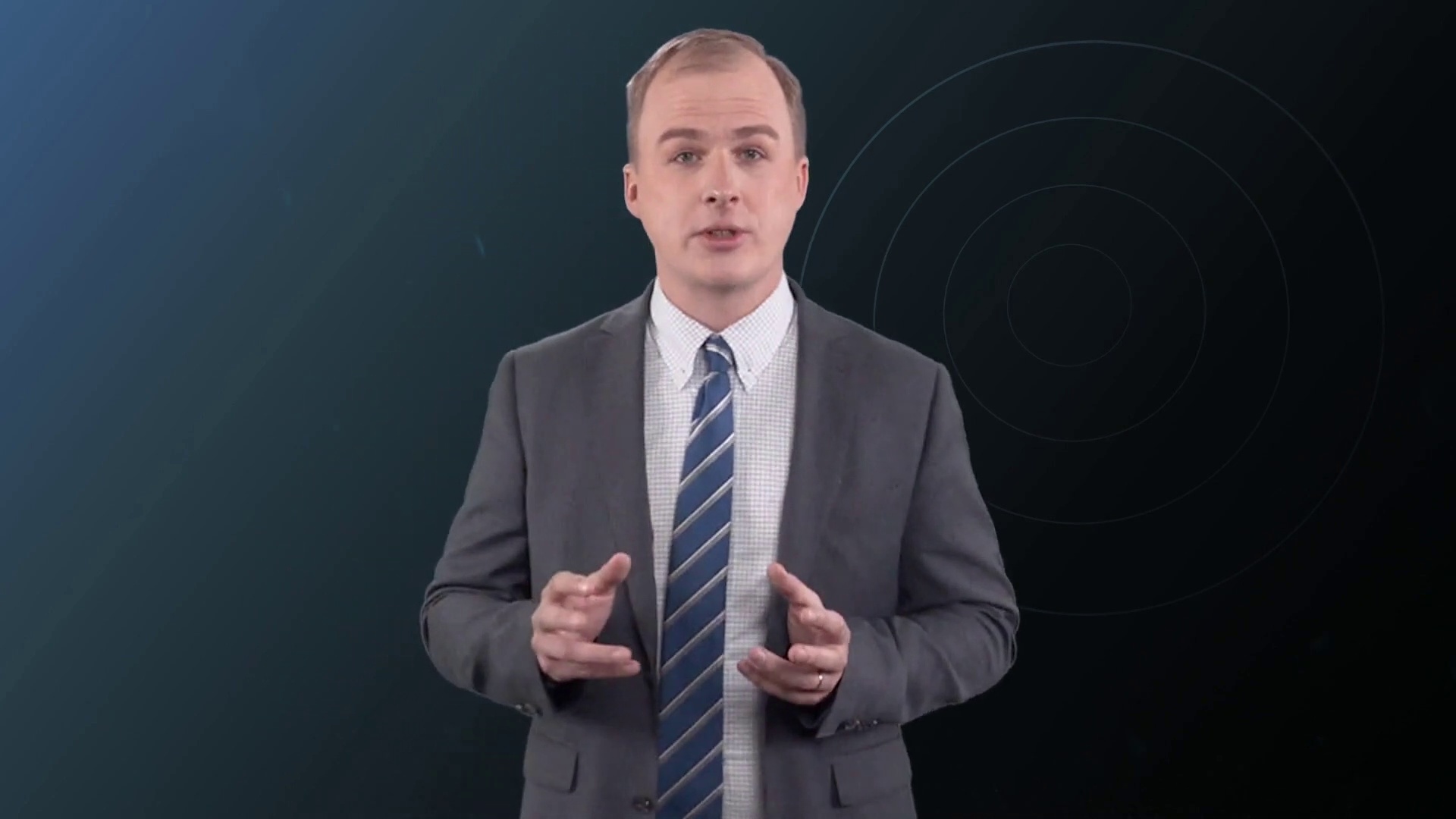} & \includegraphics[width=1.95cm]{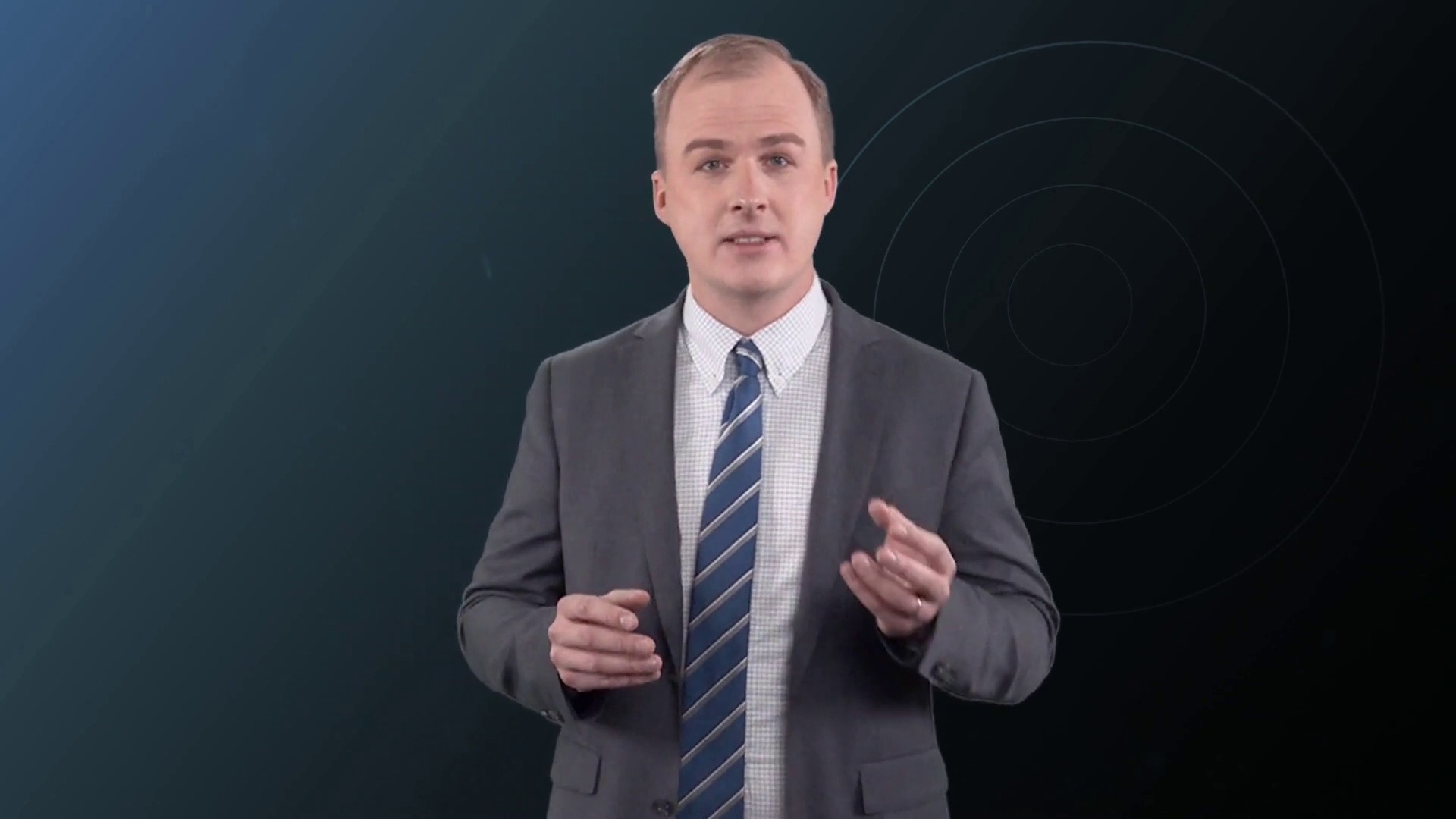} & \includegraphics[width=1.95cm]{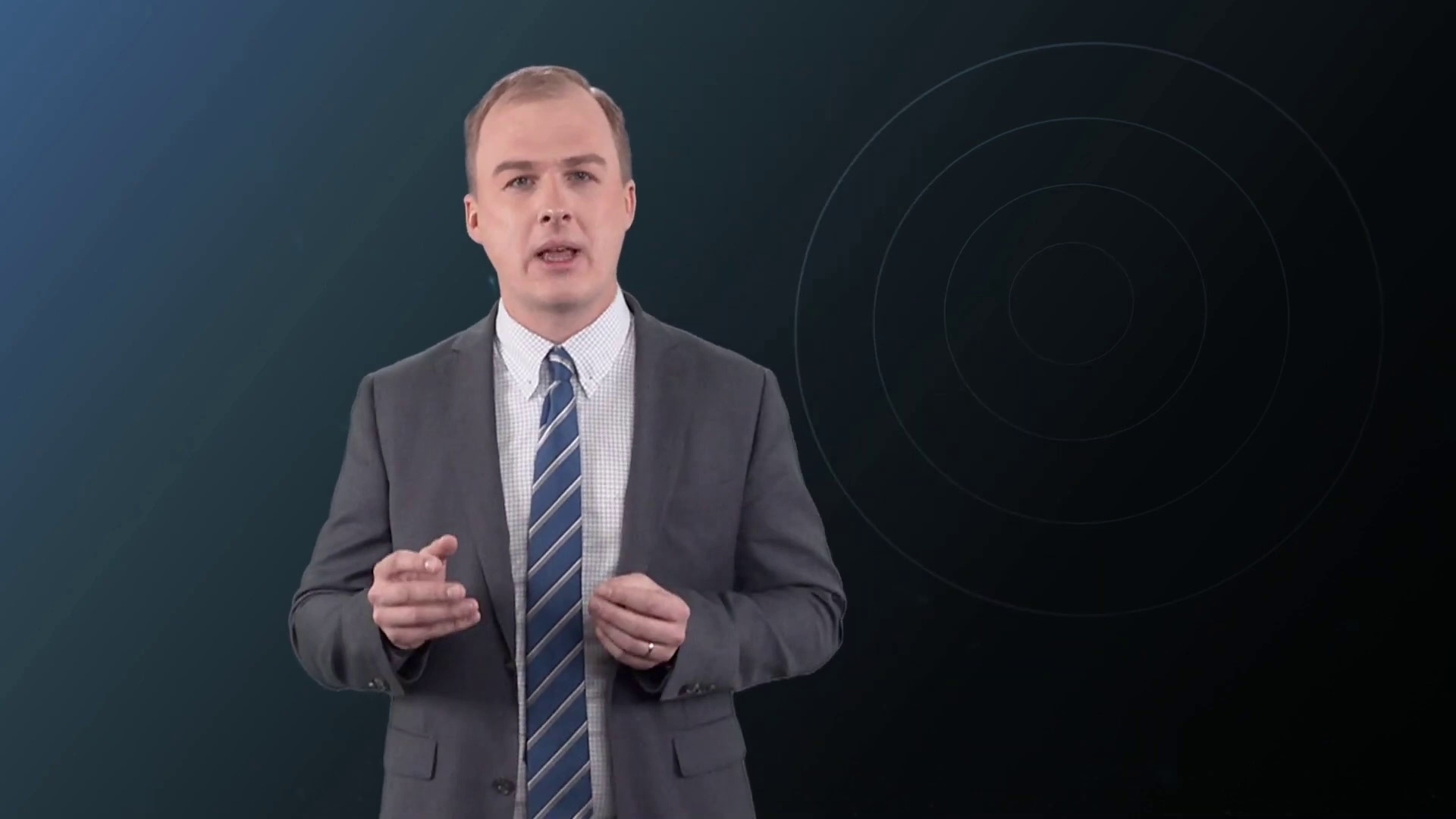} \\
    
    \multicolumn{4}{c}{\scriptsize{Left 30-degree View, Bottom View}} & & \multicolumn{4}{c}{\scriptsize{Left 30-degree View, Top View}} \\
    \includegraphics[width=1.95cm]{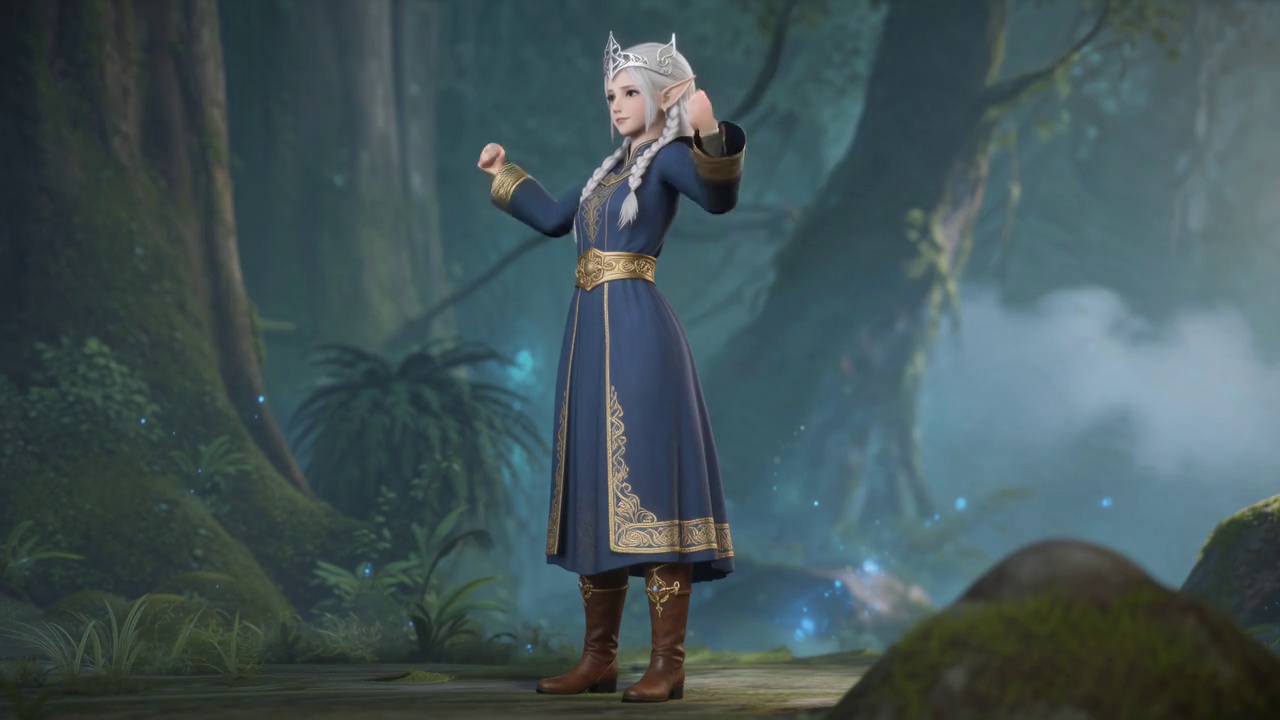} & \includegraphics[width=1.95cm]{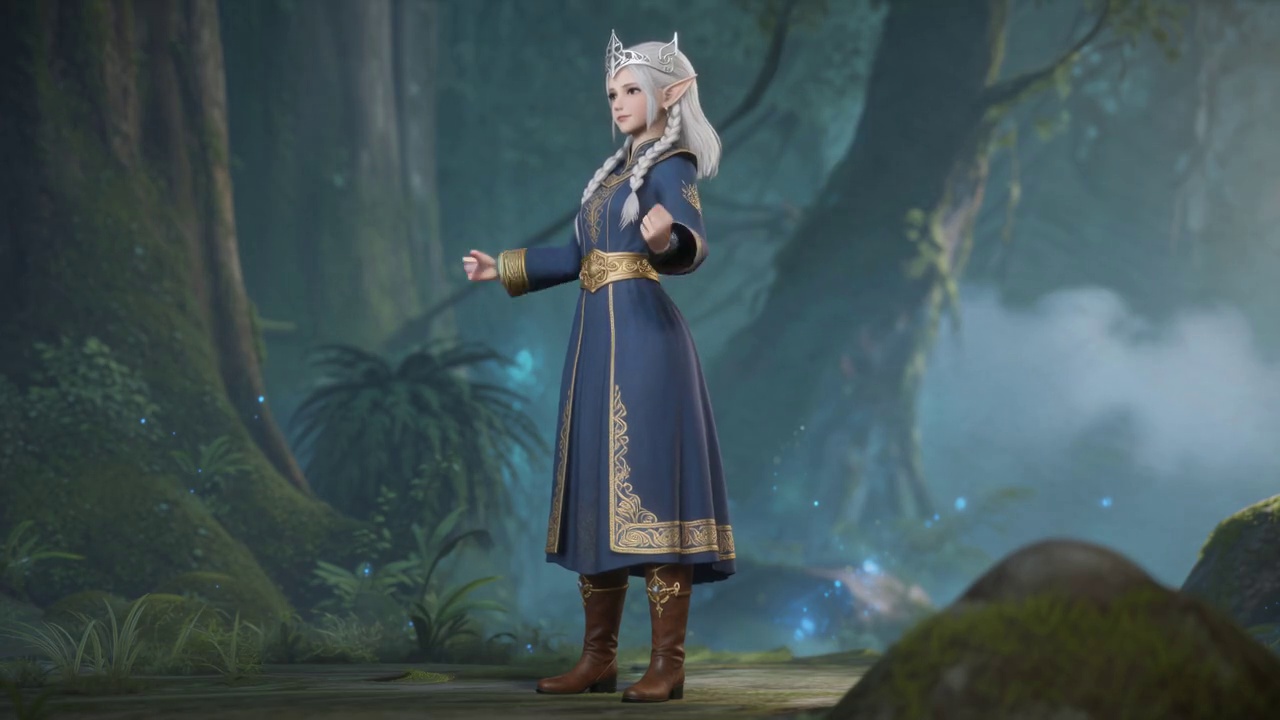} & \includegraphics[width=1.95cm]{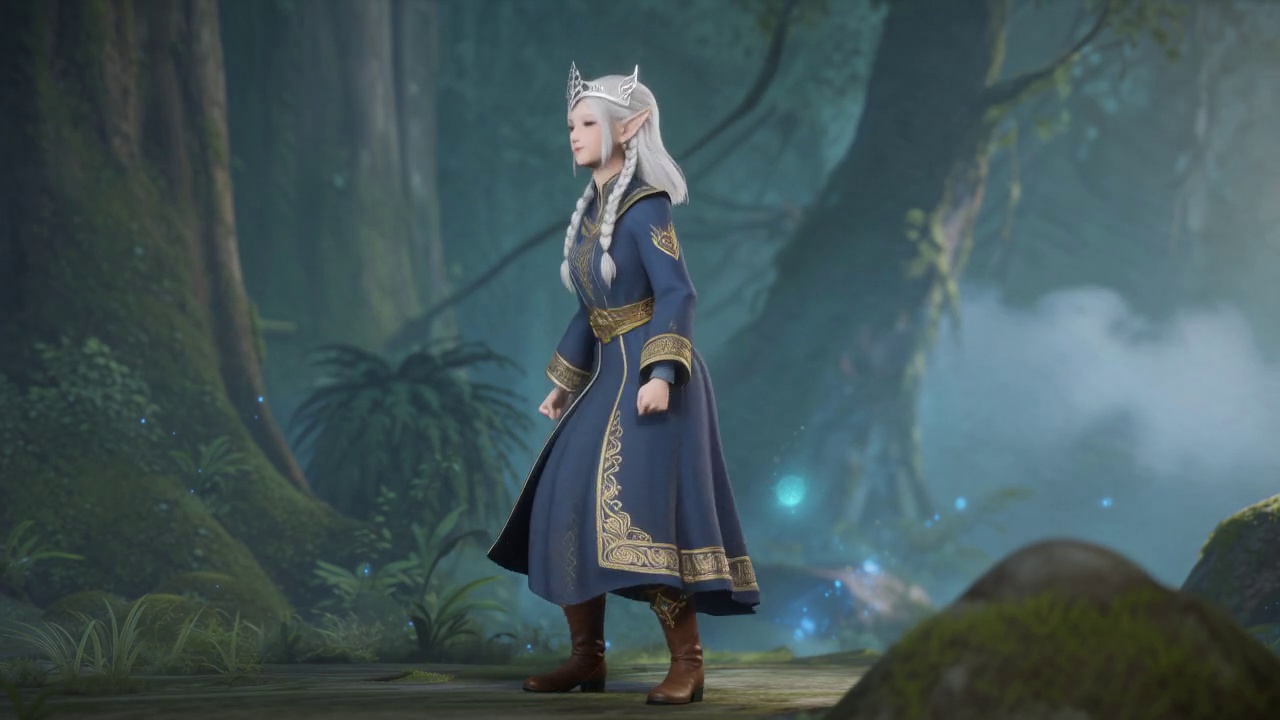} & \includegraphics[width=1.95cm]{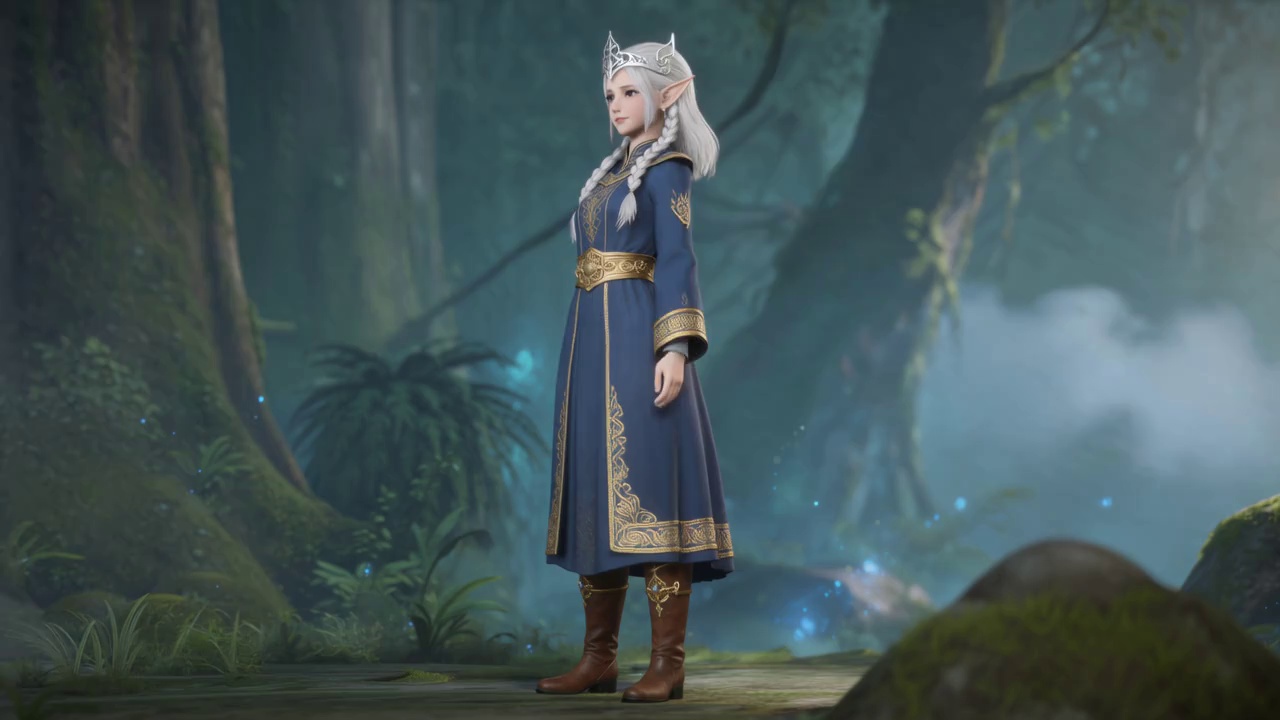} & & \includegraphics[width=1.95cm]{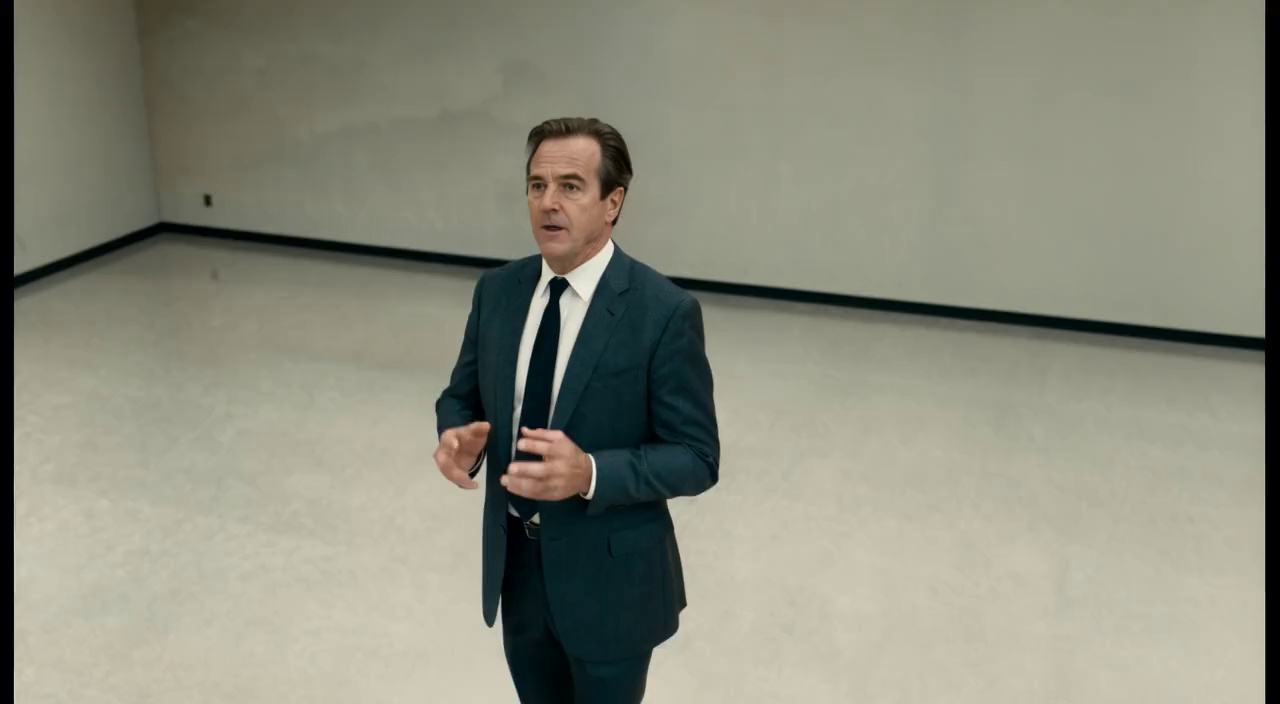} & \includegraphics[width=1.95cm]{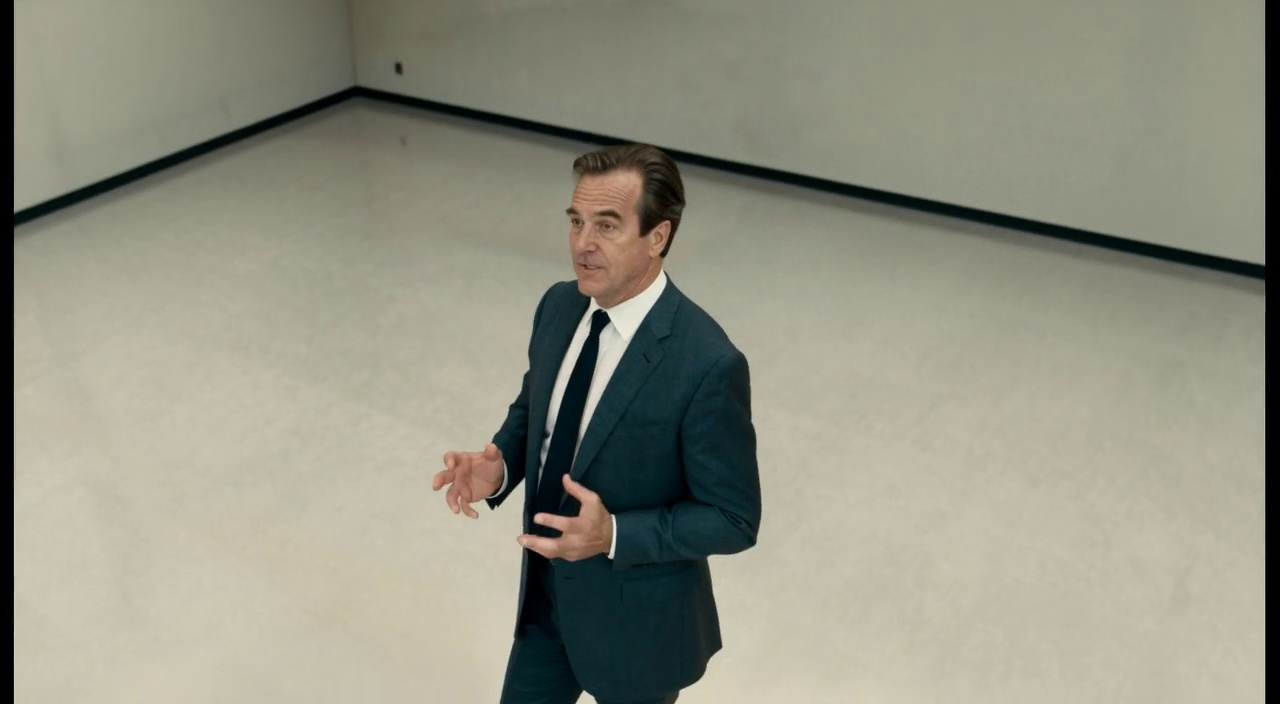} & \includegraphics[width=1.95cm]{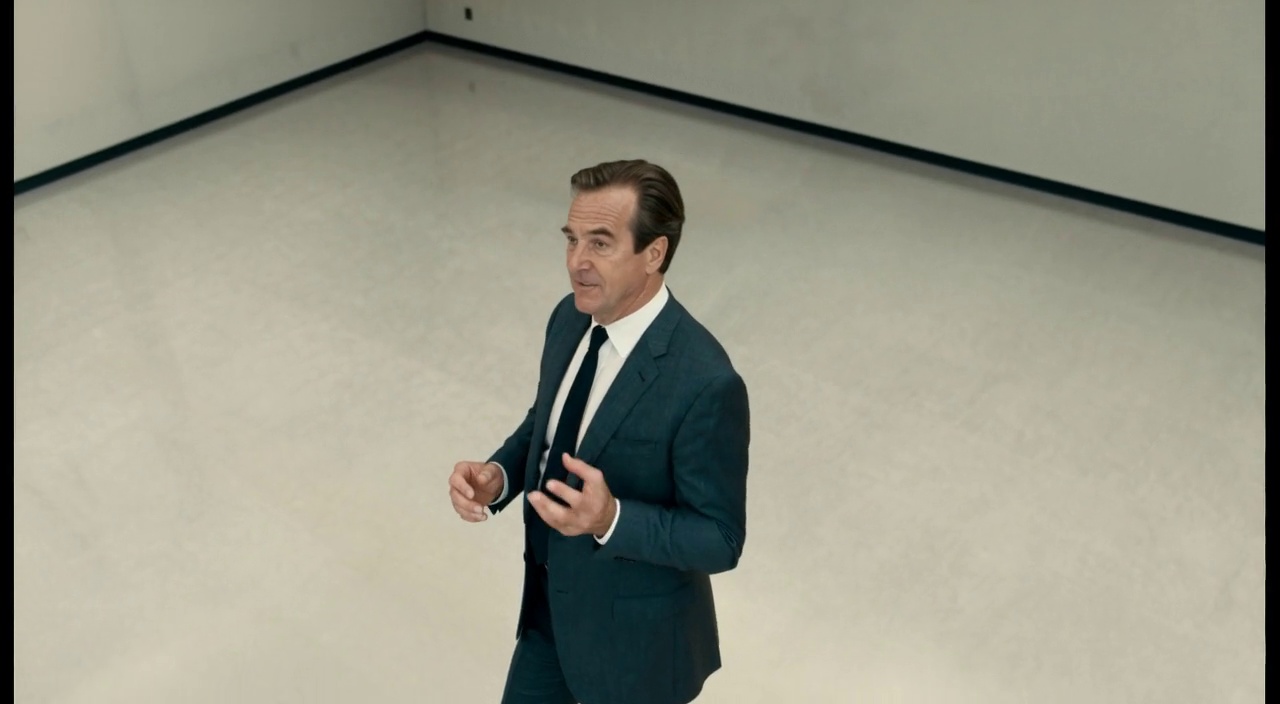} & \includegraphics[width=1.95cm]{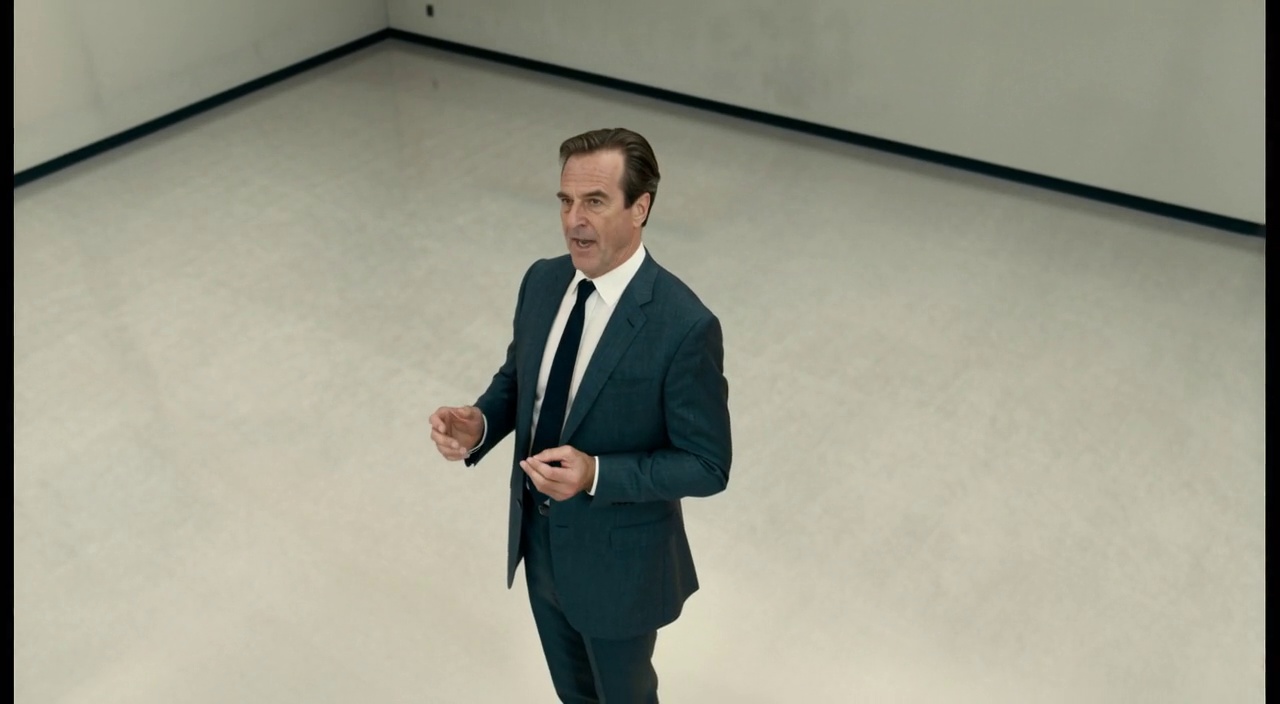} \\

    \multicolumn{4}{c}{\scriptsize{Right 60-degree View, Eye Level}} & & \multicolumn{4}{c}{\scriptsize{Front View, Eye Level}} \\
    \includegraphics[width=1.95cm]{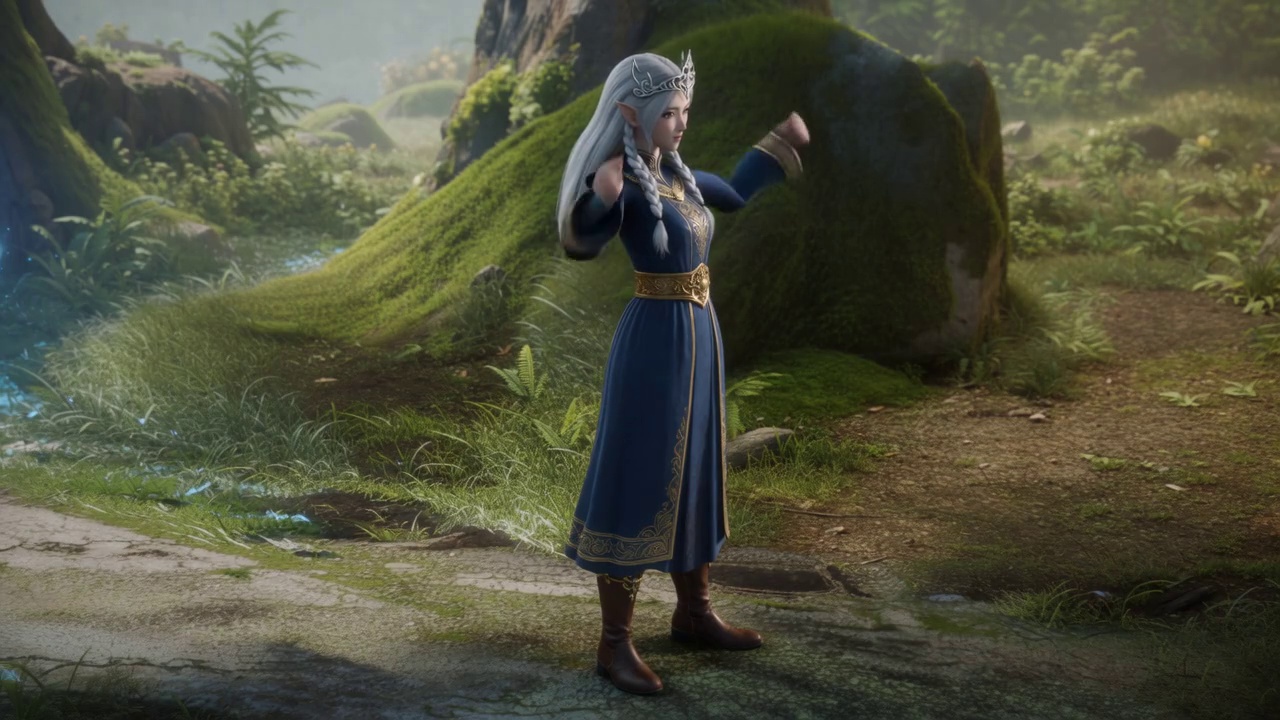} & \includegraphics[width=1.95cm]{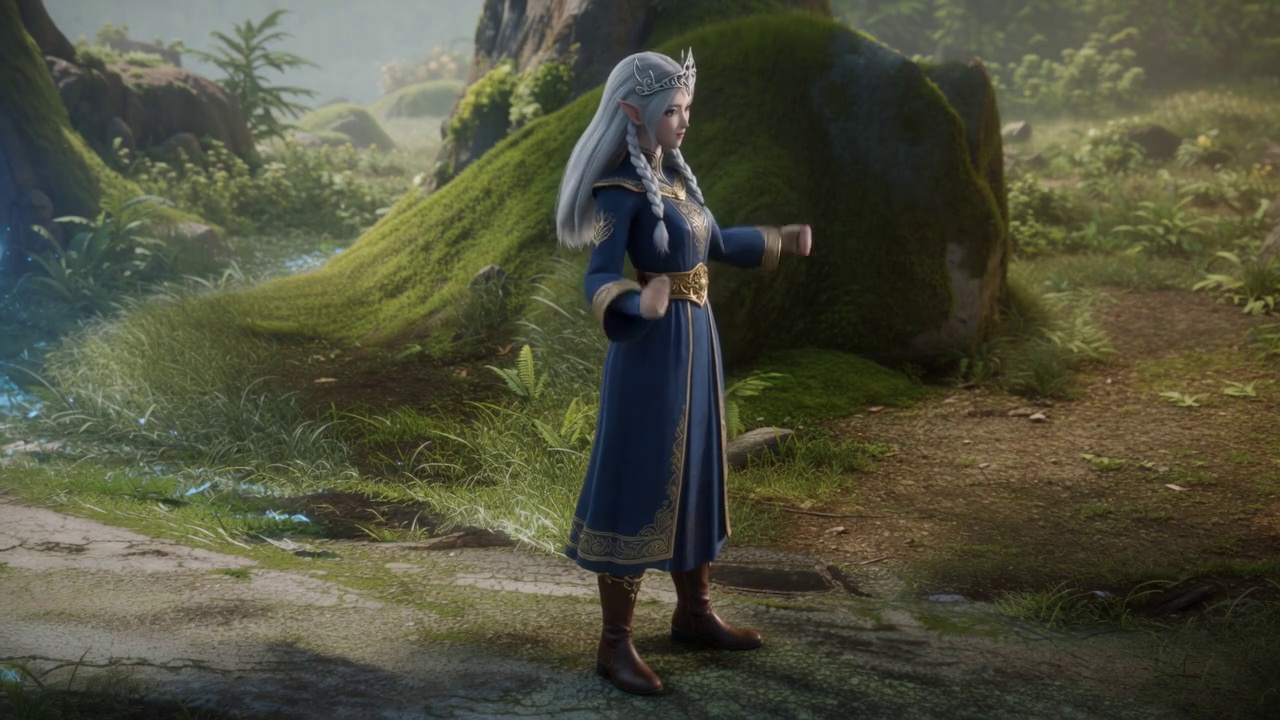} & \includegraphics[width=1.95cm]{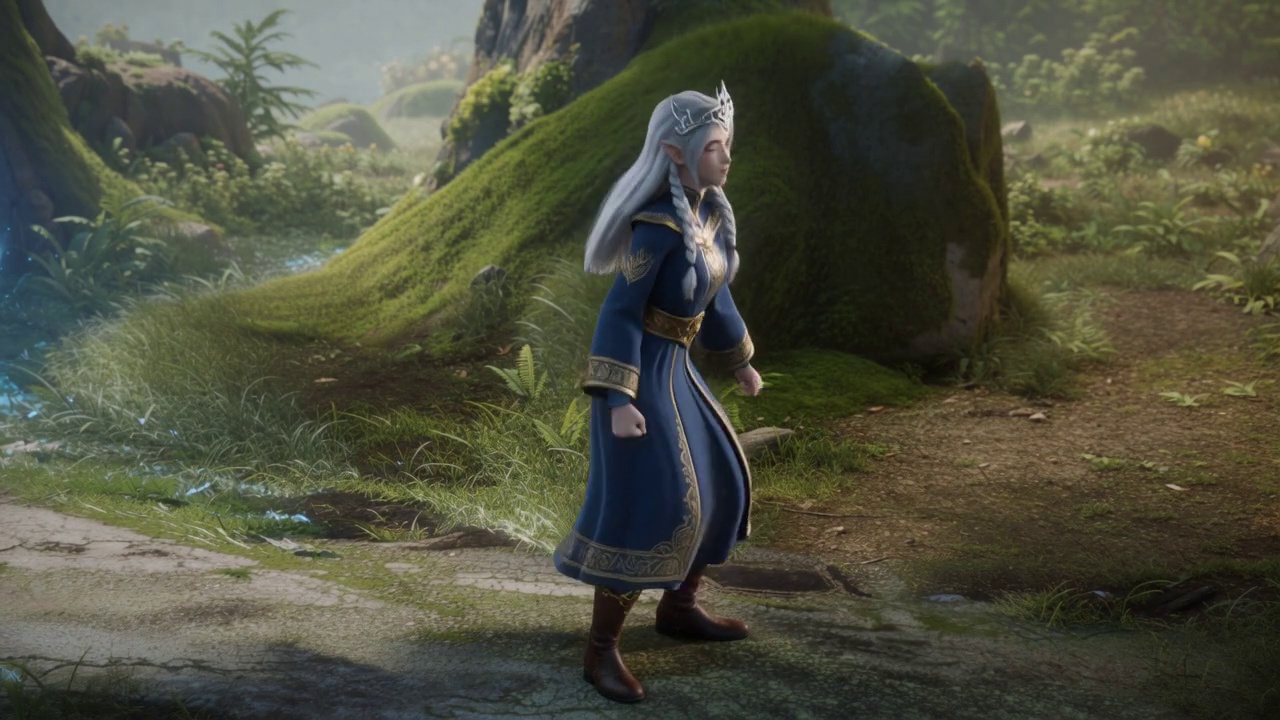} & \includegraphics[width=1.95cm]{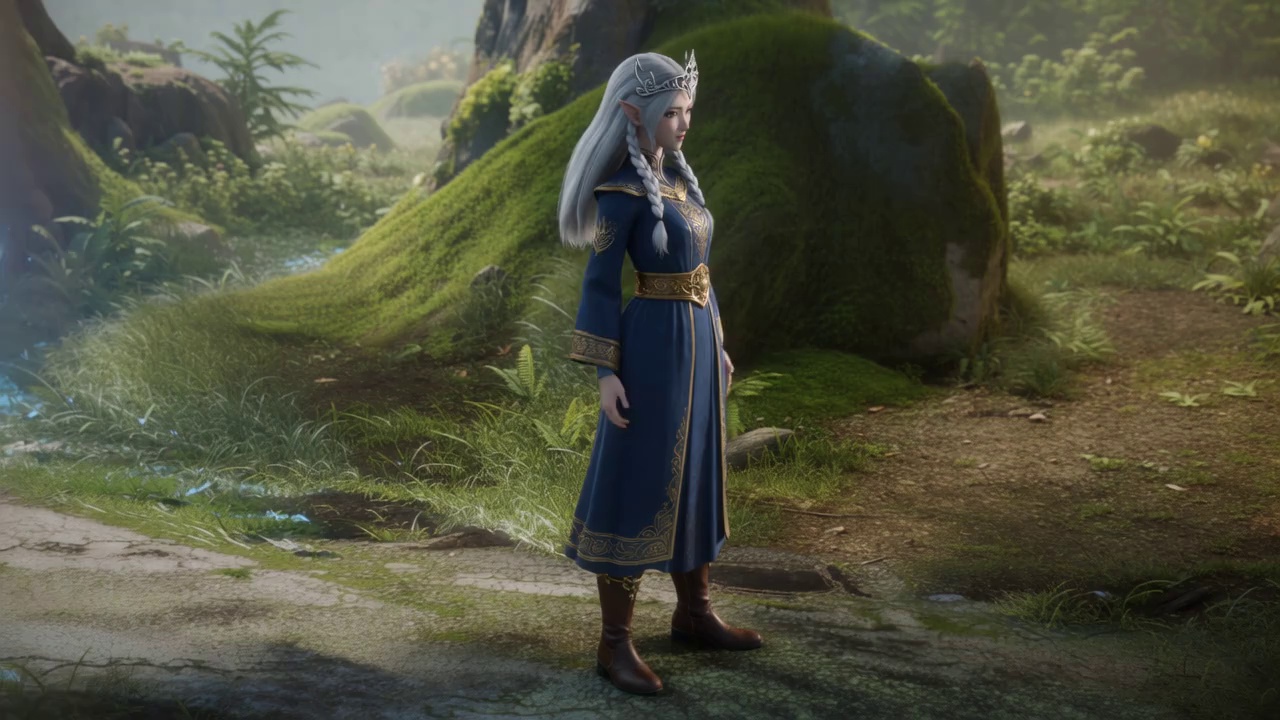} & & \includegraphics[width=1.95cm]{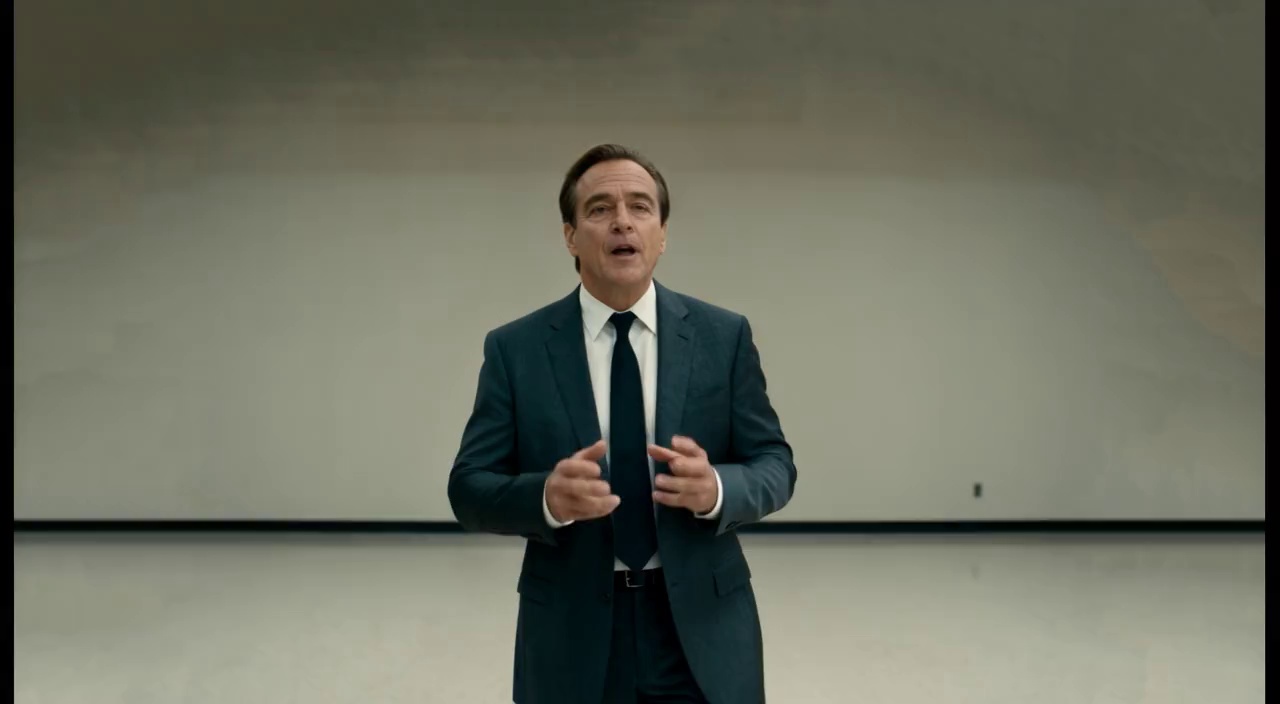} & \includegraphics[width=1.95cm]{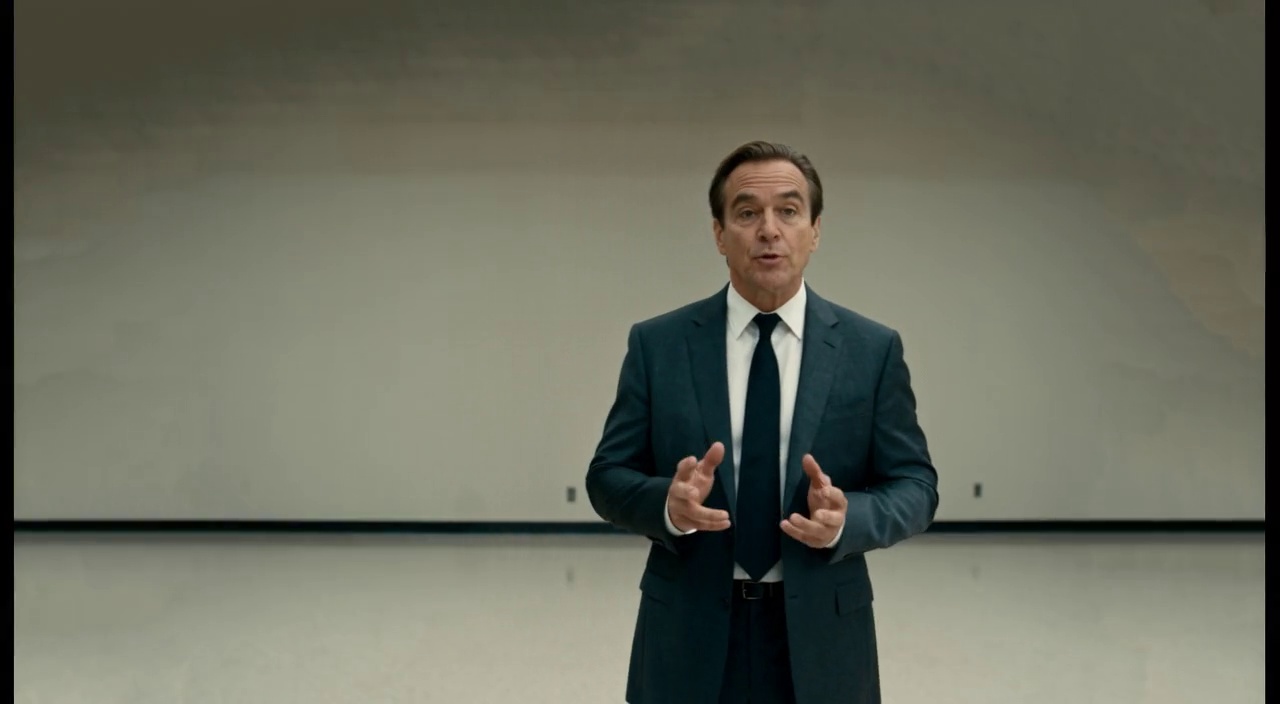} & \includegraphics[width=1.95cm]{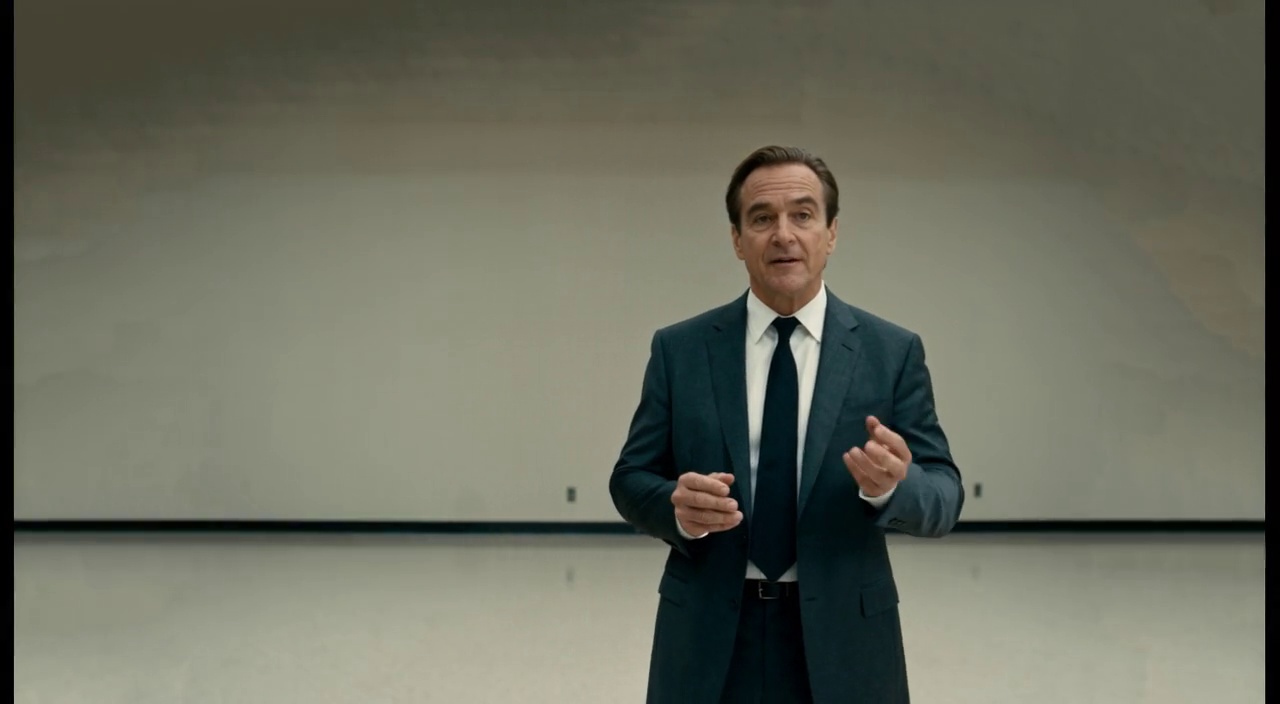} & \includegraphics[width=1.95cm]{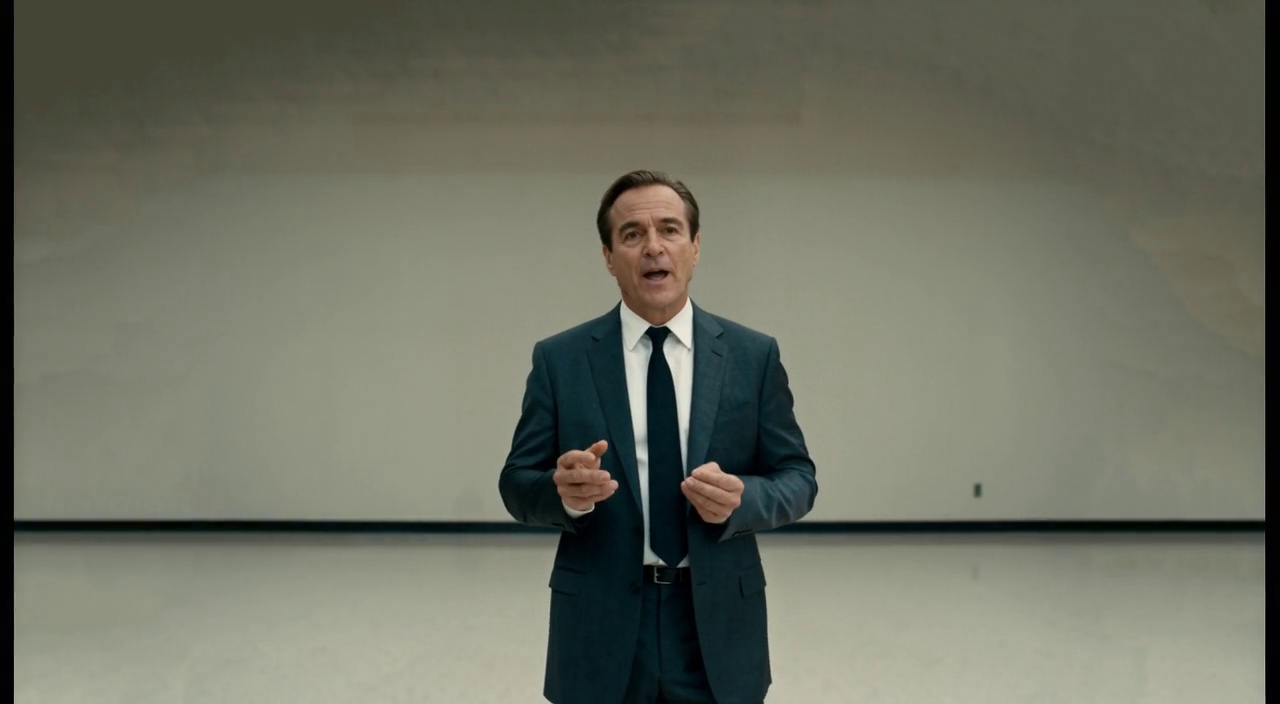} \\

    \multicolumn{4}{c}{\scriptsize{Left 30-degree View, Top View}} & & \multicolumn{4}{c}{\scriptsize{Right 60-degree View, Top View}} \\
    \includegraphics[width=1.95cm]{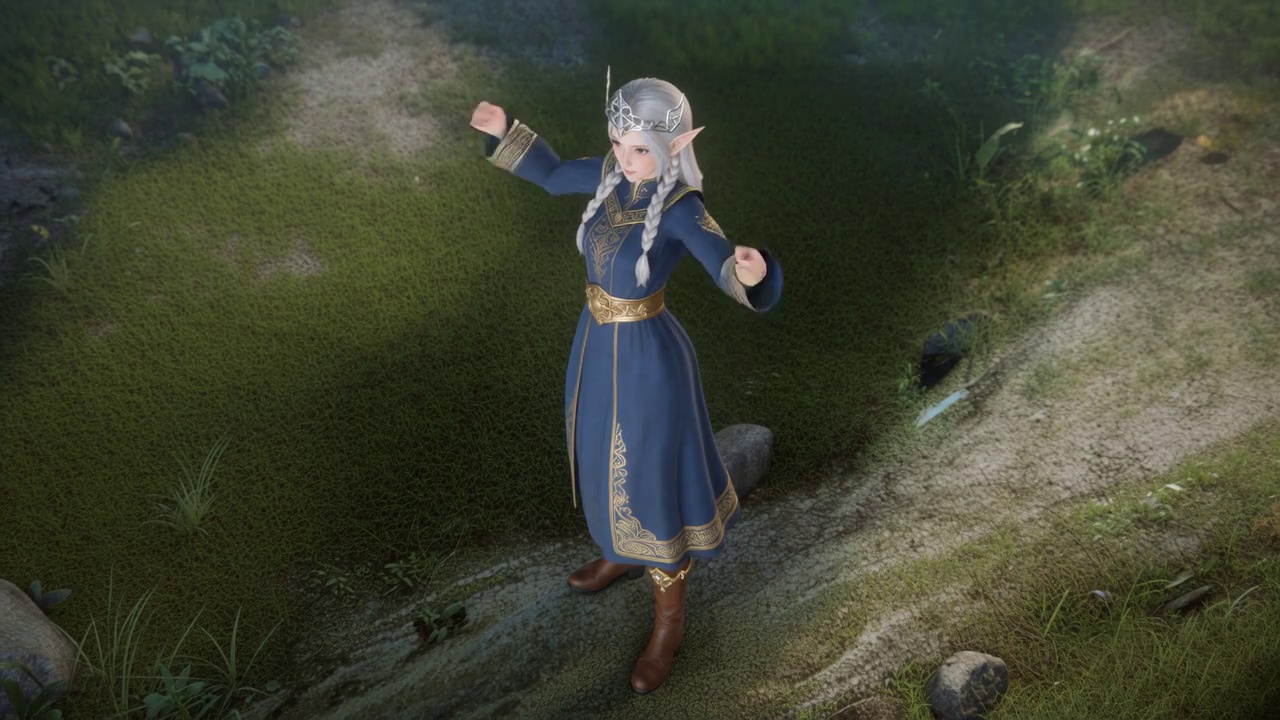} & \includegraphics[width=1.95cm]{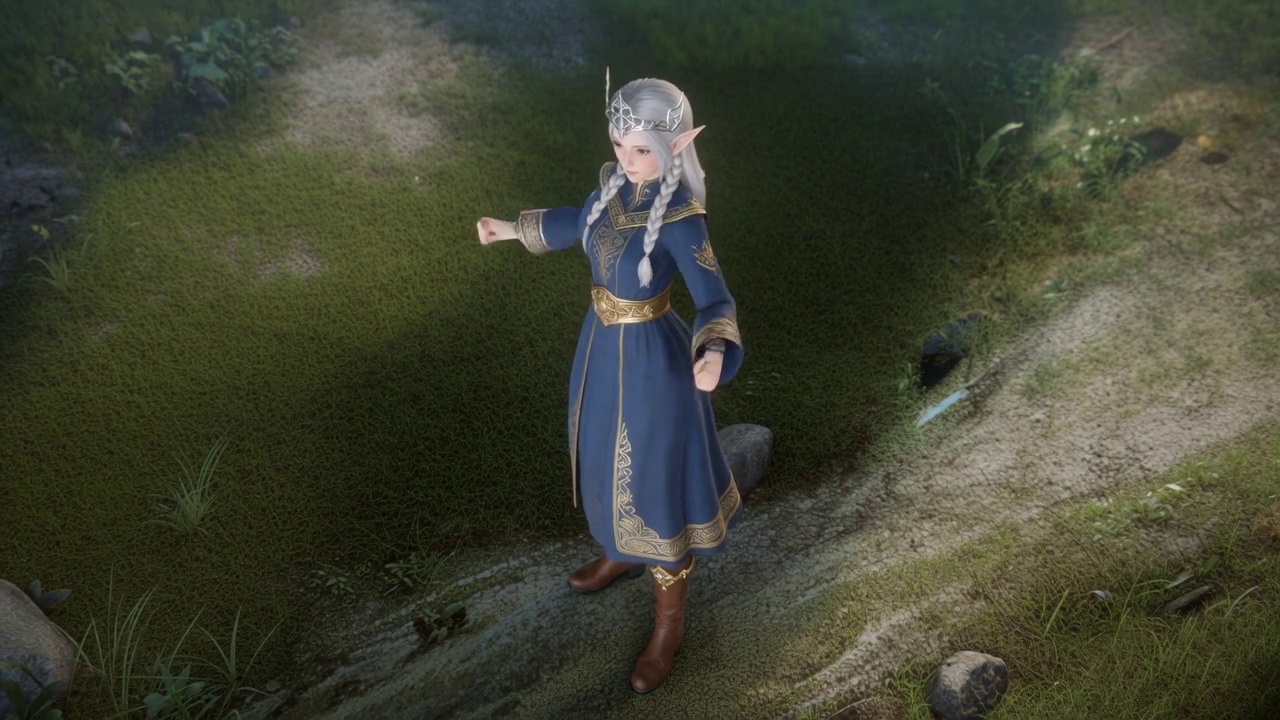} & \includegraphics[width=1.95cm]{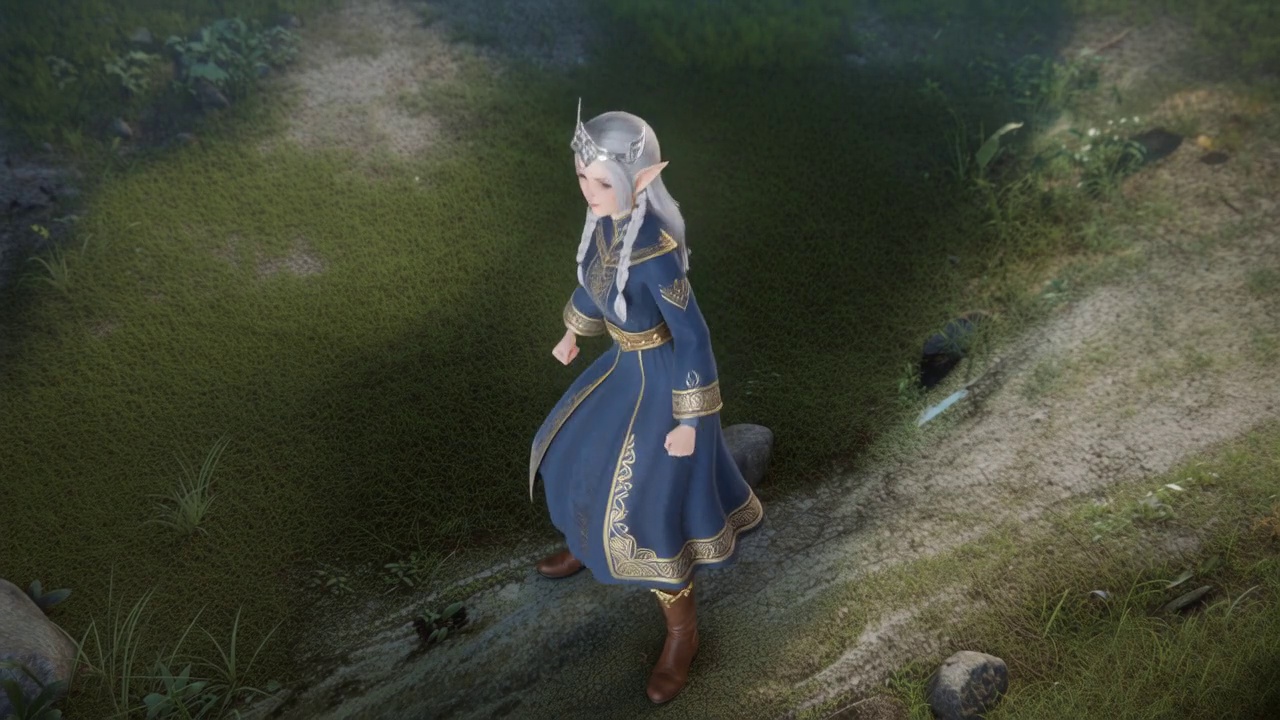} & \includegraphics[width=1.95cm]{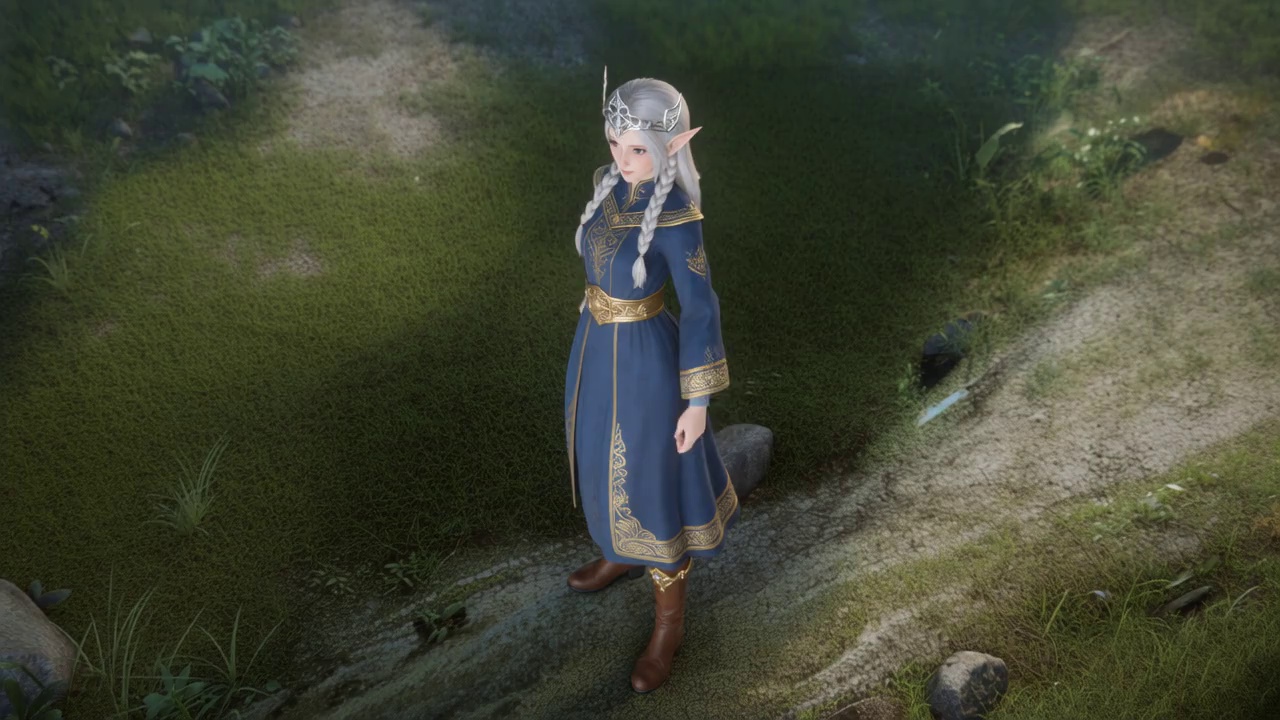} & & \includegraphics[width=1.95cm]{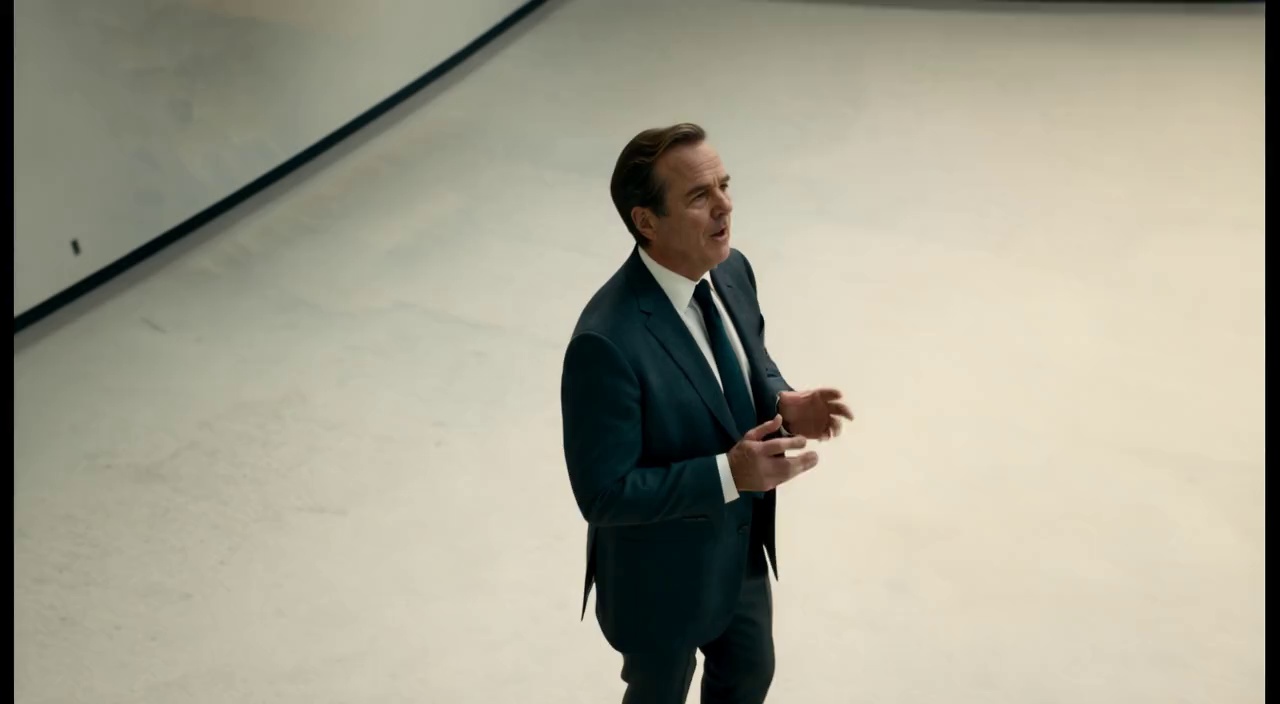} & \includegraphics[width=1.95cm]{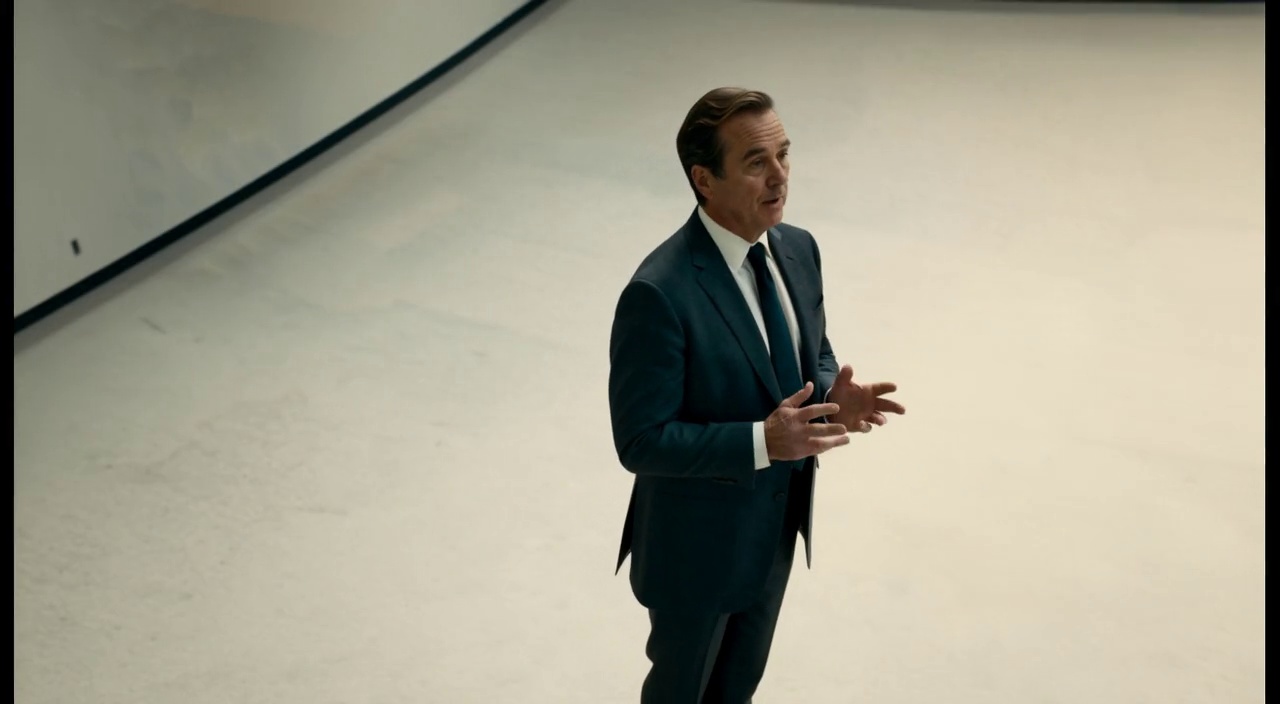} & \includegraphics[width=1.95cm]{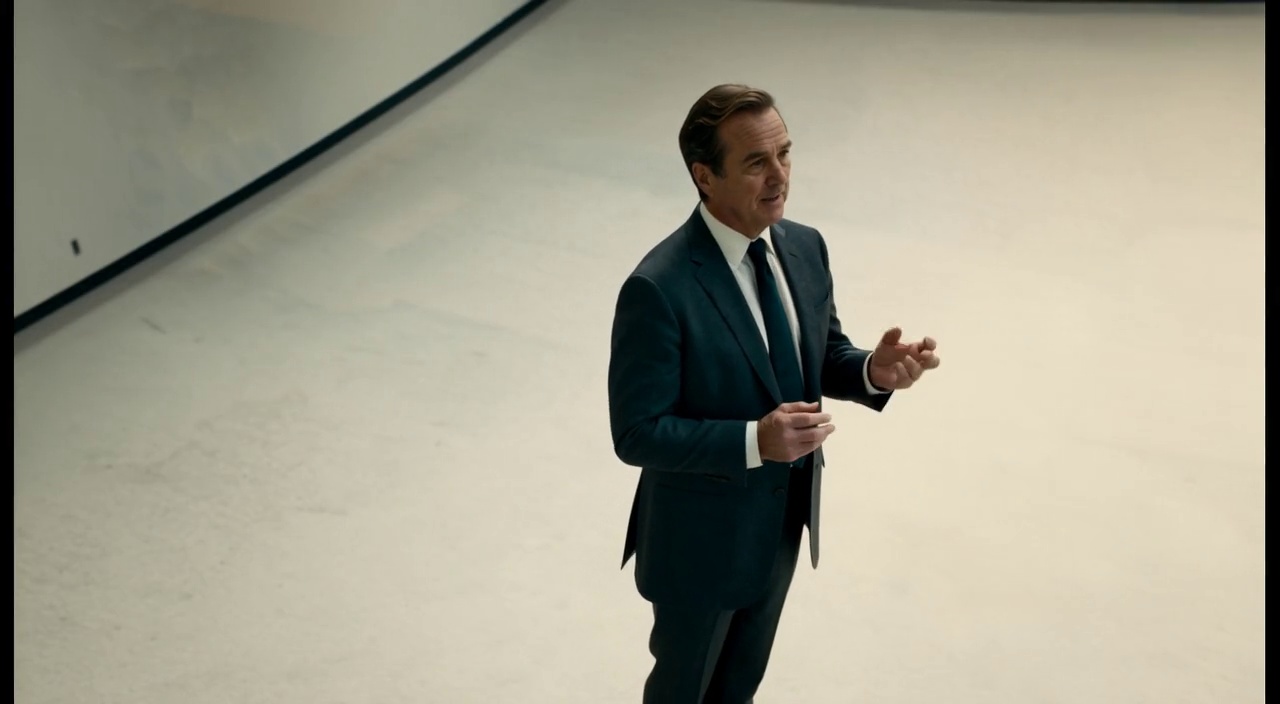} & \includegraphics[width=1.95cm]{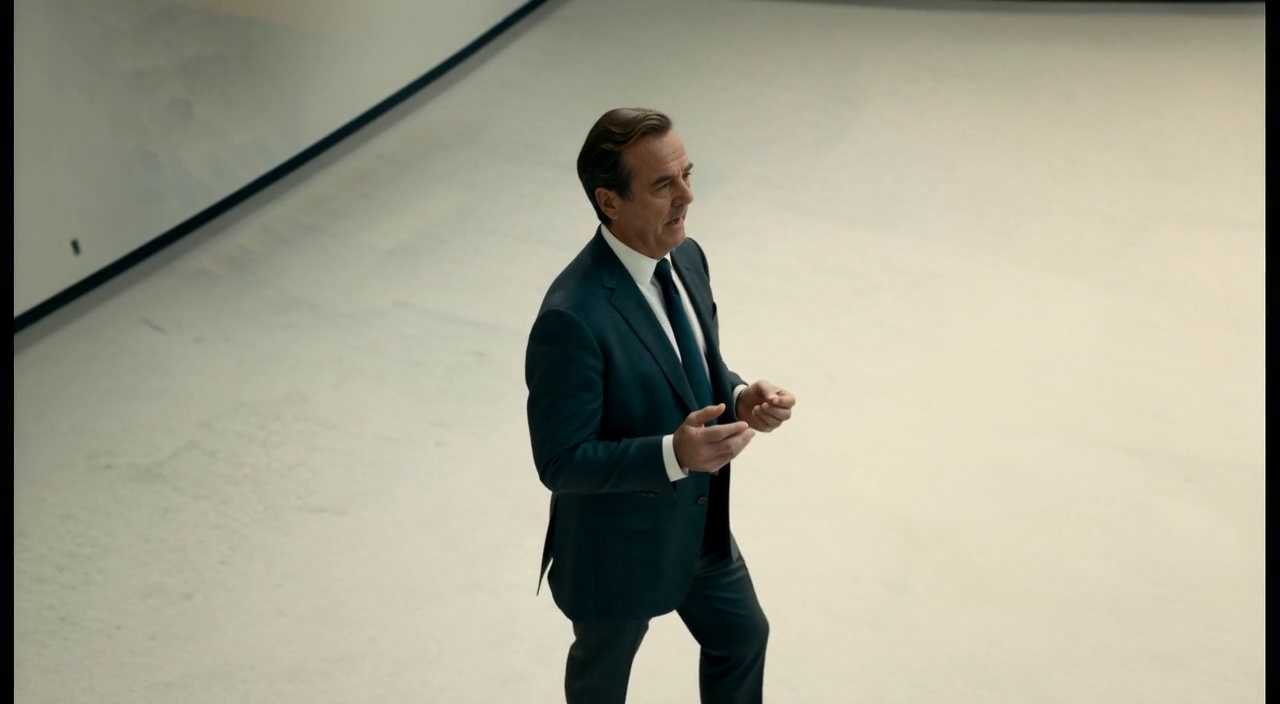} \\

    \end{tabular}
    \end{center}
    \caption{Text-guided viewpoint control results generated by \name. Our method not only animates the character according to the reference video, but also enables control of the camera viewpoint through text prompts, producing consistent animations from diverse views such as left/right side views, top views, and eye-level front views.} 
    \label{fig:viewpoint}
    
\end{figure*}

In this section, we evaluate the performance of our Viewpoint LoRA regarding camera viewpoint control. We randomly selected several viewpoint descriptions to include in our prompts and tested them on real-world in-the-wild reference images and videos, with results shown in Figure~\ref{fig:viewpoint}. The experimental results demonstrate that our Viewpoint LoRA effectively decouples camera position from character motion dynamics. Furthermore, it maintains high consistency in the surrounding environment even when the camera viewpoint is significantly altered, confirming the robust spatial control enabled by our approach.

\subsection{Comparison with State-of-the-Art}

\begin{figure*}[htbp]
    \begin{center}
    \setlength{\tabcolsep}{0.5pt}
    \begin{tabular}{m{1.75cm}<{\centering}m{1.75cm}<{\centering}m{1.75cm}<{\centering}m{1.75cm}<{\centering}m{1.75cm}<{\centering}m{1.75cm}<{\centering}m{1.75cm}<{\centering}m{1.75cm}<{\centering}m{1.75cm}<{\centering}}

    \shortstack[b]{\scriptsize Reference\\[-0.1em]\scriptsize Image \\ [-0.15em] \includegraphics[width=1.7cm]{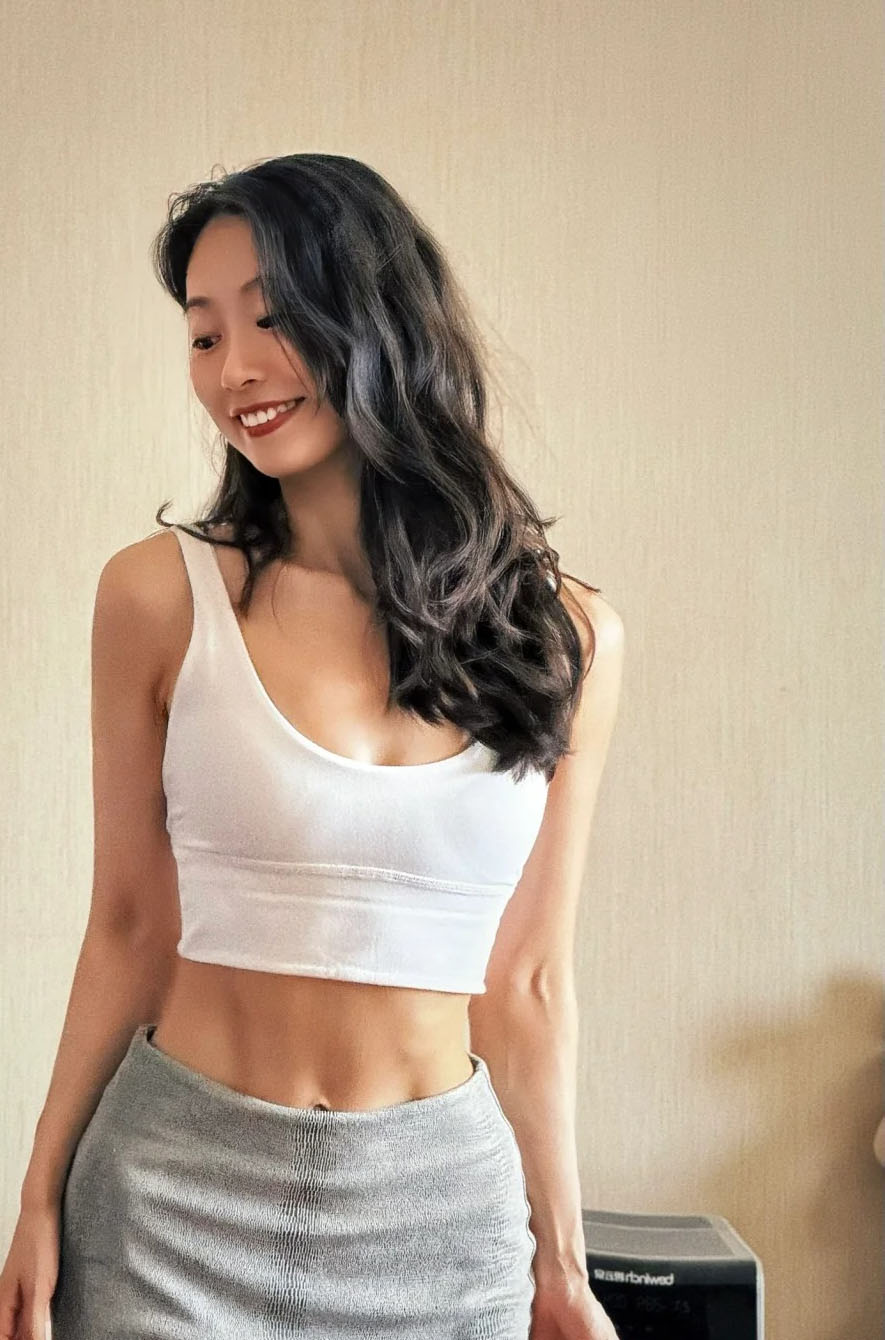}} & \includegraphics[width=1.7cm]{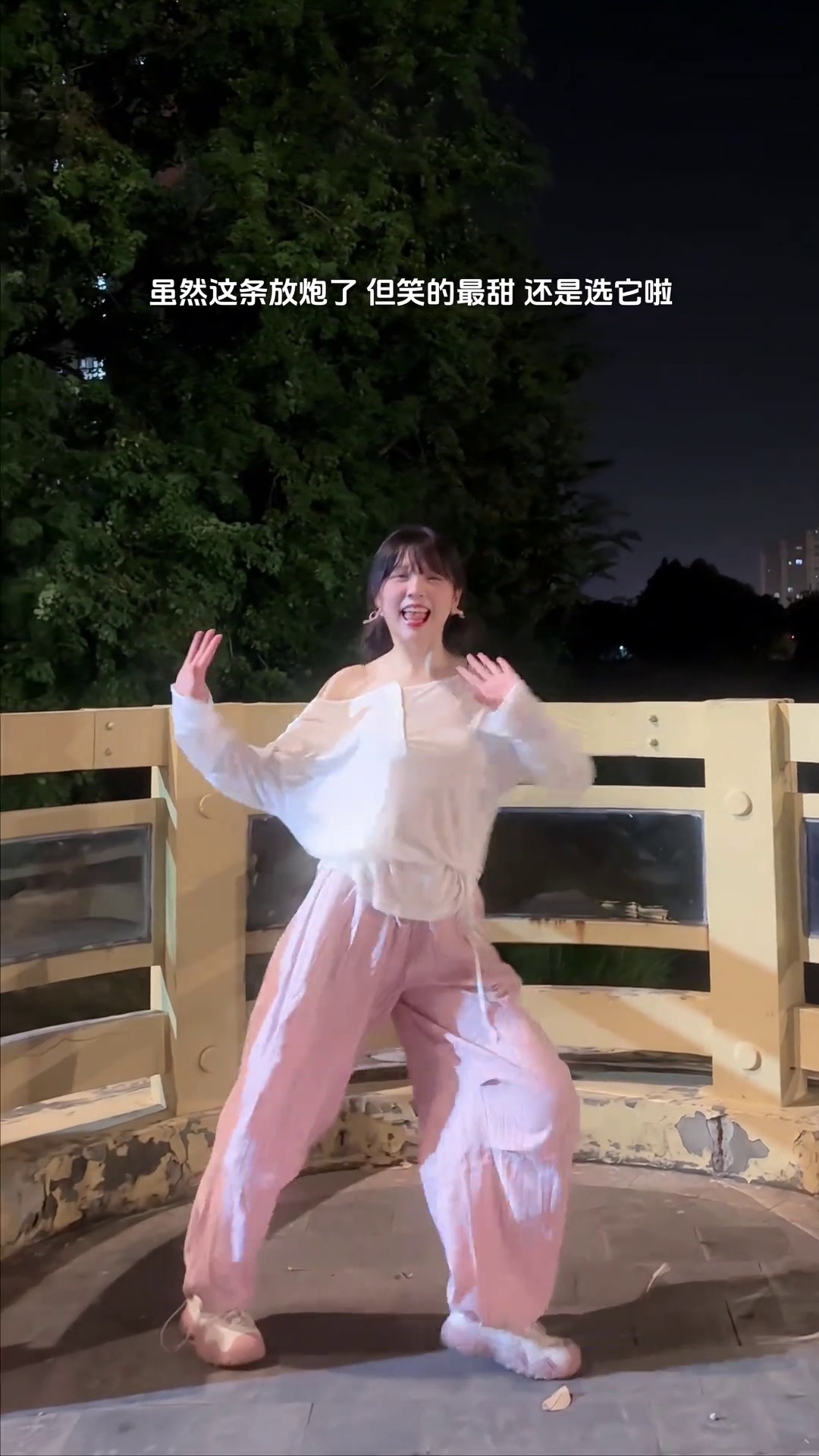} & \includegraphics[width=1.7cm]{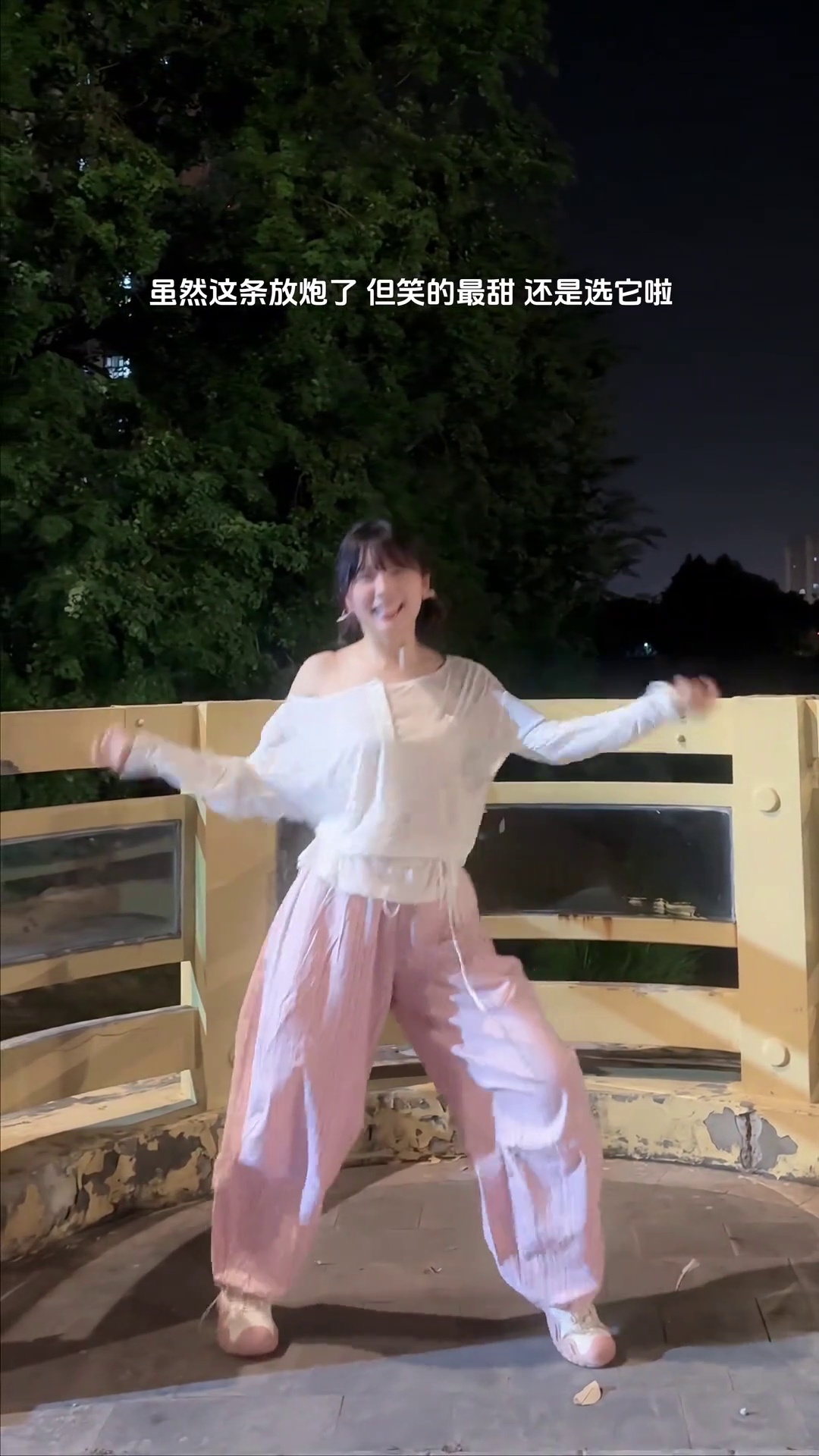} & \includegraphics[width=1.7cm]{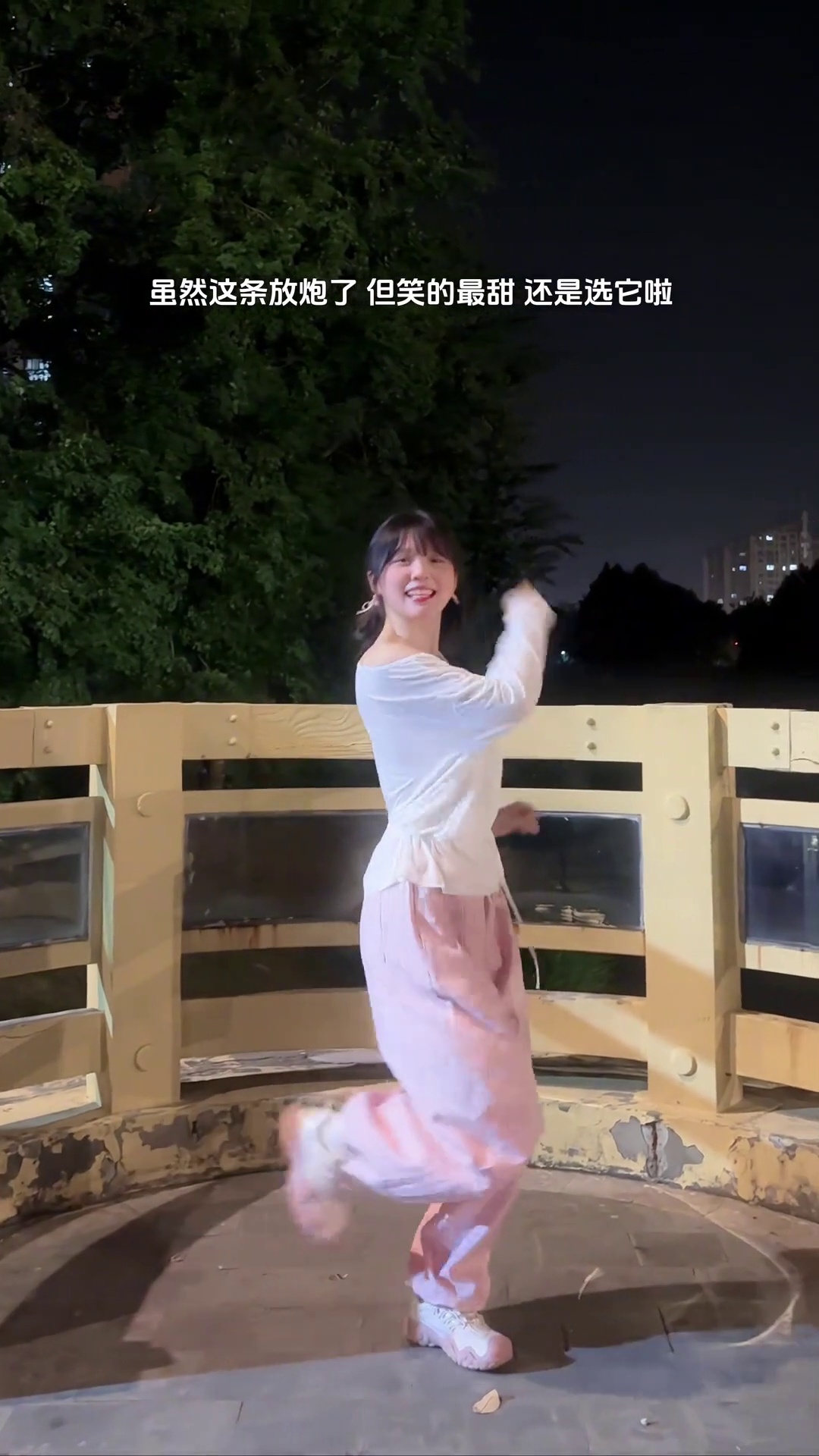} & \includegraphics[width=1.7cm]{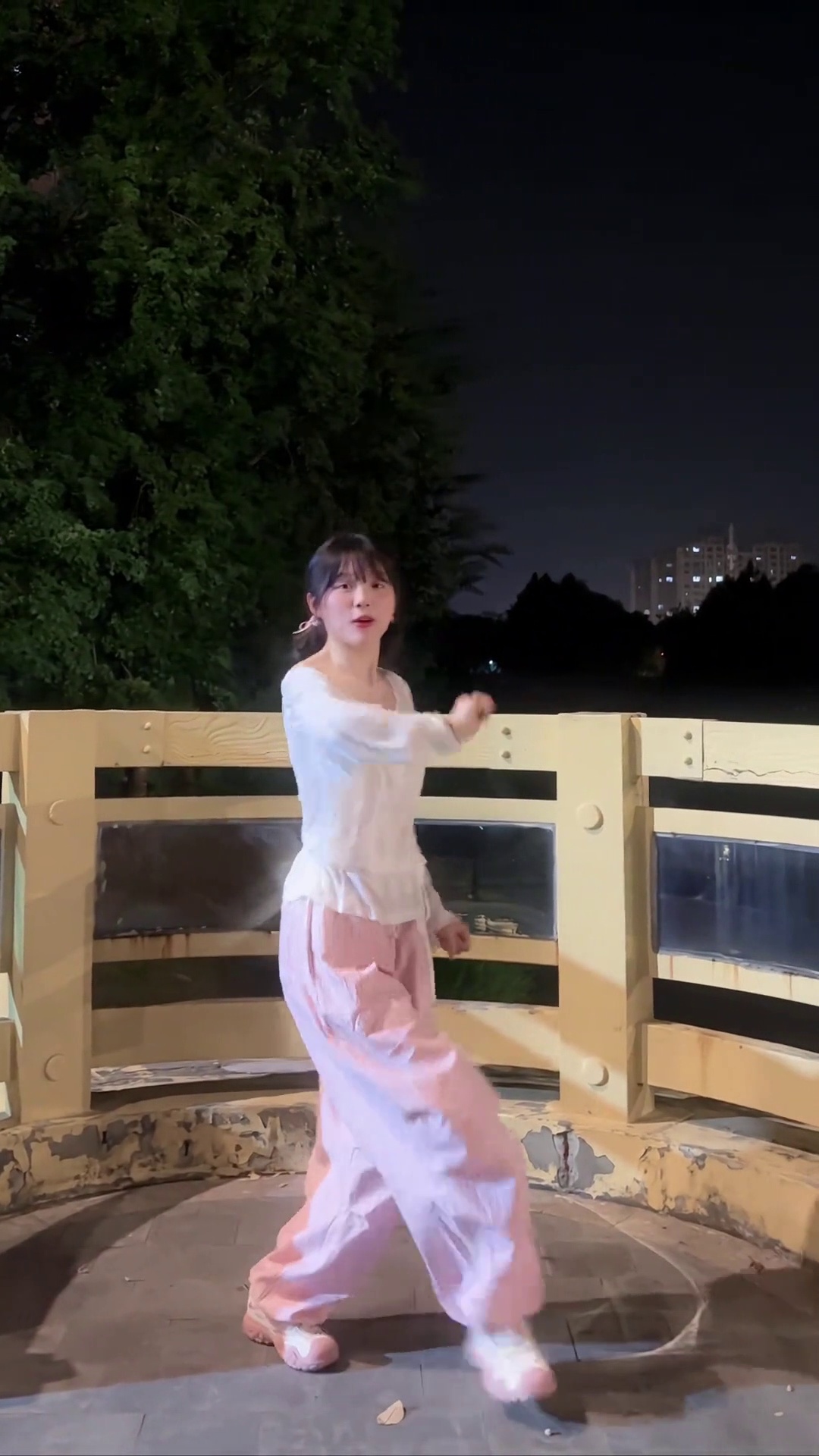} & \includegraphics[width=1.7cm]{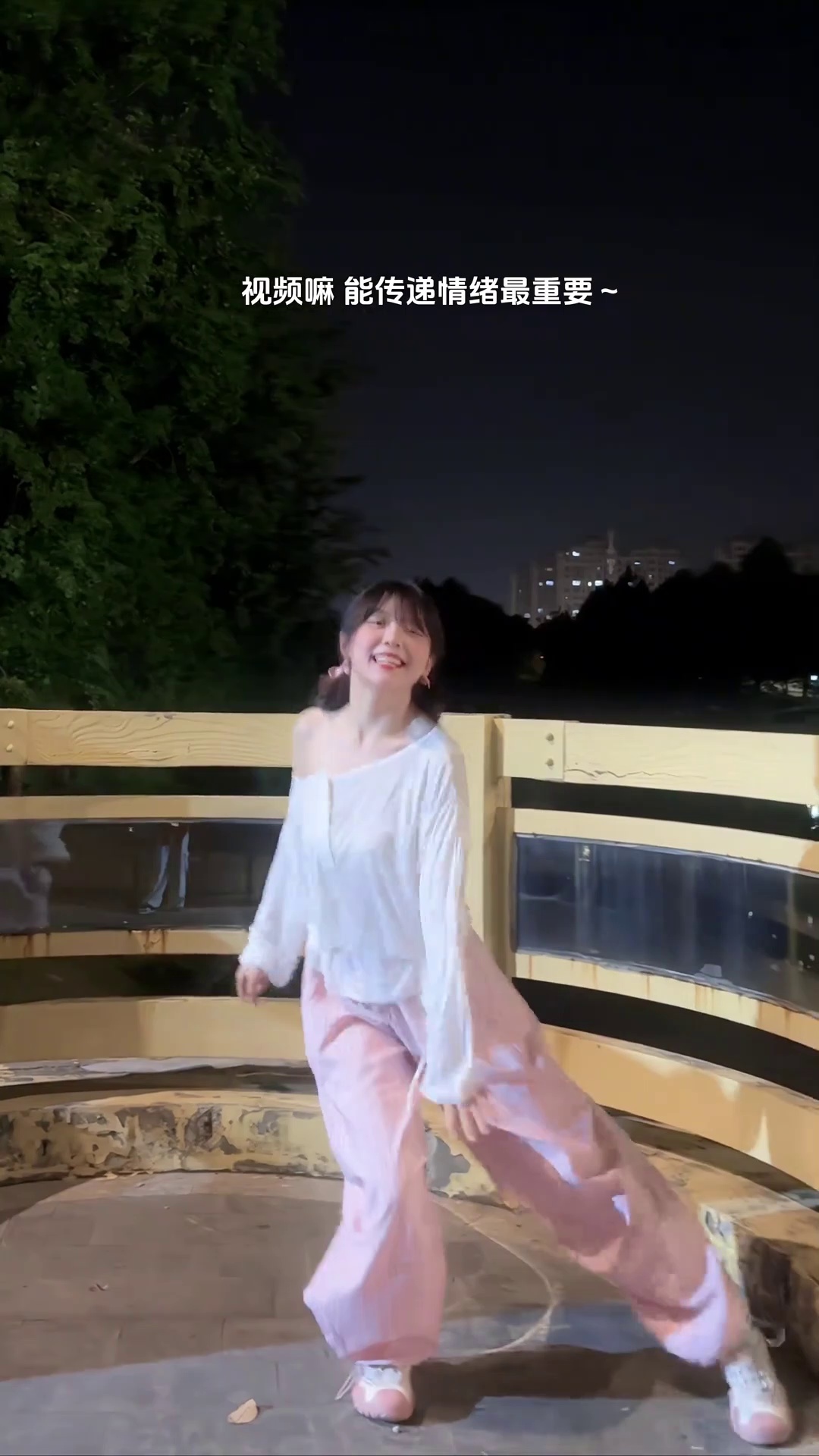} & \includegraphics[width=1.7cm]{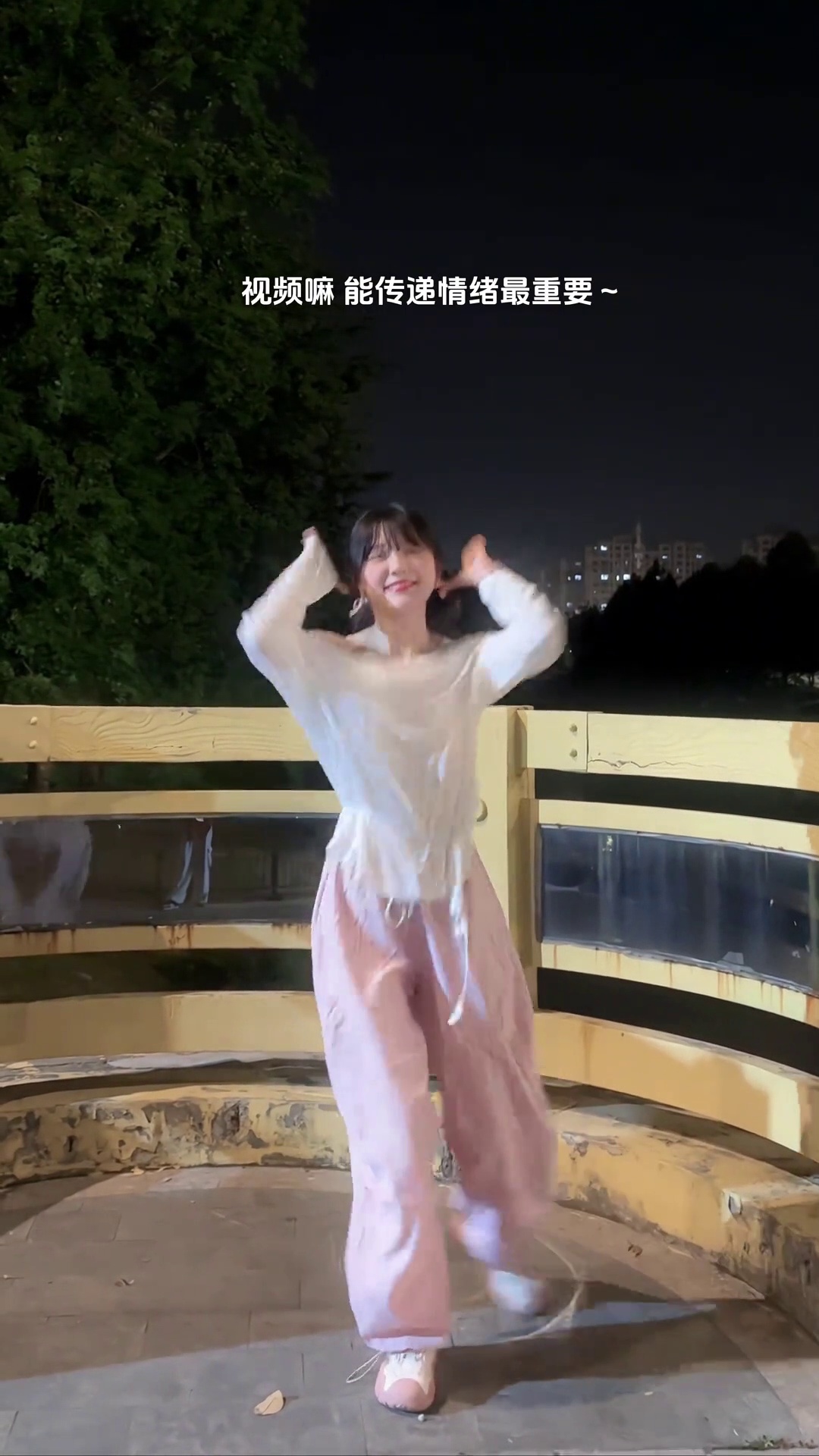} & \includegraphics[width=1.7cm]{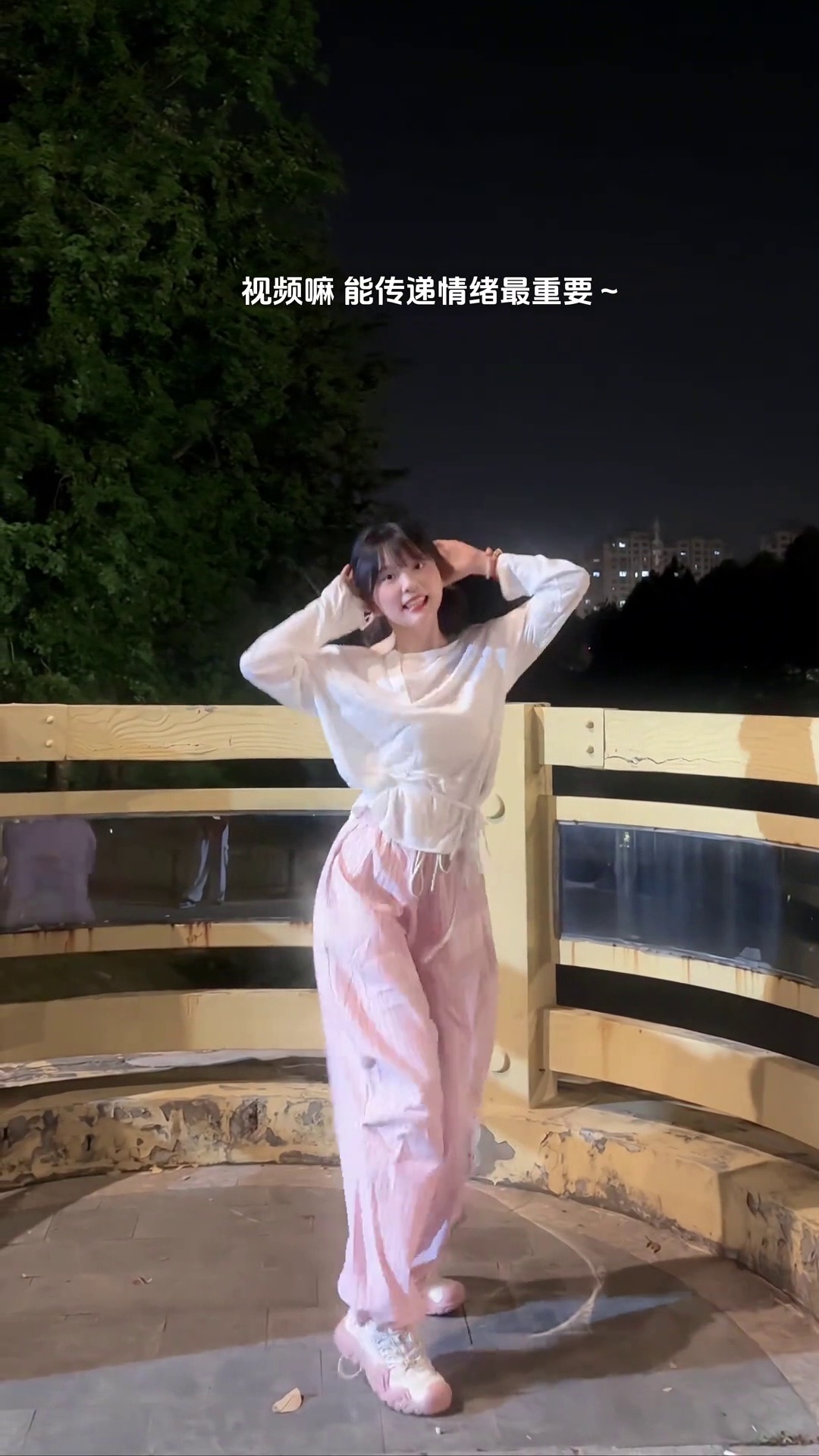} & \includegraphics[width=1.7cm]{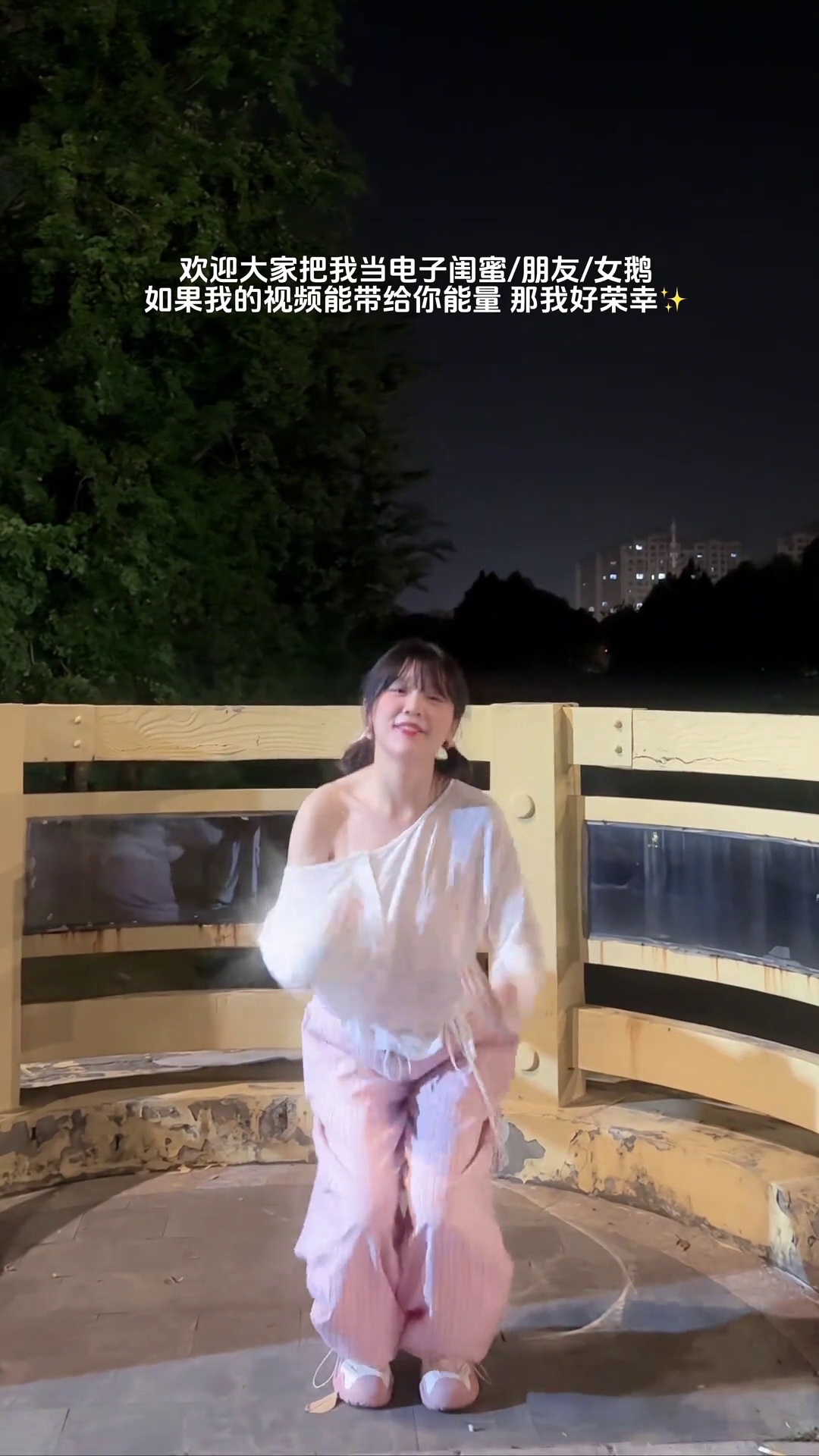} \\

    \scriptsize{Wan-Animate} & \includegraphics[width=1.7cm]{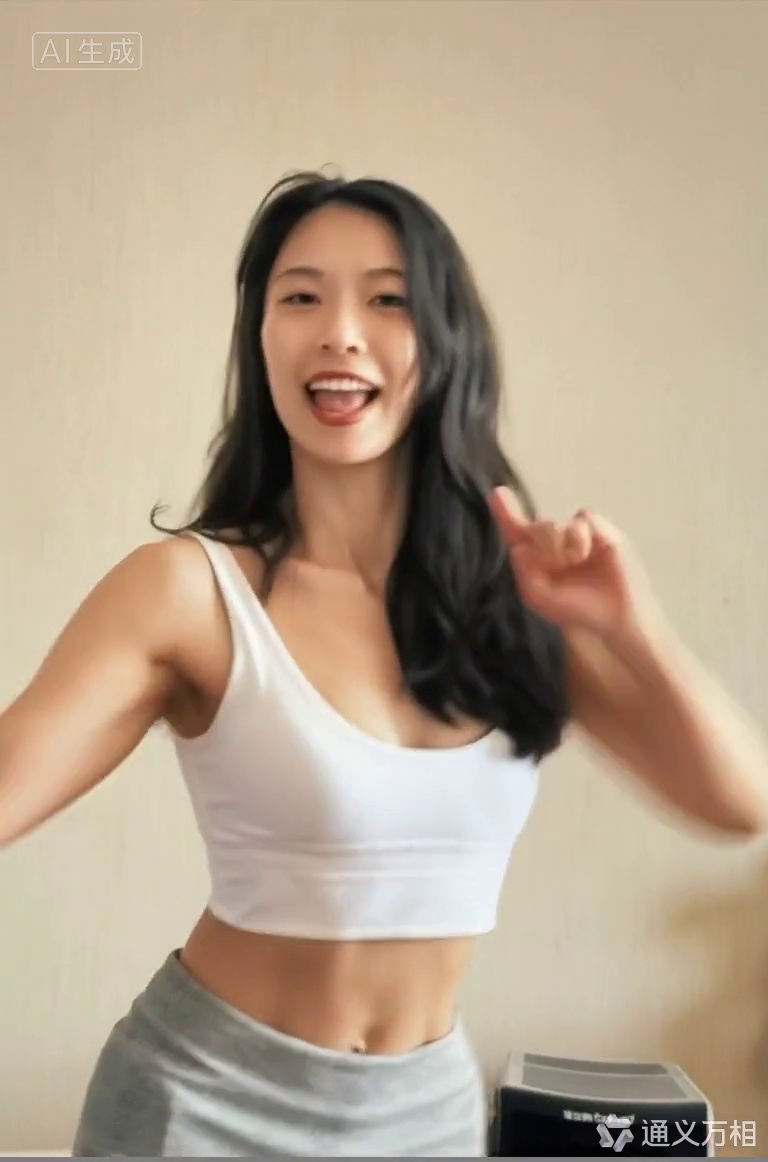} & \includegraphics[width=1.7cm]{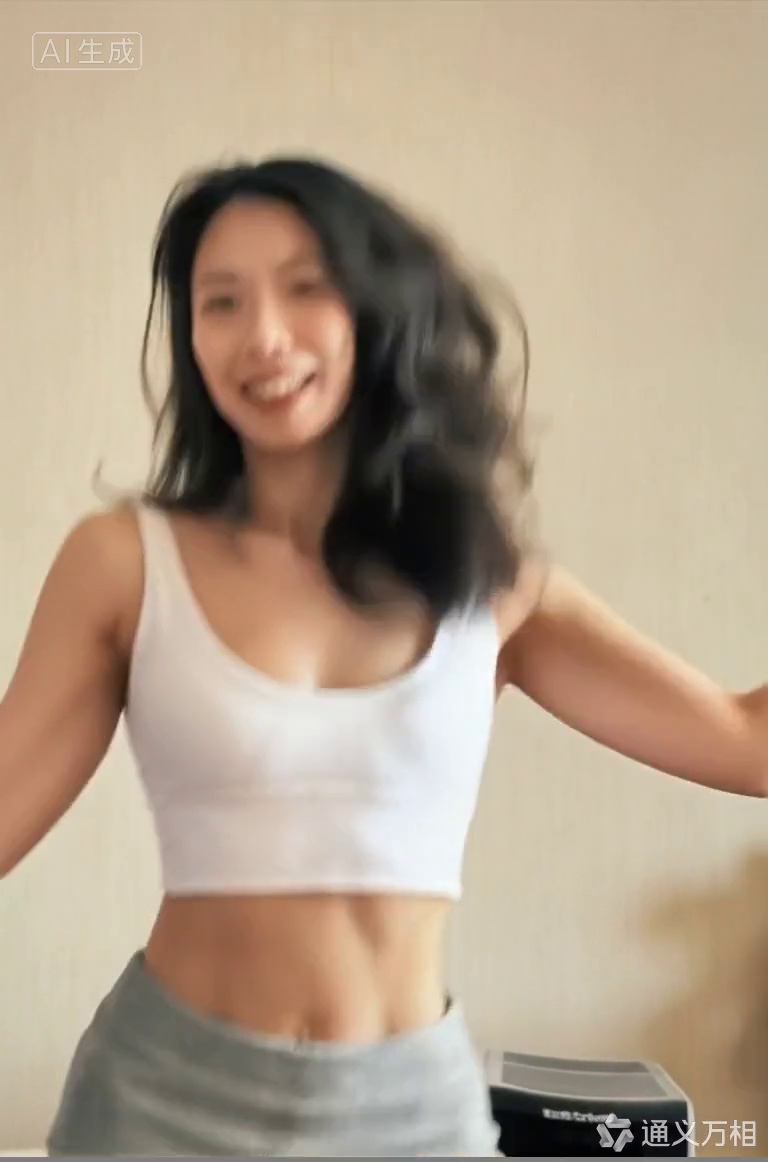} & \includegraphics[width=1.7cm]{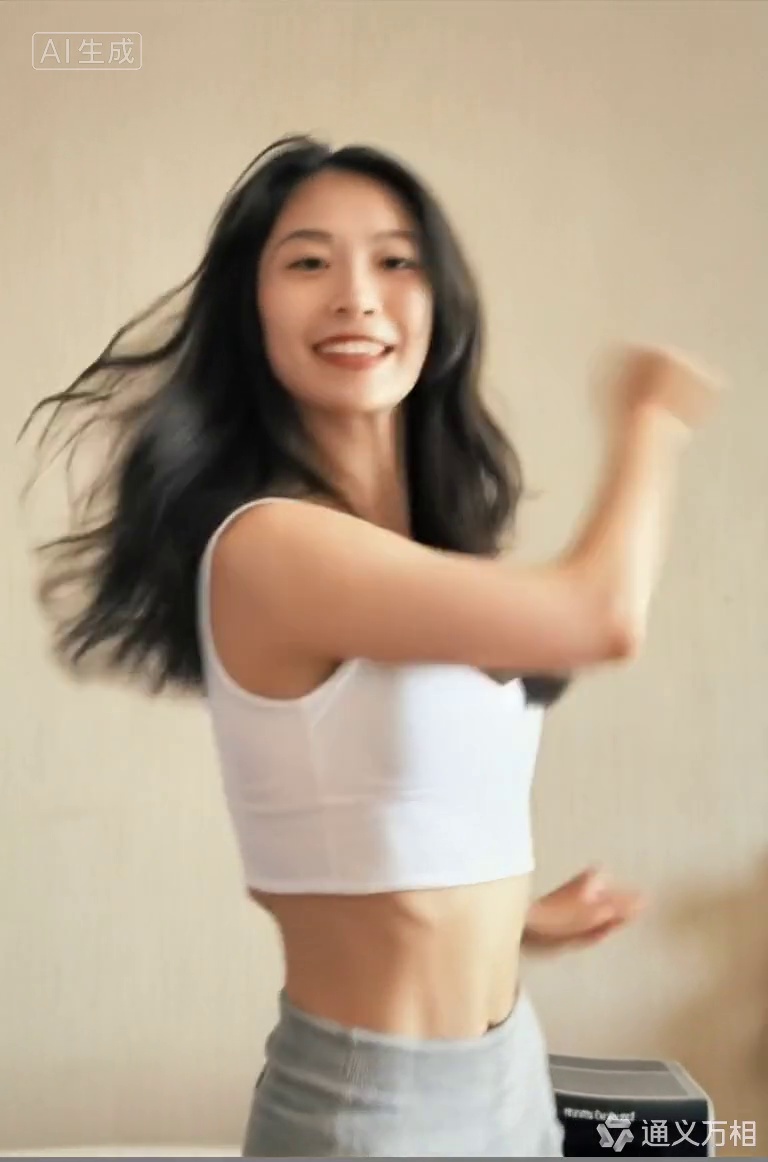} & \includegraphics[width=1.7cm]{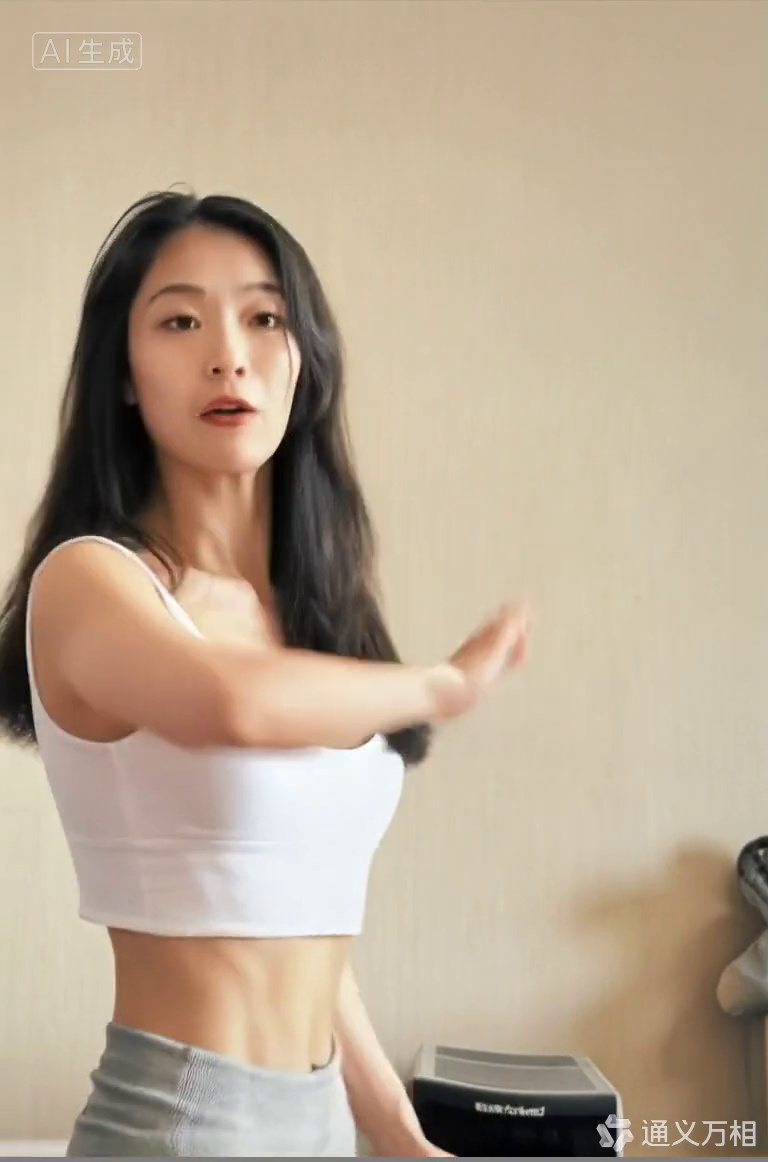} & \includegraphics[width=1.7cm]{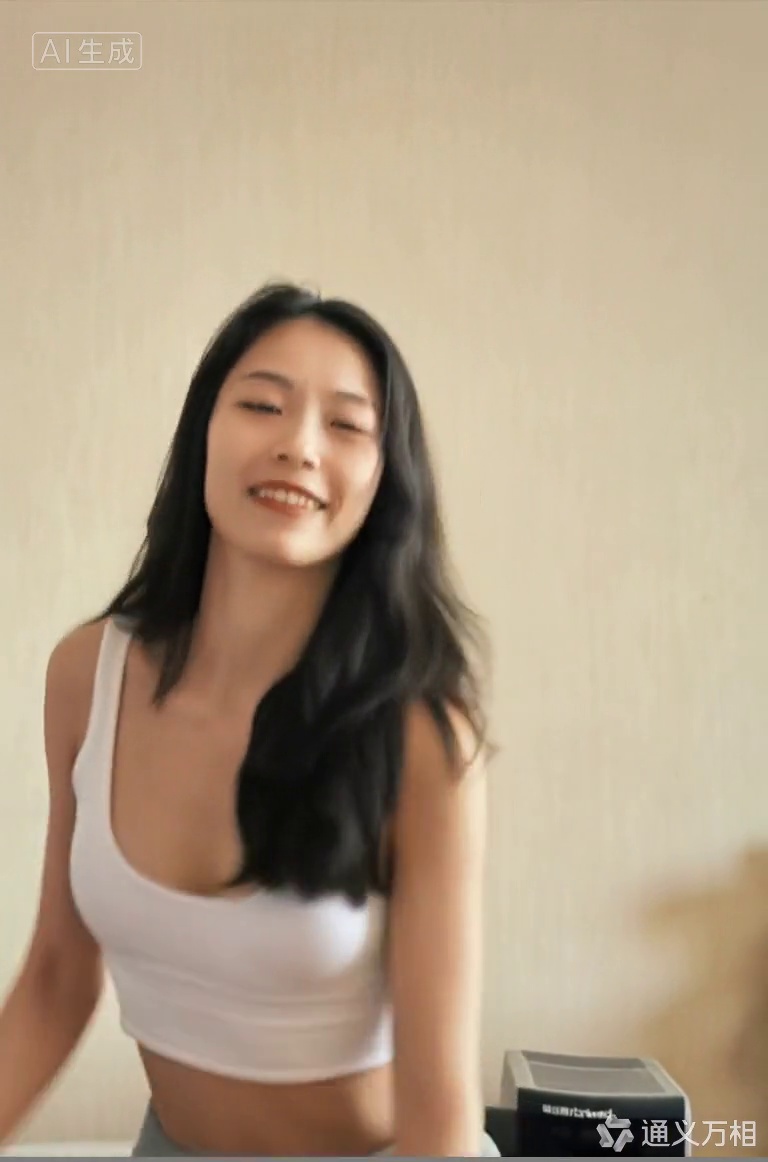} & \includegraphics[width=1.7cm]{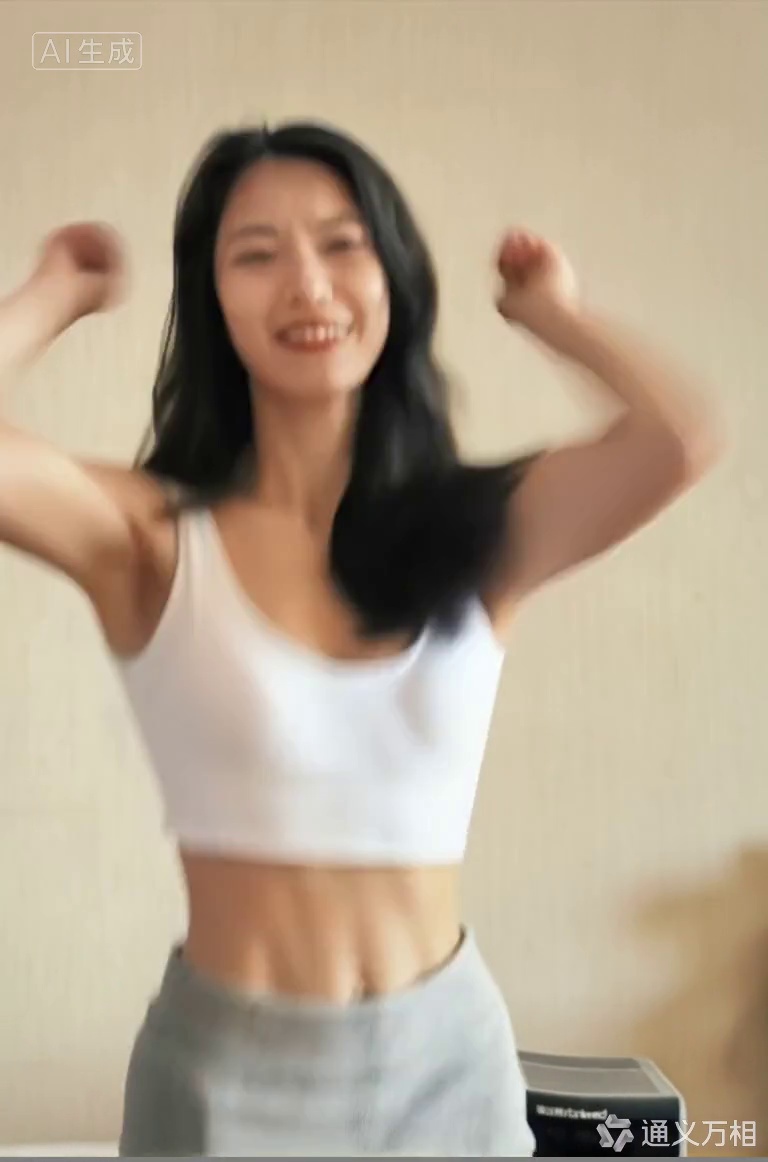} & \includegraphics[width=1.7cm]{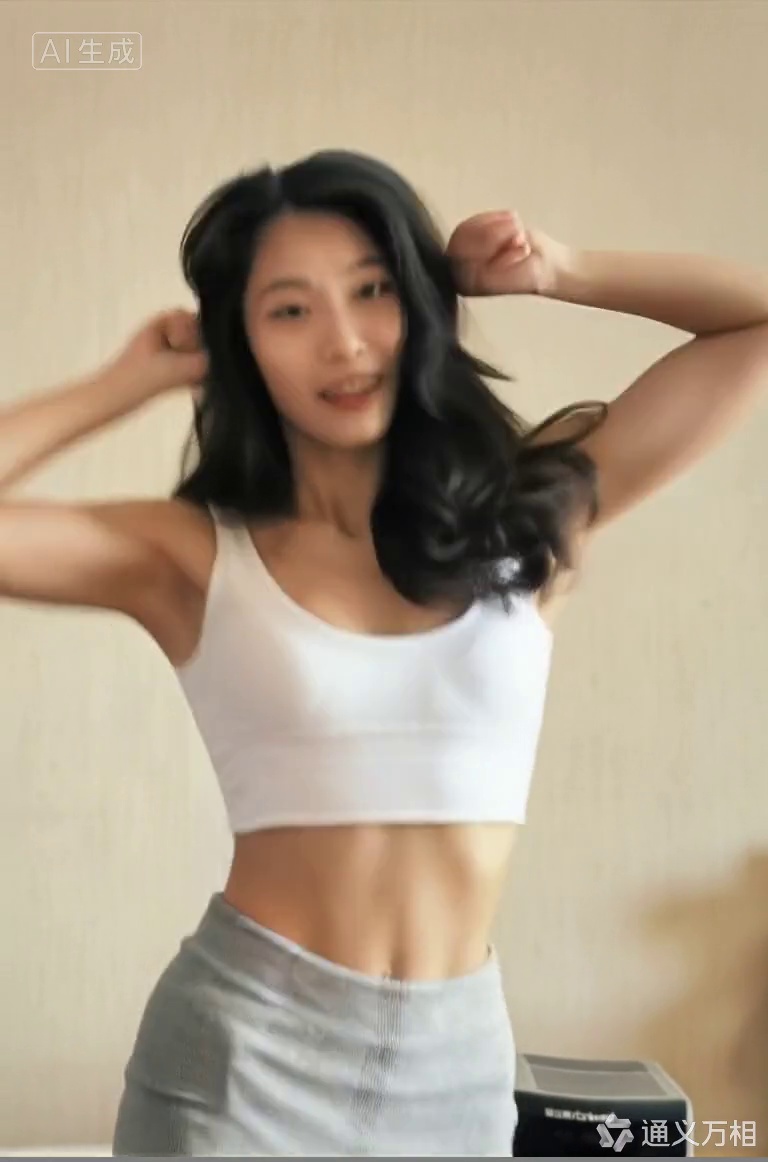} & \includegraphics[width=1.7cm]{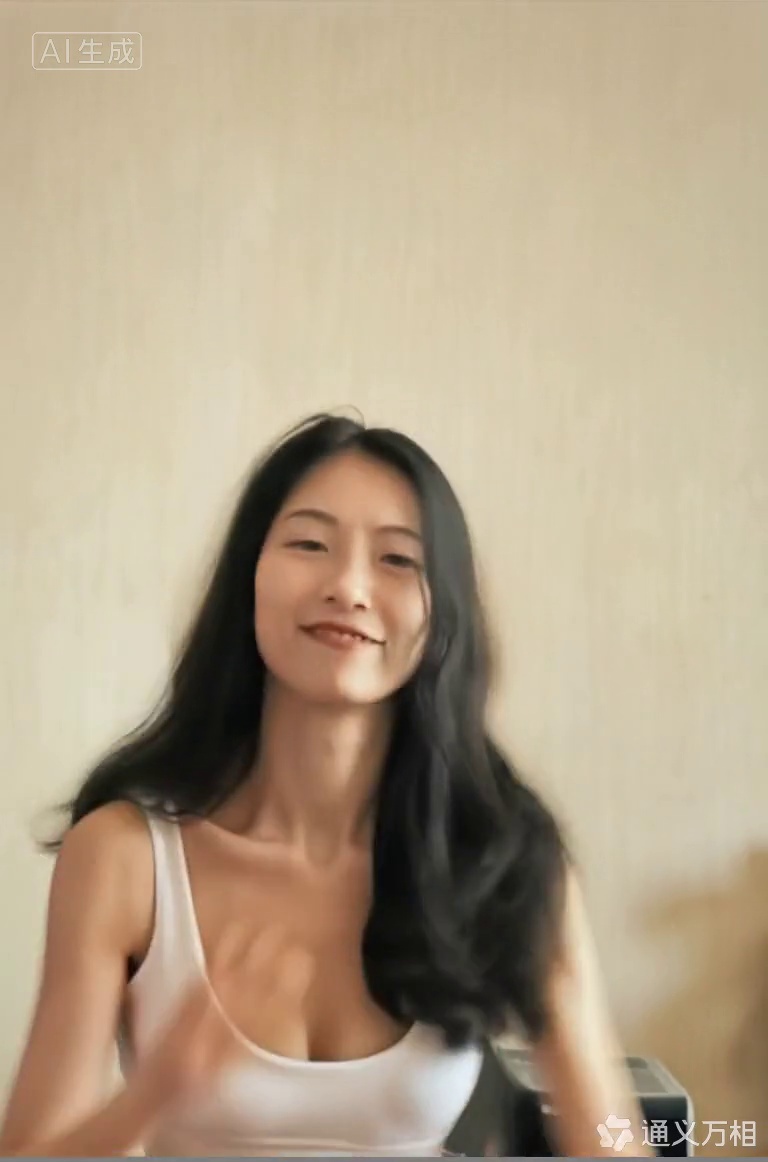} \\

    \scriptsize{Dreamina} & \includegraphics[width=1.7cm]{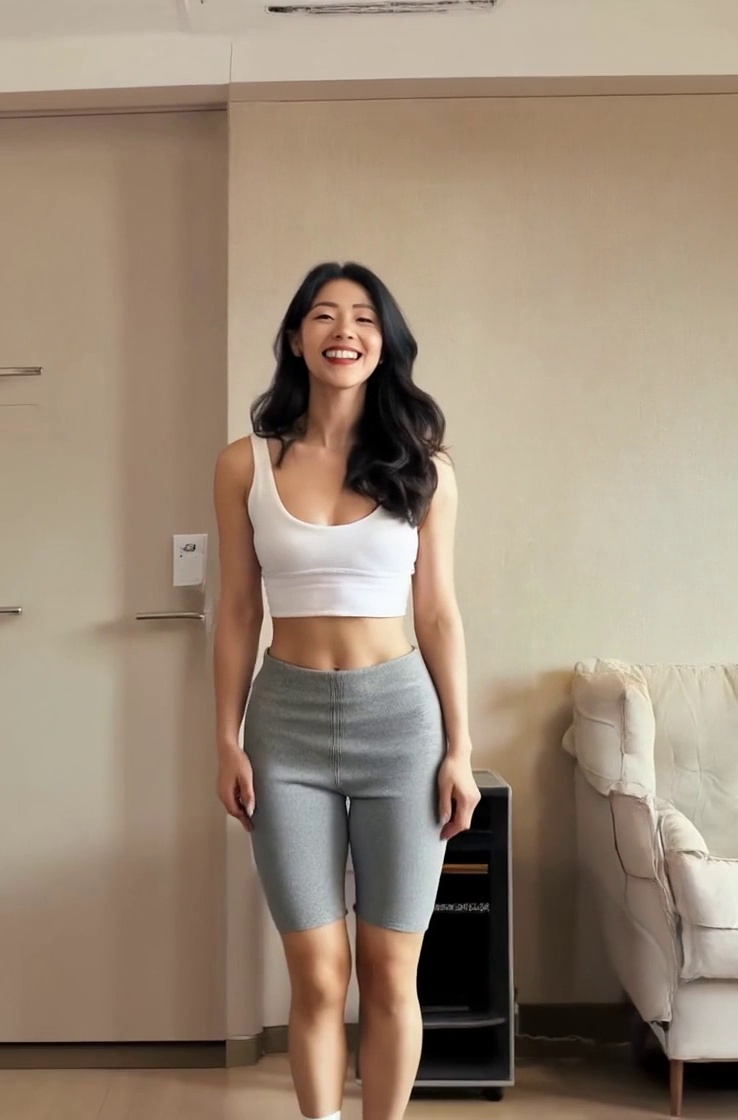} & \includegraphics[width=1.7cm]{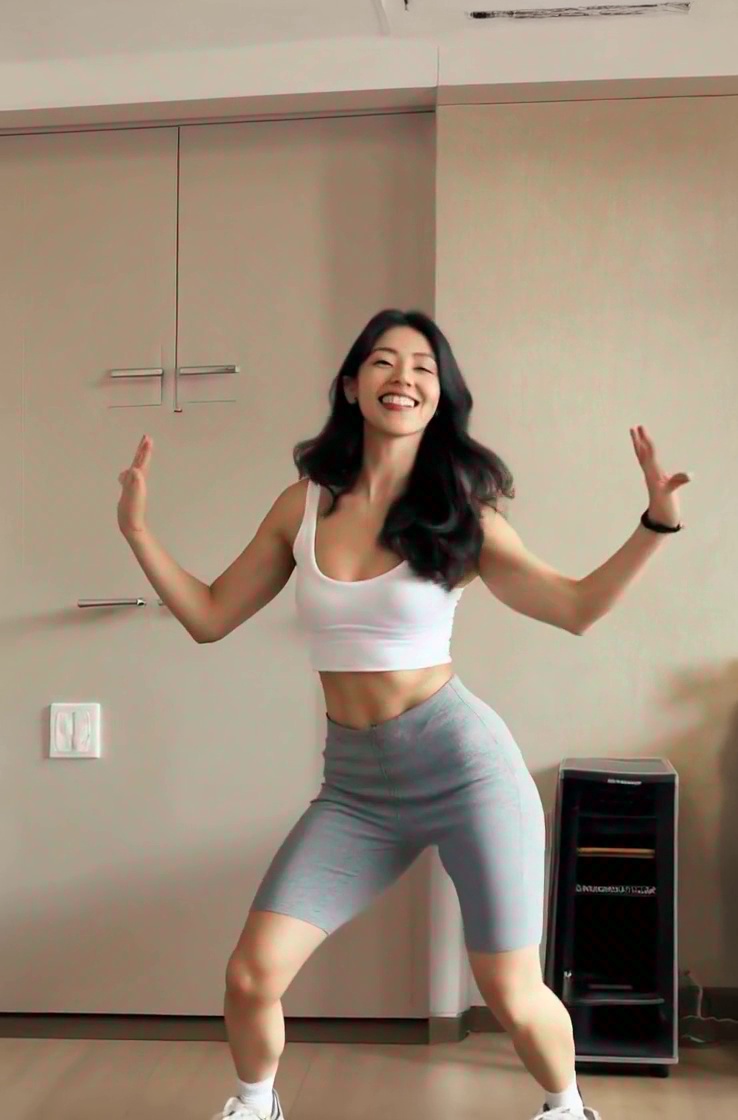} & \includegraphics[width=1.7cm]{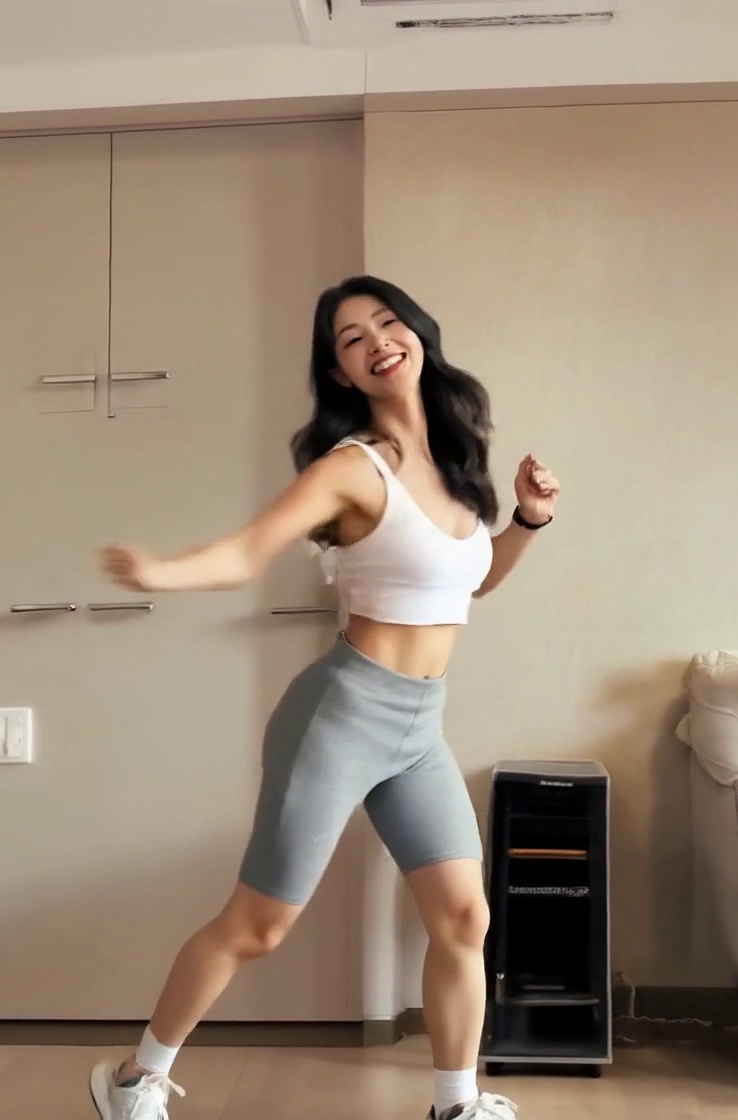} & \includegraphics[width=1.7cm]{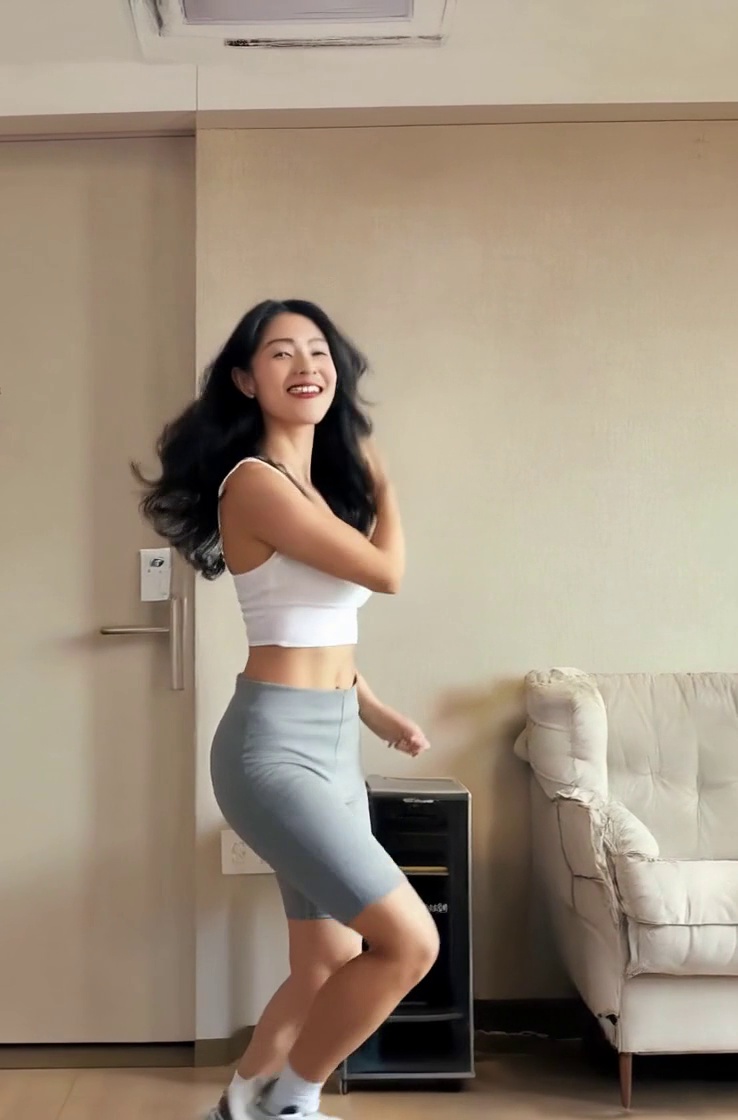} & \includegraphics[width=1.7cm]{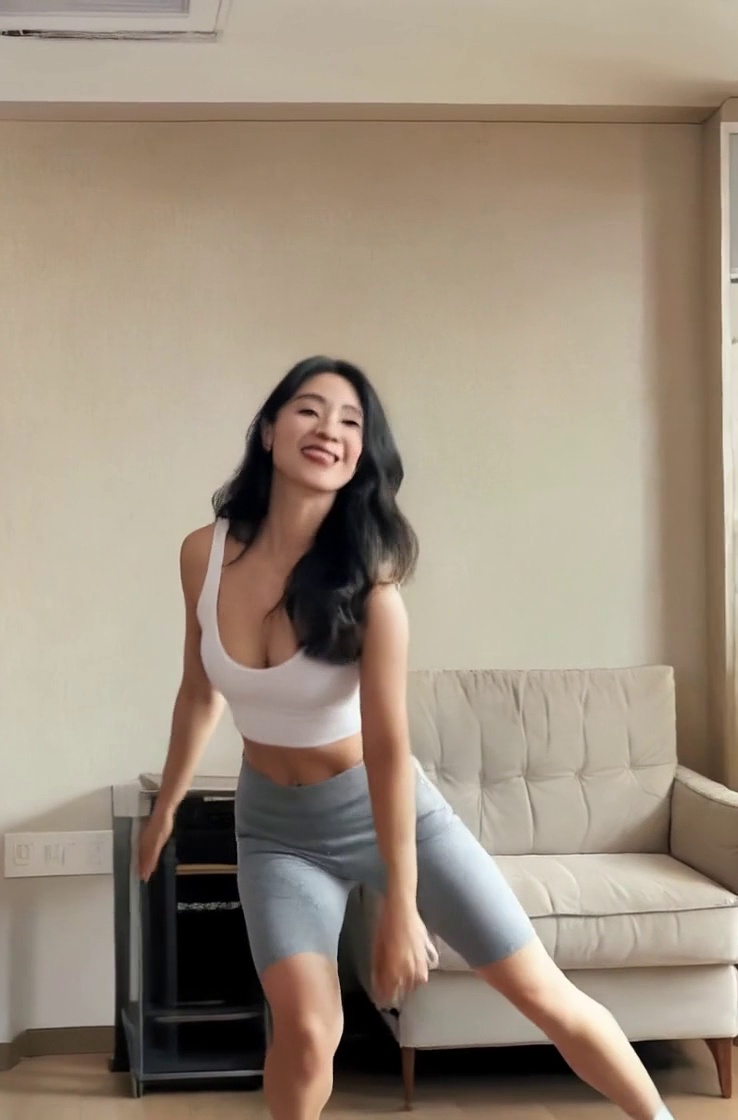} & \includegraphics[width=1.7cm]{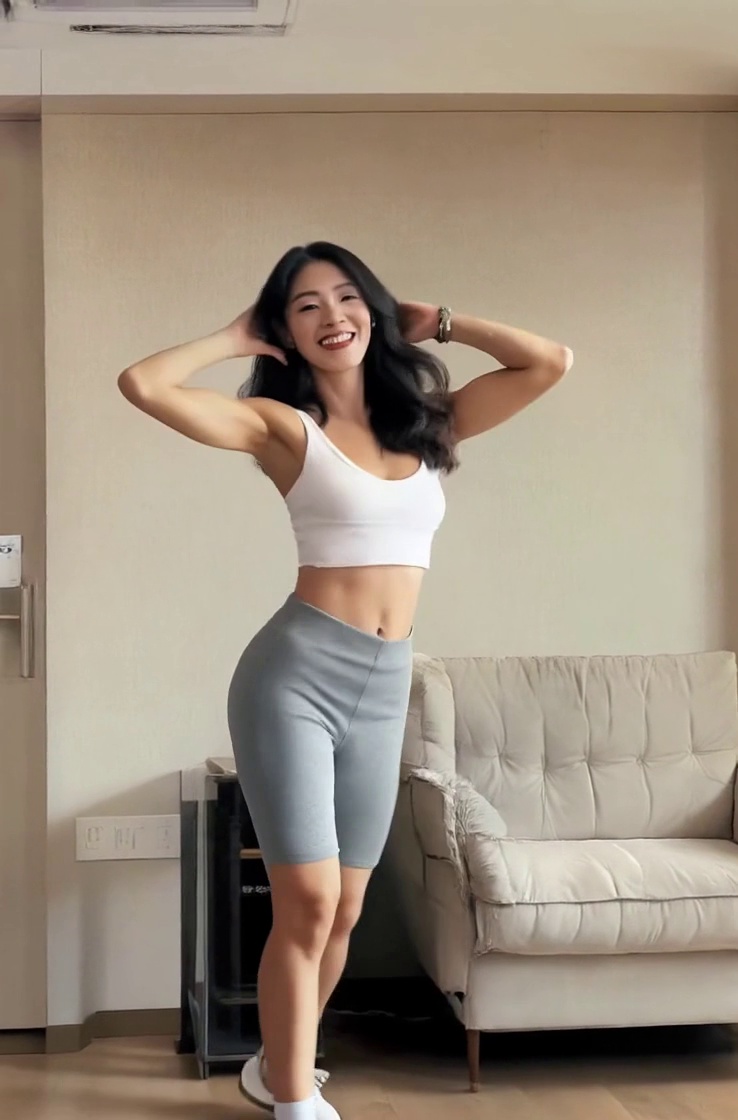} & \includegraphics[width=1.7cm]{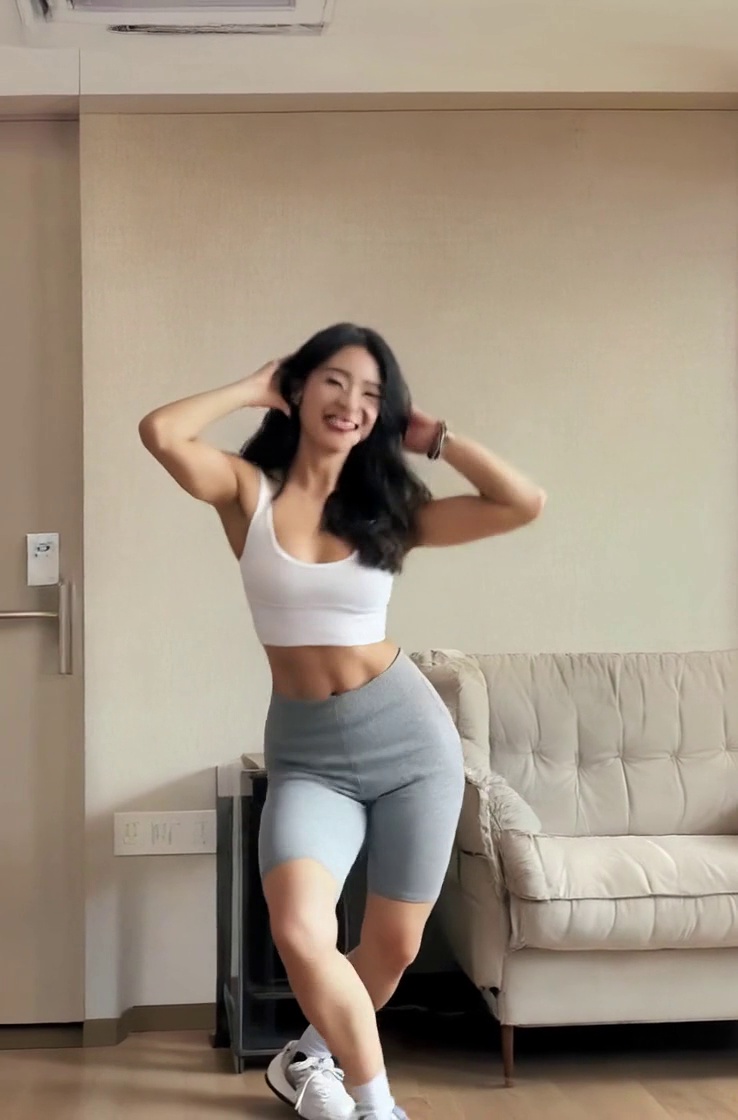} & \includegraphics[width=1.7cm]{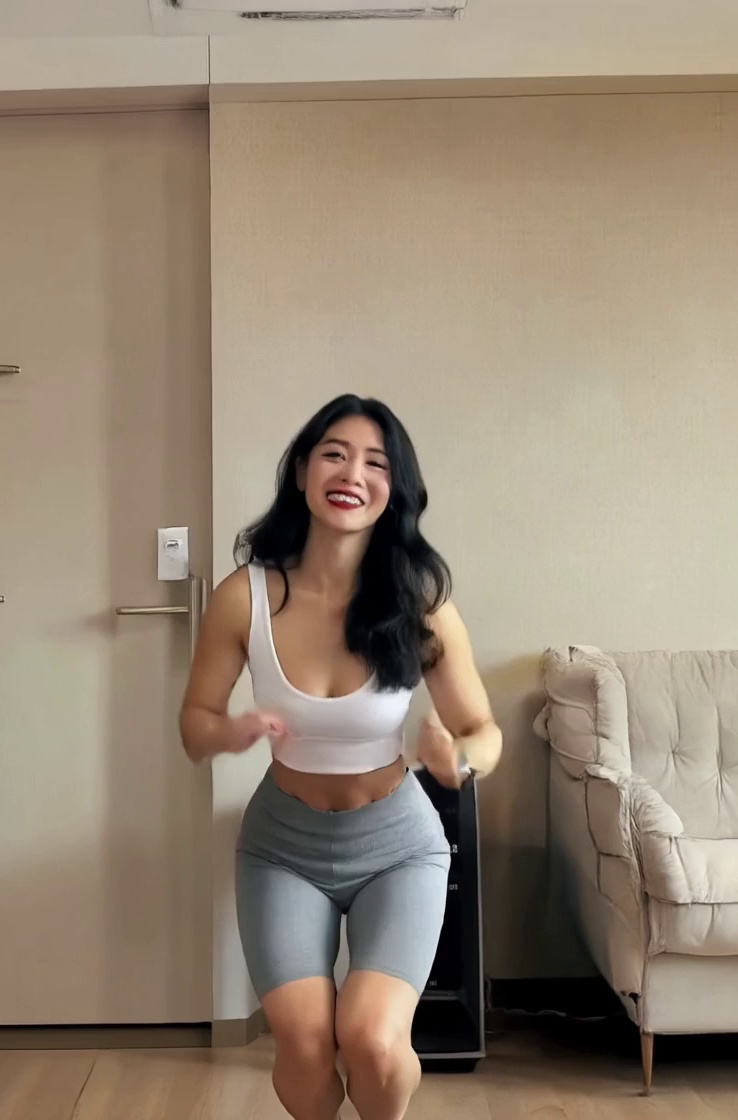} \\

    \scriptsize{Kling-MotionControl} & \includegraphics[width=1.7cm]{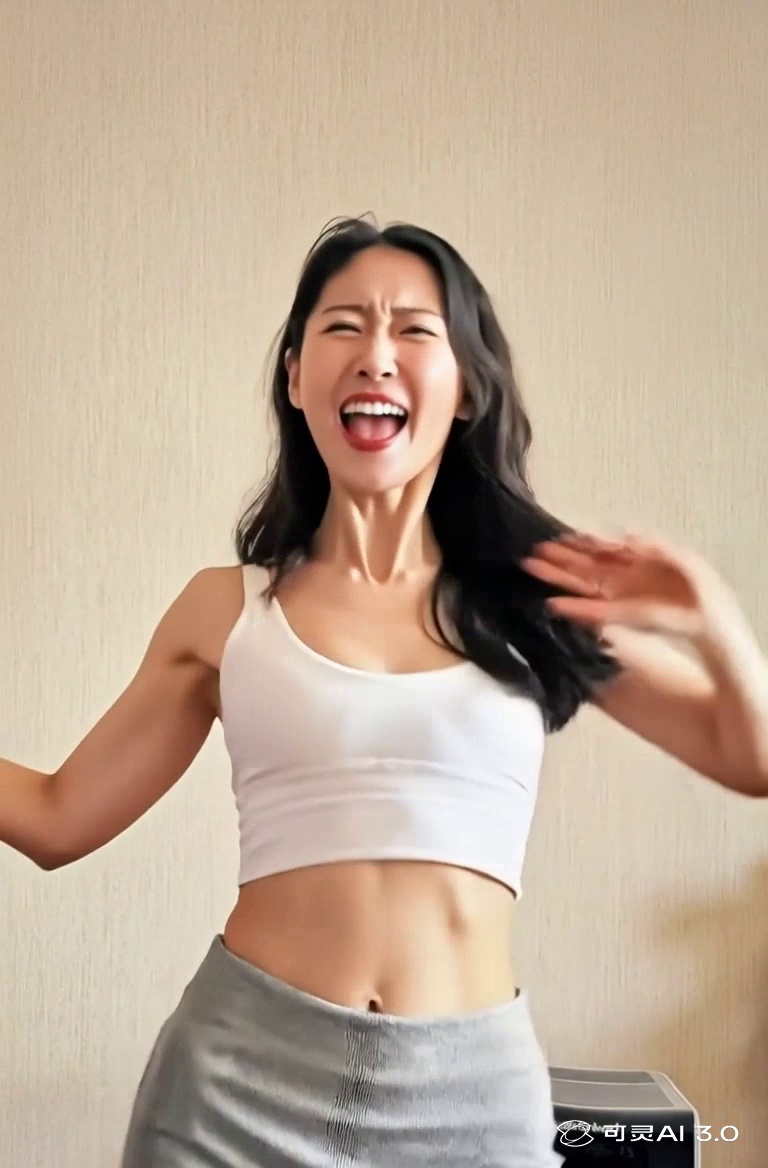} & \includegraphics[width=1.7cm]{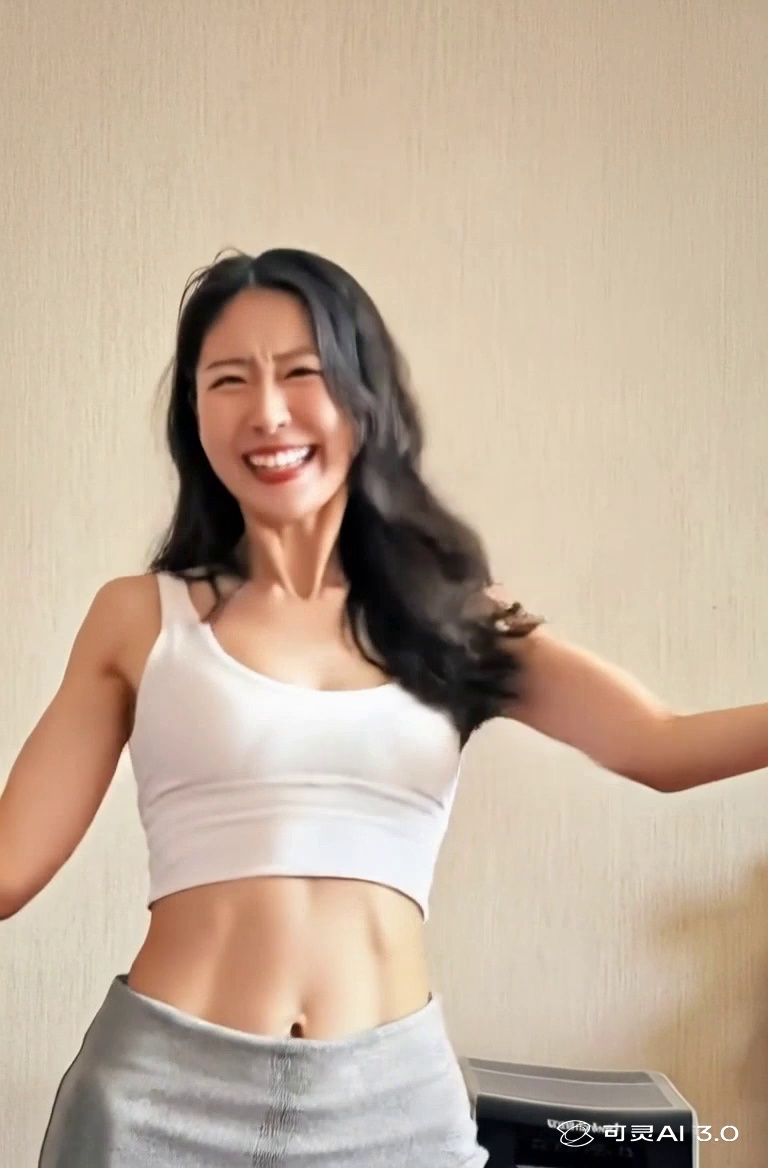} & \includegraphics[width=1.7cm]{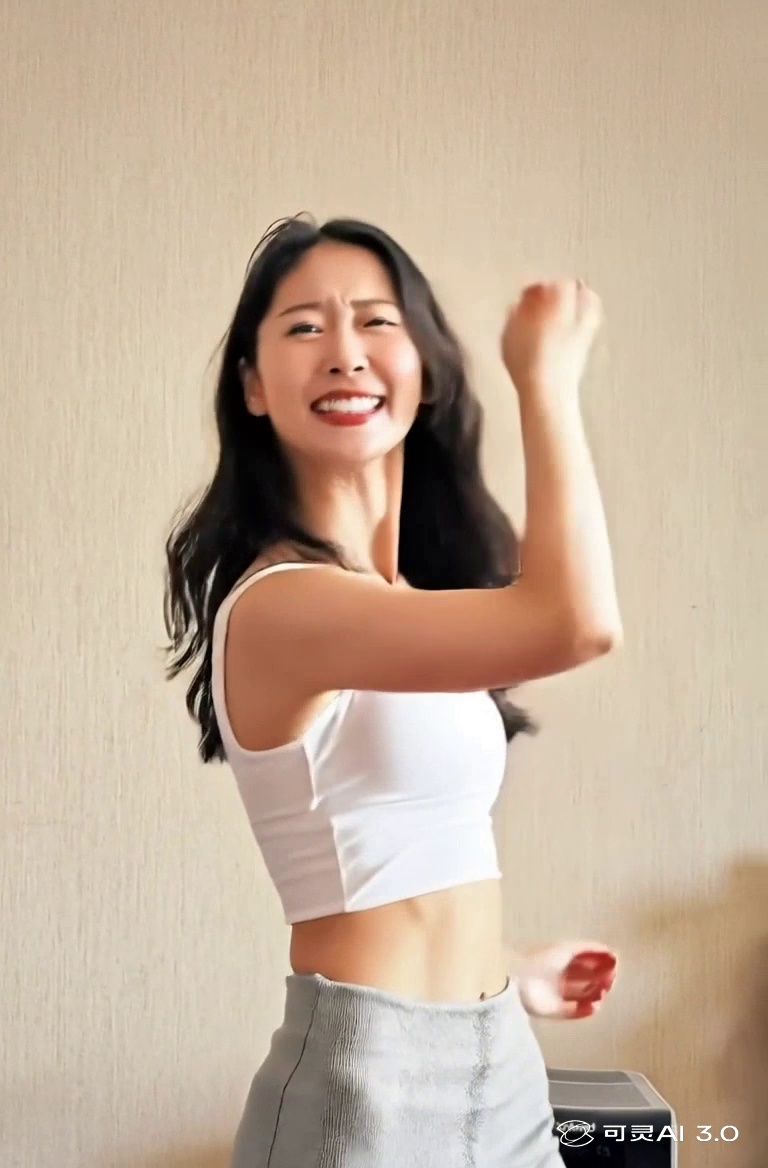} & \includegraphics[width=1.7cm]{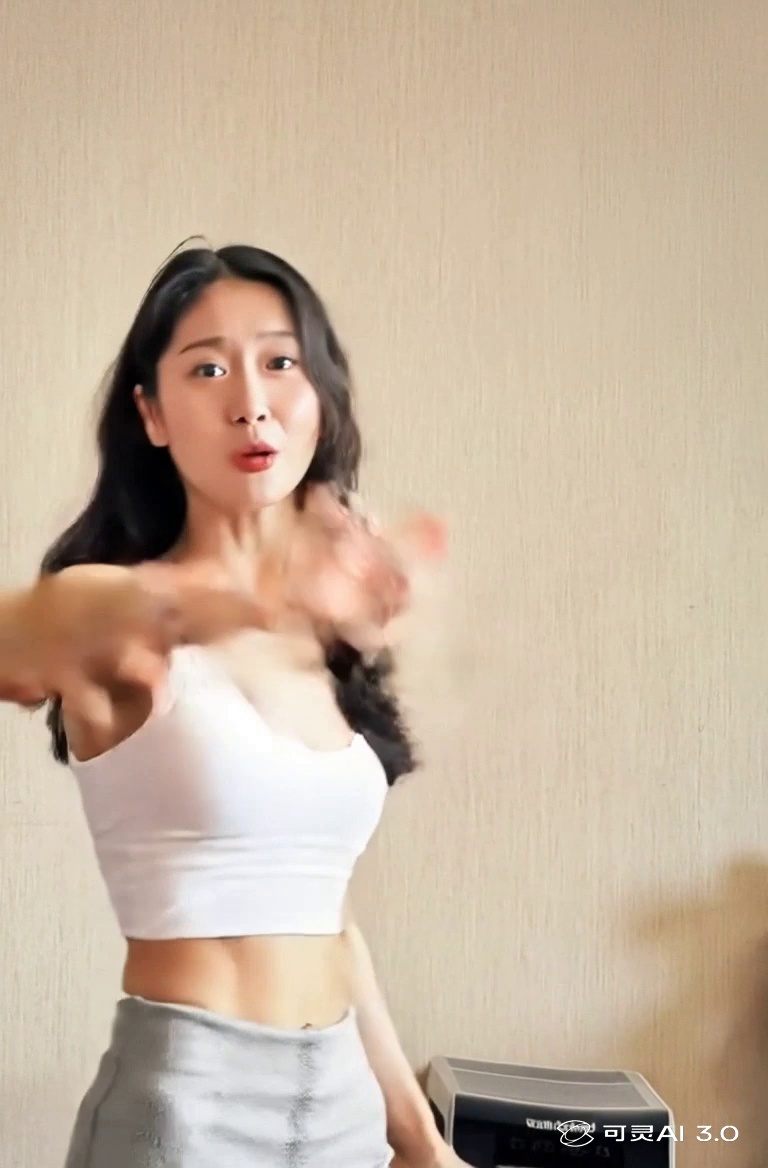} & \includegraphics[width=1.7cm]{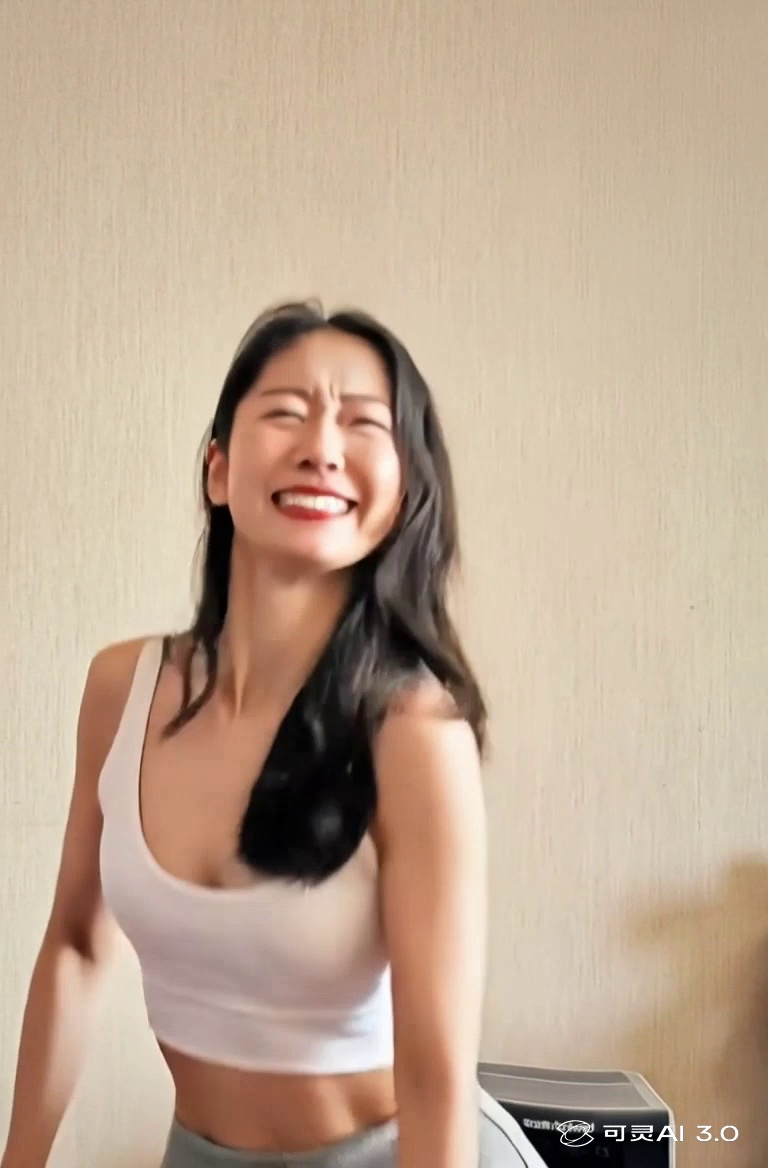} & \includegraphics[width=1.7cm]{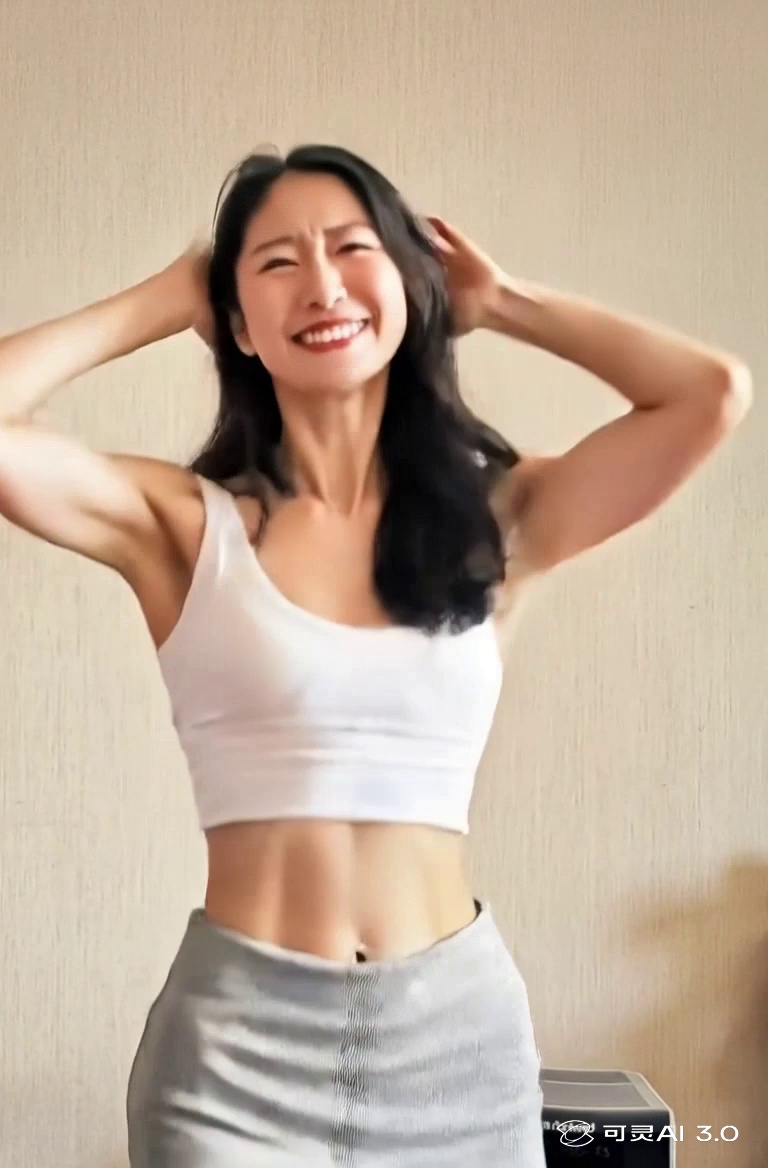} & \includegraphics[width=1.7cm]{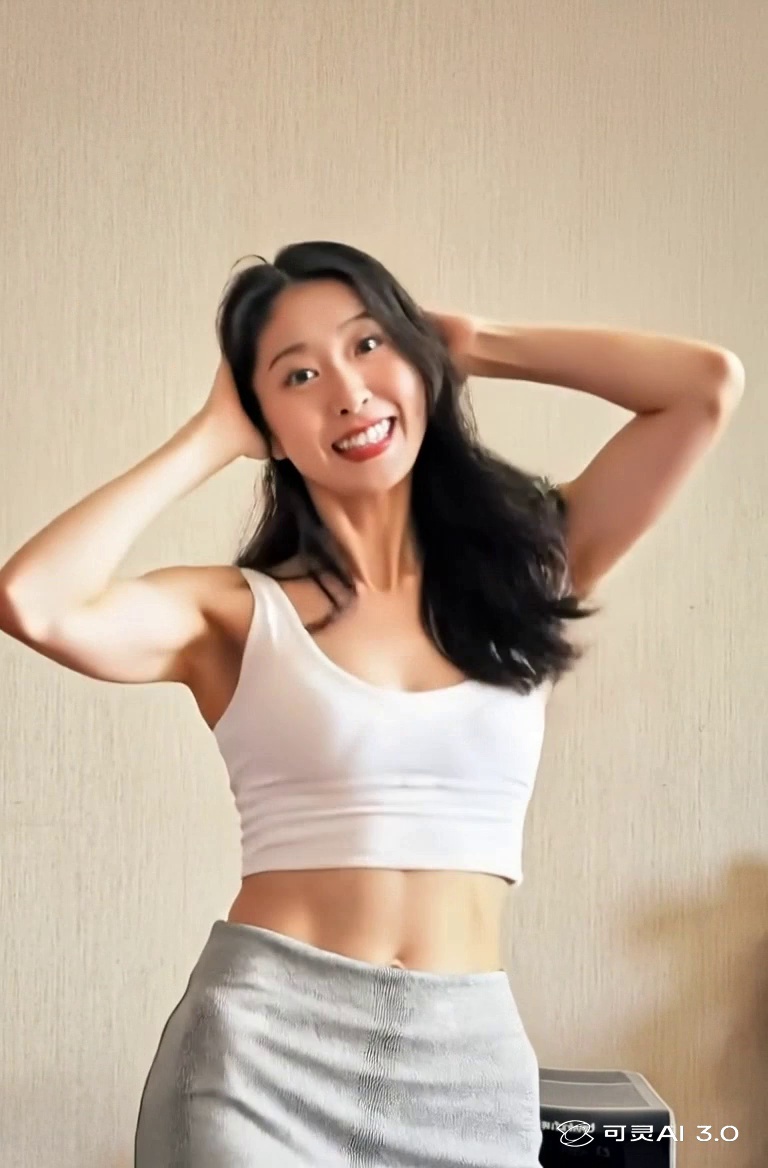} & \includegraphics[width=1.7cm]{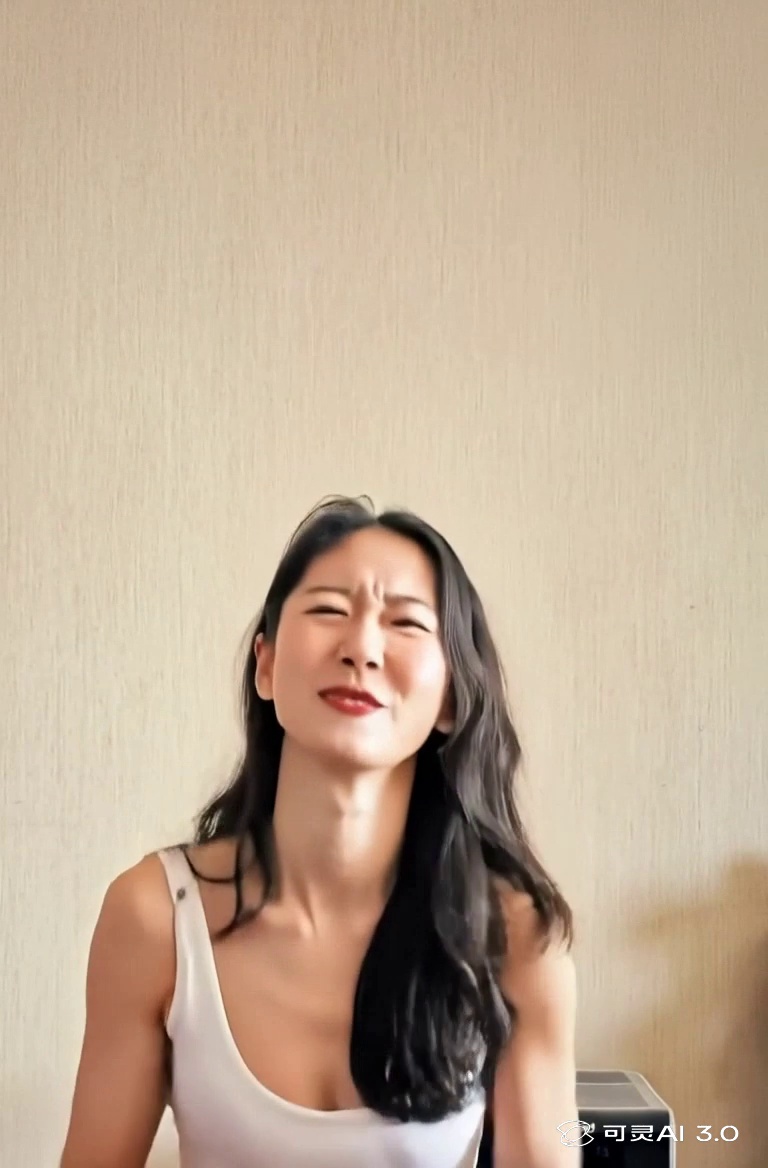} \\

    \scriptsize{Ours} & \includegraphics[width=1.7cm]{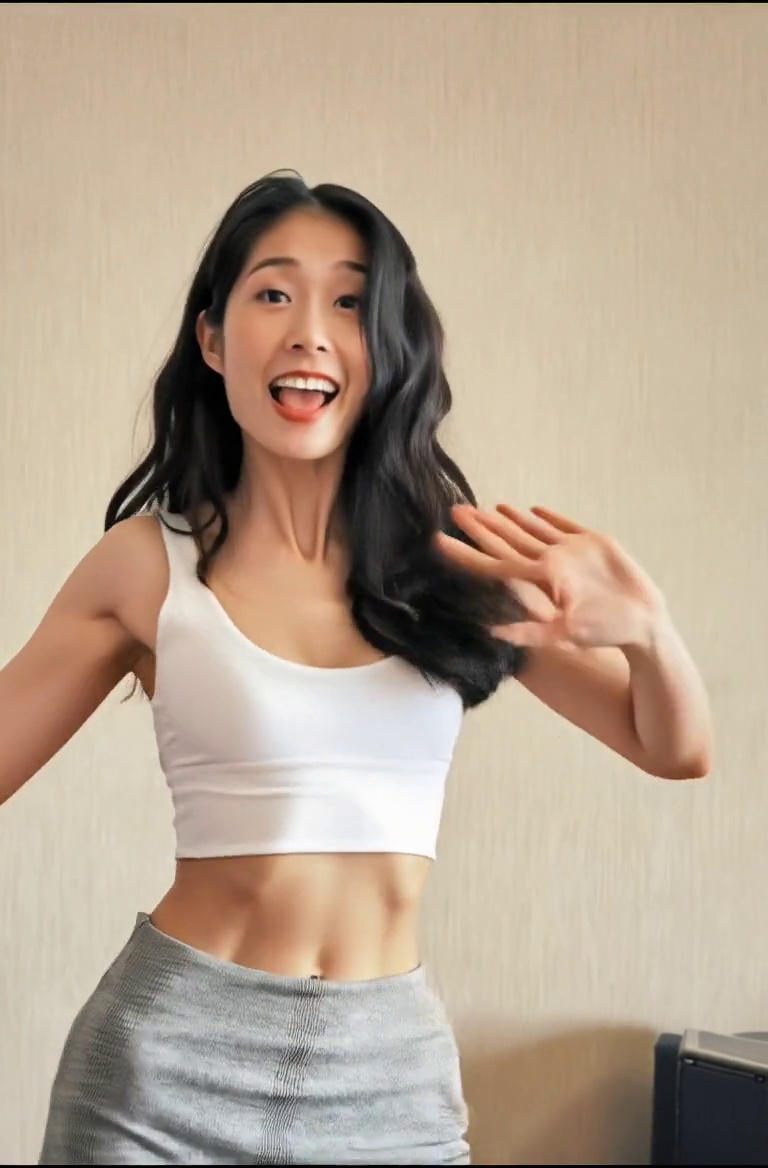} & \includegraphics[width=1.7cm]{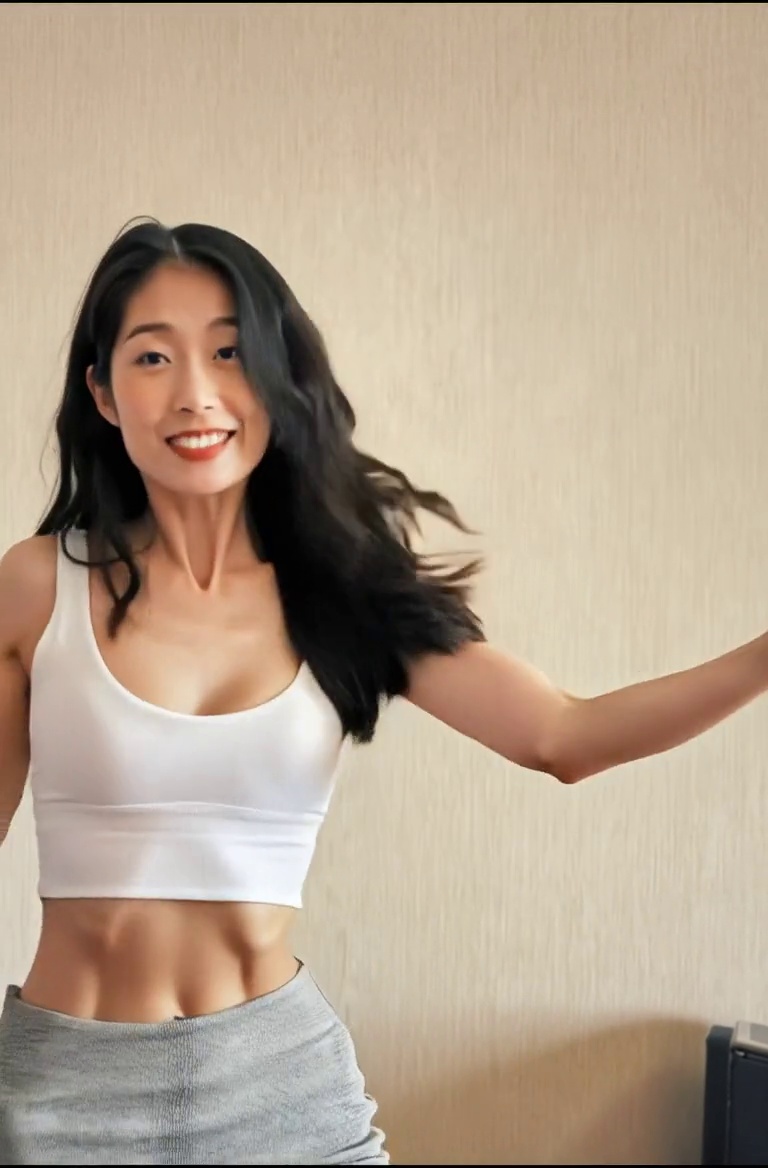} & \includegraphics[width=1.7cm]{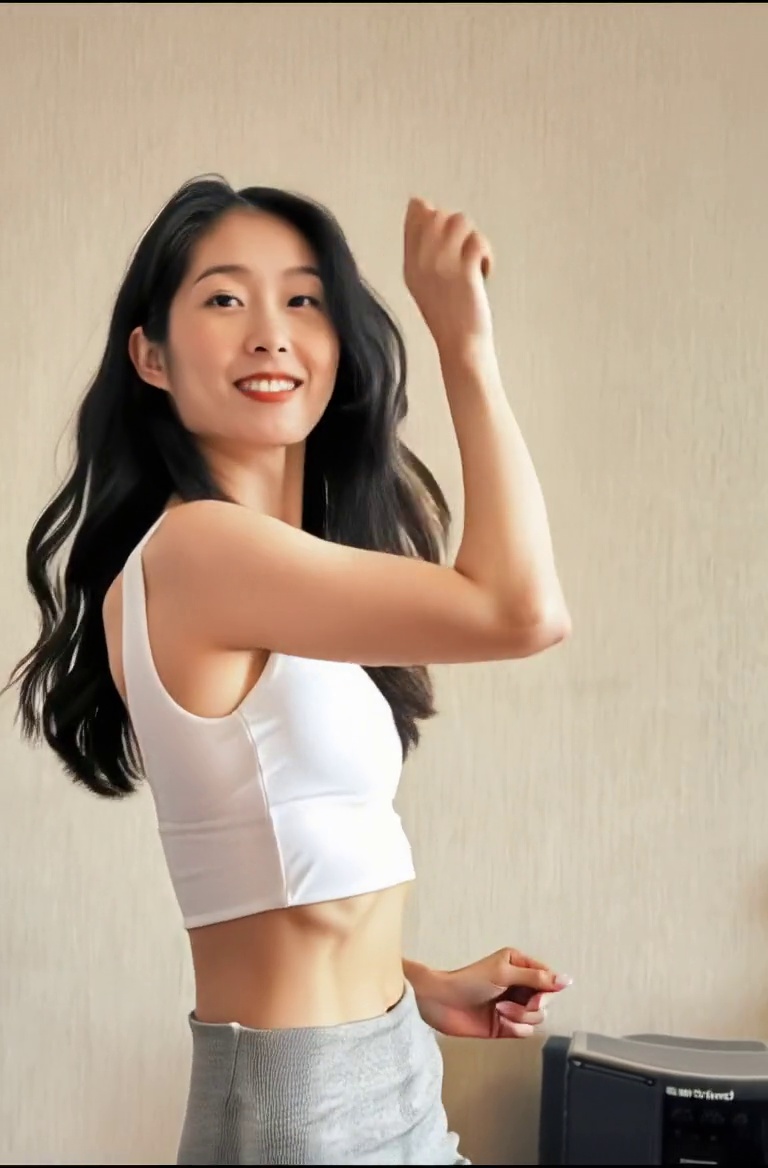} & \includegraphics[width=1.7cm]{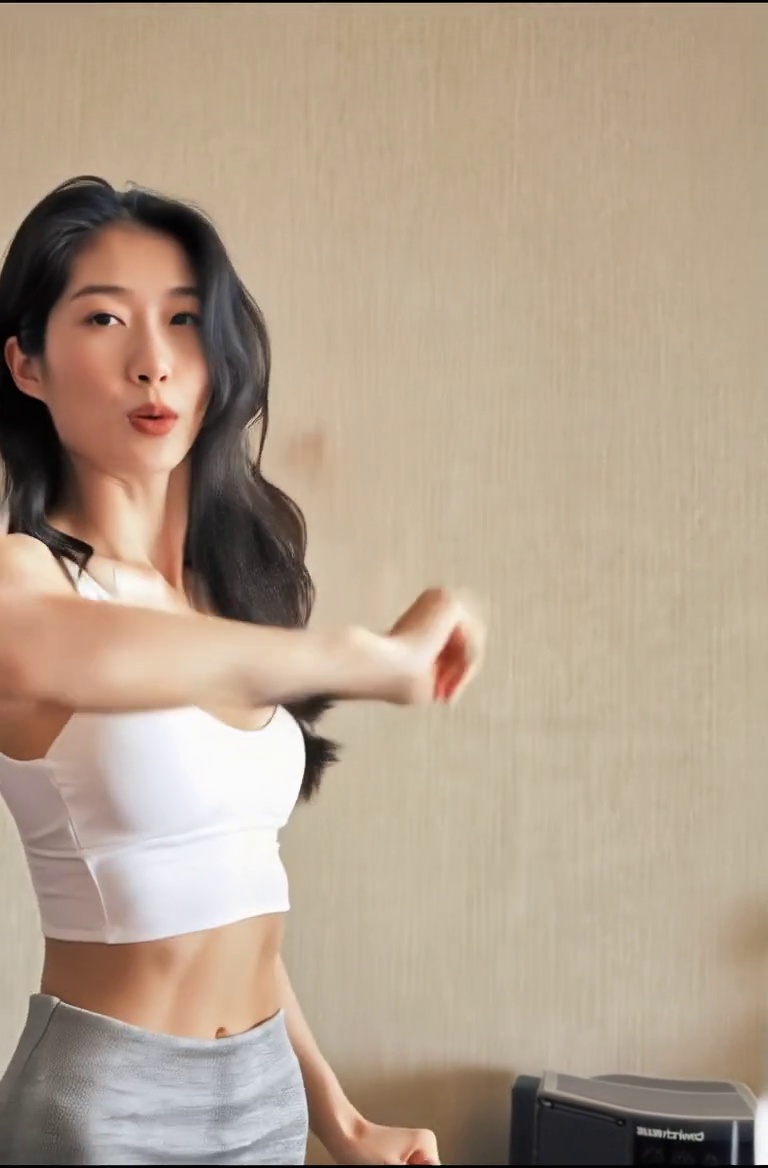} & \includegraphics[width=1.7cm]{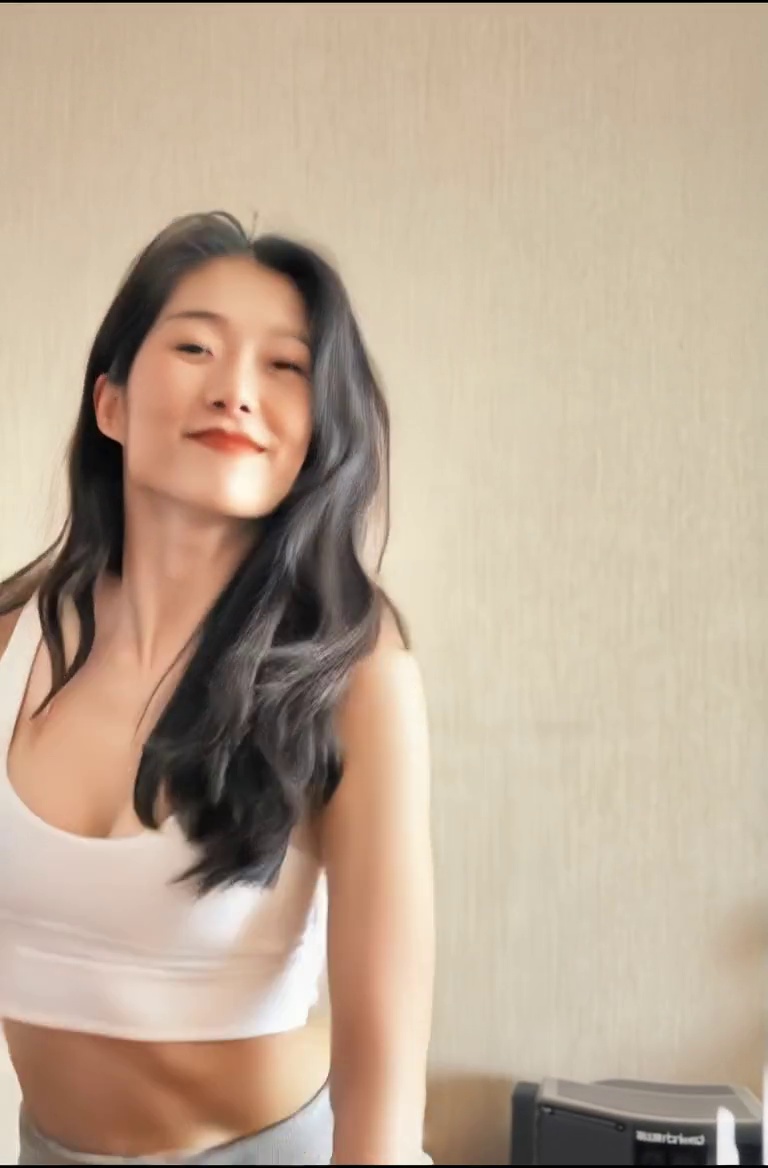} & \includegraphics[width=1.7cm]{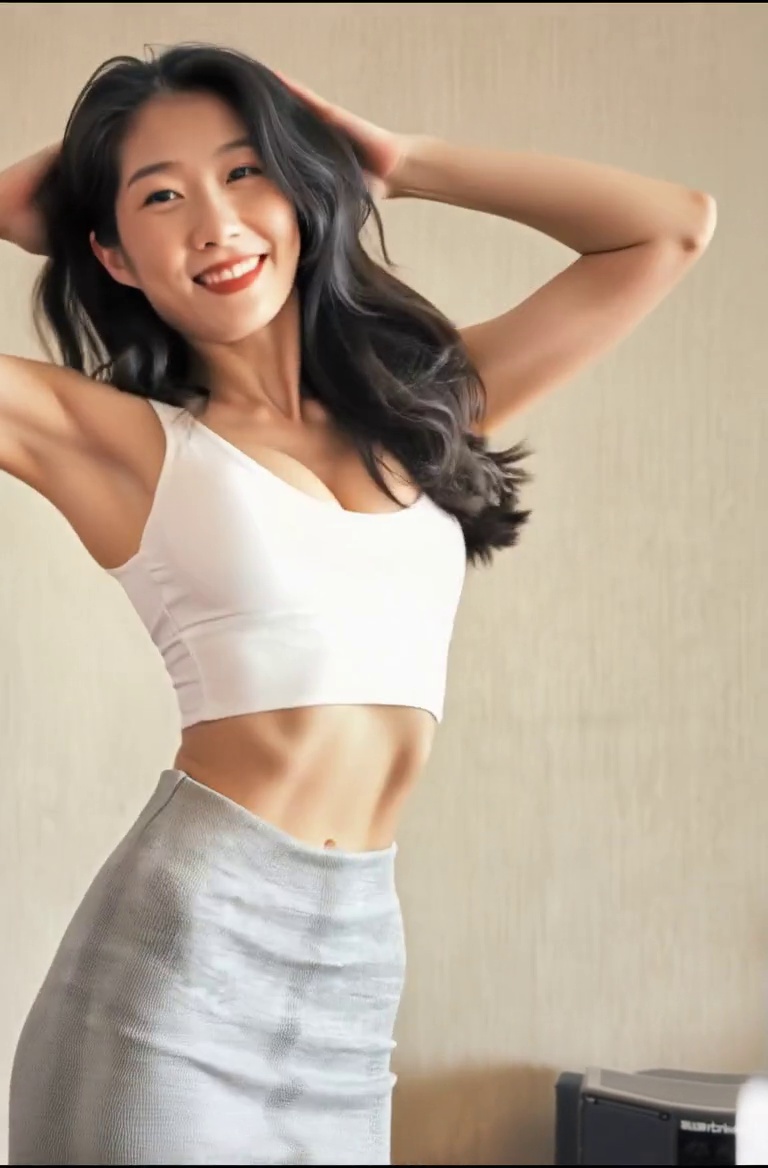} & \includegraphics[width=1.7cm]{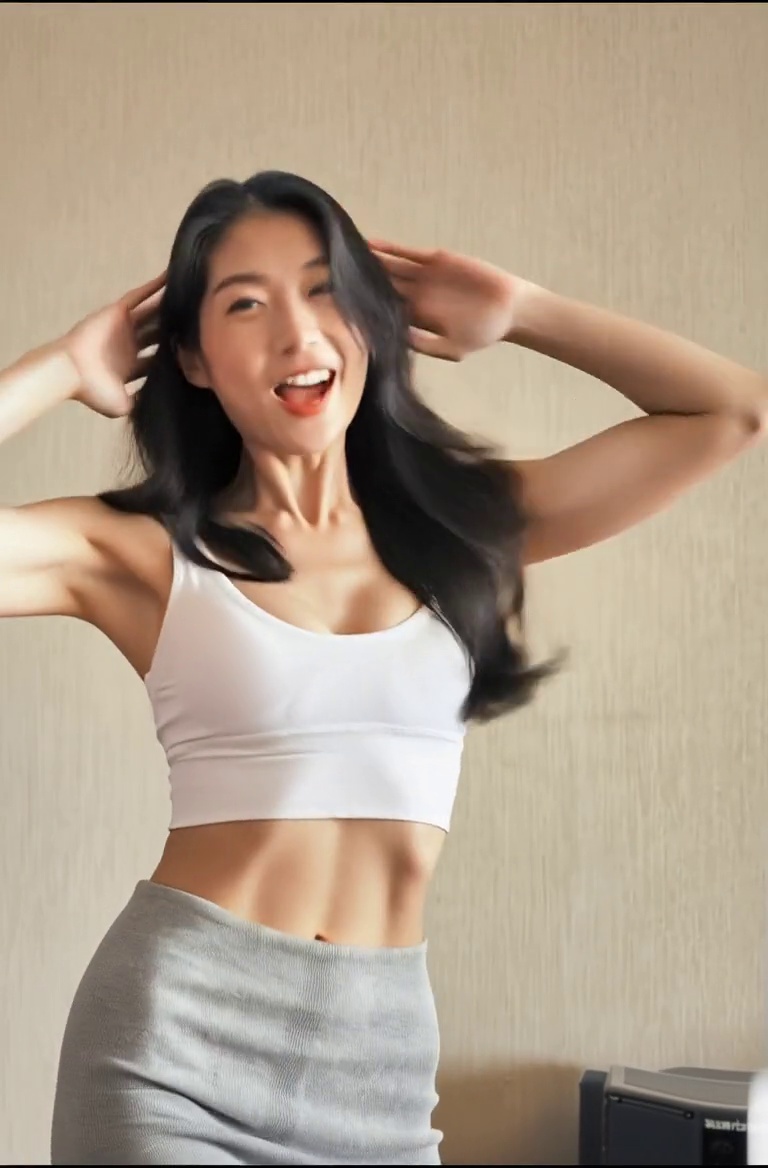} & \includegraphics[width=1.7cm]{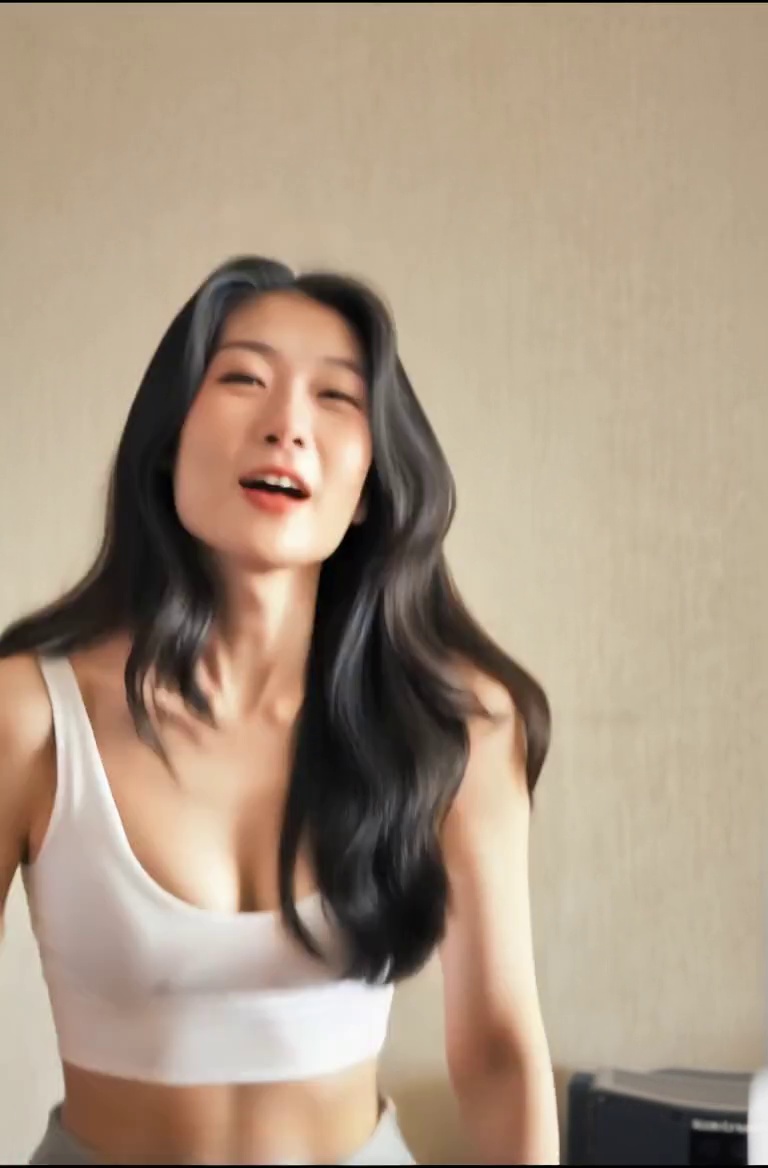} \\
    \end{tabular}
    \end{center}
    \caption{Qualitative comparison of cross-identity character image animation. Our method successfully animates the target character while preserving its original shape, whereas other approaches exhibit noticeable artifacts or unintended distortions.} 
    \label{fig:qual-comp}
    
\end{figure*}

We present a qualitative comparison of our animation method with state-of-the-art methods, namely Wan-Animate~\cite{cheng2025wananimateunifiedcharacteranimation}, Dreamina~\cite{dreamina}, and Kling-MotionControl~\cite{kling}. The comparative results are presented in Figure~\ref{fig:qual-comp}. Our method achieves automatic retargeting, animating the reference image according to its actual size, and effectively replicates highly complex facial expressions and hand movements. In contrast, competing methods struggle with expression fidelity, detailed hand articulation, and body shape misalignment.

\begin{figure*}[htbp]
    \centering
    \includegraphics[width=0.95\linewidth]{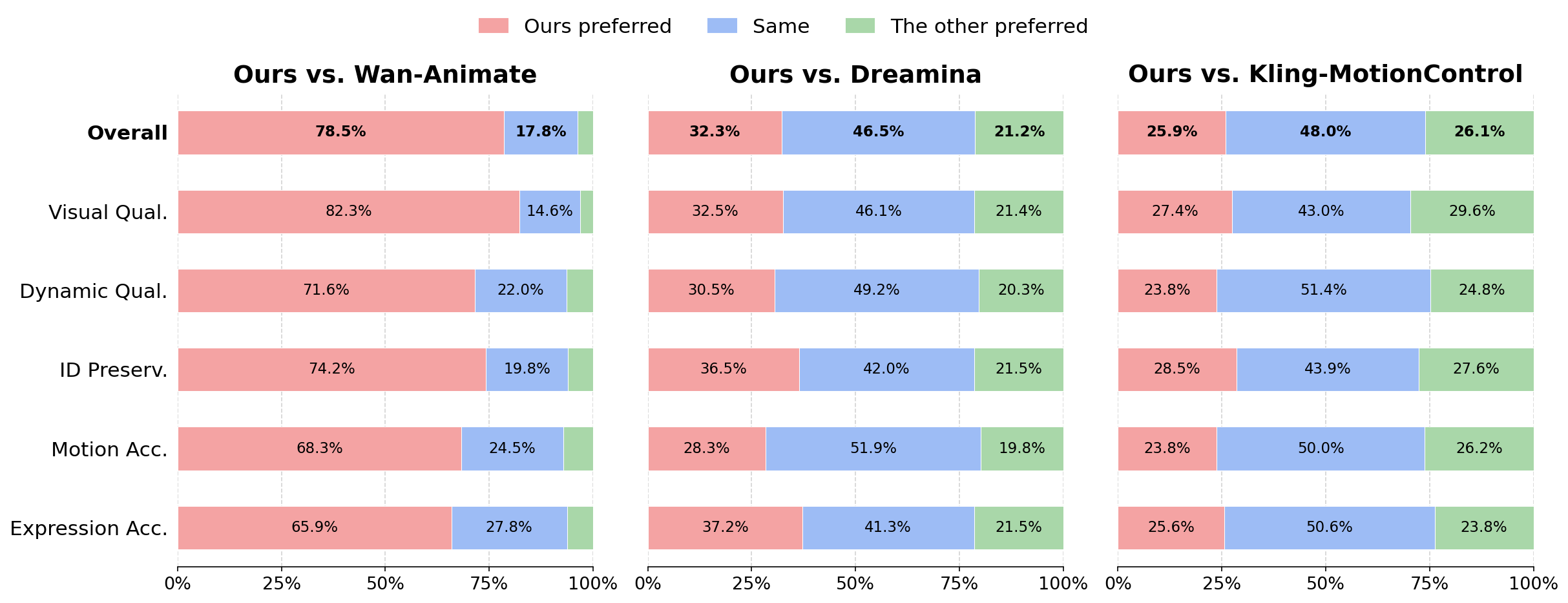}
    \caption{The visualization of our user study results demonstrates that \name, built entirely on an open-source base model, significantly outperforms existing open-source methods and surpasses or matches commercial closed-source platforms (Dreamina and Kling-MotionControl) that leverage larger proprietary foundation models.}
    \label{fig:user-study}
\end{figure*}

\subsection{User Study}

To evaluate the performance of \name, we conducted a blind user study comparing our framework against current open-source methods and prominent commercial solutions, including Wan-Animate~\cite{cheng2025wananimateunifiedcharacteranimation}, Dreamina~\cite{dreamina}, and Kling-MotionControl~\cite{kling}. Participants assessed the generated video sequences across five specific dimensions: visual quality, dynamic naturalness, identity preservation, motion accuracy, and facial expression accuracy, alongside an overall quality assessment.

The results, as illustrated in Figure~\ref{fig:user-study}, demonstrate that \name consistently outperforms Wan-Animate across all metrics, with over 70\% of pairwise comparisons favoring our method in overall quality. Notably, \name also surpasses Dreamina, a leading proprietary commercial platform, with participants preferring our results in the majority of comparisons. When compared against Kling-MotionControl, another prominent closed-source solution, \name achieves comparable performance, as most participants rated the generated videos to be of equivalent quality. It is worth emphasizing that both Dreamina and Kling-MotionControl are built upon larger, closed-source video generation foundation models, whereas \name is developed entirely on top of an open-source base model. The fact that an open-source system can match or even exceed the quality of these proprietary platforms underscores the effectiveness of our architectural design and training methodology, and demonstrates the viability of open-source approaches for high-fidelity character animation.

\subsection{Real-Time Inference}

\begin{figure*}[t]
    \centering
    \includegraphics[width=\linewidth]{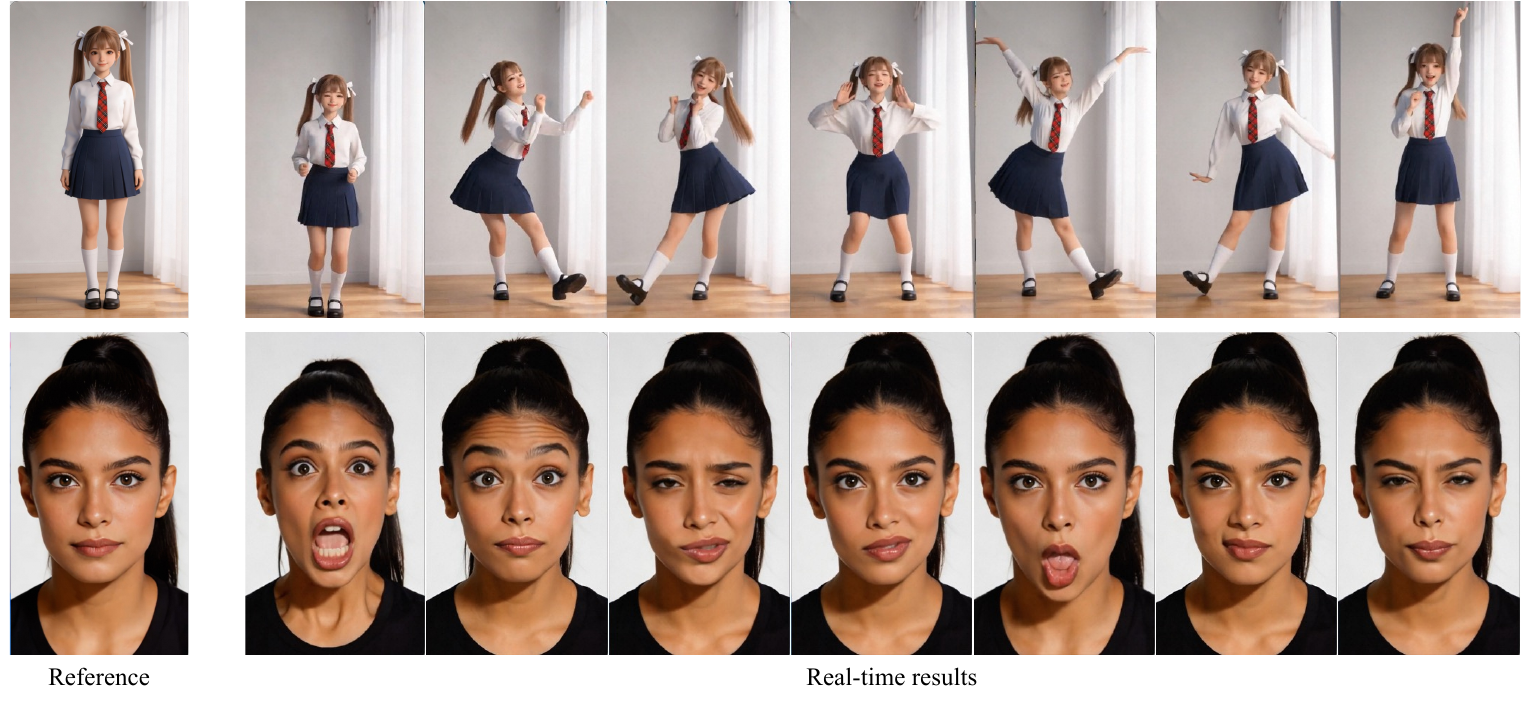}
    \caption{Streaming inference results of \name-Lite. The top row shows a dancing scenario and the bottom row shows an expression scenario, demonstrating stable, high-fidelity generation over extended sequences without error accumulation.}
    \label{fig:realtime-result}
\end{figure*}

To evaluate the real-time inference capability of Wan-Animate-2-Lite, we deploy the system on a 4-GPU cluster with a pipeline-parallel architecture. Specifically, we allocate the four GPUs as follows: one GPU handles VAE encoding of the input reference image and conditioning signals; two GPUs collaboratively execute the 3-step DiT denoising process using Sequence Parallelism, distributing the temporal dimension across devices to minimize per-GPU memory pressure; and the remaining GPU performs VAE decoding to produce the final pixel-space video frames. This pipelined design enables continuous, overlapping execution across stages, maximizing hardware utilization.

At a resolution of $400 \times 720$, the system achieves a throughput of 24 frames per second (fps), surpassing the real-time threshold and enabling smooth, interactive animation generation. As shown in Figure~\ref{fig:realtime-result}, \name-Lite produces visually consistent results across diverse scenarios, including dancing and expression. Notably, thanks to the error buffer mechanism introduced during teacher forcing pretraining, \name-Lite sustains autoregressive chunk-wise generation over extended sequences without observable error accumulation or quality degradation, making it suitable for applications requiring continuous video synthesis such as live streaming avatars and interactive virtual environments.

%% file: sec/6_conclusion.tex
\section{Conclusion}

This report presents \name, an end-to-end character animation framework that pushes the application boundaries of character animation in three complementary directions. First, we propose a redesigned Diffusion Transformer architecture that directly consumes driving videos without intermediate motion extractors, achieving superior motion fidelity and identity preservation through decoupled dual-branch attention, time-aligned positional encoding, and sparse reference attention. Second, we introduce text-driven viewpoint control that decouples the output camera perspective from the driving video, enabling flexible camera manipulation via natural language prompts—a capability rarely supported by prior character animation methods that rely on explicit motion representations. Third, we present \name-Lite, an efficient variant that reduces inference latency to real-time thresholds through a three-stage training paradigm (teacher forcing, error buffer, and self-forcing distillation), unlocking interactive deployment scenarios such as digital avatars and live-streaming hosts. Together, these contributions transform character animation from an offline, single-viewpoint generation task into a real-time, viewpoint-controllable, and high-fidelity interactive capability, broadening its applicability to digital humans, virtual production, and live entertainment.

%% file: main.bib
@String(CVPR   = {IEEE Conf. Comput. Vis. Pattern Recog.})

@String(ECCV   = {Eur. Conf. Comput. Vis.})

@String(NIPS   = {Adv. Neural Inform. Process. Syst.})

@String(ICLR   = {Int. Conf. Learn. Represent.})

@String(SIGGRAPH   = {SIGGRAPH})

@inproceedings{yin2024one,
  title={One-step diffusion with distribution matching distillation},
  author={Yin, Tianwei and Gharbi, Micha{\"e}l and Zhang, Richard and Shechtman, Eli and Durand, Fredo and Freeman, William T and Park, Taesung},
  booktitle=CVPR,
  pages={6613--6623},
  year={2024}
}

@article{yin2024improved,
  title={Improved distribution matching distillation for fast image synthesis},
  author={Yin, Tianwei and Gharbi, Micha{\"e}l and Park, Taesung and Zhang, Richard and Shechtman, Eli and Durand, Fredo and Freeman, William T},
  journal=NIPS,
  volume={37},
  pages={47455--47487},
  year={2024}
}

@article{wu2025qwen,
  title={Qwen-image technical report},
  author={Wu, Chenfei and Li, Jiahao and Zhou, Jingren and Lin, Junyang and Gao, Kaiyuan and Yan, Kun and Yin, Sheng-ming and Bai, Shuai and Xu, Xiao and Chen, Yilei and others},
  journal={arXiv preprint arXiv:2508.02324},
  year={2025}
}

@article{cai2025z,
  title={Z-image: An efficient image generation foundation model with single-stream diffusion transformer},
  author={Cai, Huanqia and Cao, Sihan and Du, Ruoyi and Gao, Peng and Hoi, Steven and Hou, Zhaohui and Huang, Shijie and Jiang, Dengyang and Jin, Xin and Li, Liangchen and others},
  journal={arXiv preprint arXiv:2511.22699},
  year={2025}
}

@misc{cheng2025wananimateunifiedcharacteranimation,
      title={Wan-Animate: Unified Character Animation and Replacement with Holistic Replication}, 
      author={Gang Cheng and Xin Gao and Li Hu and Siqi Hu and Mingyang Huang and Chaonan Ji and Ju Li and Dechao Meng and Jinwei Qi and Penchong Qiao and Zhen Shen and Yafei Song and Ke Sun and Linrui Tian and Feng Wang and Guangyuan Wang and Qi Wang and Zhongjian Wang and Jiayu Xiao and Sheng Xu and Bang Zhang and Peng Zhang and Xindi Zhang and Zhe Zhang and Jingren Zhou and Lian Zhuo},
      year={2025},
      eprint={2509.14055},
      archivePrefix={arXiv},
      primaryClass={cs.CV},
      url={https://arxiv.org/abs/2509.14055}, 
}

@article{hu2022lora,
  title={Lora: Low-rank adaptation of large language models.},
  author={Hu, Edward J and Shen, Yelong and Wallis, Phillip and Allen-Zhu, Zeyuan and Li, Yuanzhi and Wang, Shean and Wang, Liang and Chen, Weizhu and others},
  journal={Iclr},
  volume={1},
  number={2},
  pages={3},
  year={2022}
}

@inproceedings{peebles2023scalable,
  title={Scalable diffusion models with transformers},
  author={Peebles, William and Xie, Saining},
  booktitle={Proceedings of the IEEE/CVF international conference on computer vision},
  pages={4195--4205},
  year={2023}
}

@inproceedings{yang2023effective,
  title={Effective whole-body pose estimation with two-stages distillation},
  author={Yang, Zhendong and Zeng, Ailing and Yuan, Chun and Li, Yu},
  booktitle={Proceedings of the IEEE/CVF International Conference on Computer Vision},
  pages={4210--4220},
  year={2023}
}

@incollection{loper2023smpl,
  title={SMPL: A skinned multi-person linear model},
  author={Loper, Matthew and Mahmood, Naureen and Romero, Javier and Pons-Moll, Gerard and Black, Michael J},
  booktitle={Seminal Graphics Papers: Pushing the Boundaries, Volume 2},
  pages={851--866},
  year={2023}
}

@inproceedings{zhu2024champ,
      title={Champ: Controllable and Consistent Human Image Animation with 3D Parametric Guidance},
      author={Shenhao Zhu and Junming Leo Chen and Zuozhuo Dai and Yinghui Xu and Xun Cao and Yao Yao and Hao Zhu and Siyu Zhu},
      booktitle={European Conference on Computer Vision (ECCV)},
      year={2024}
}

@article{hu2023animateanyone,
  title={Animate Anyone: Consistent and Controllable Image-to-Video Synthesis for Character Animation},
  author={Li Hu and Xin Gao and Peng Zhang and Ke Sun and Bang Zhang and Liefeng Bo},
  journal={arXiv preprint arXiv:2311.17117},
  website={https://humanaigc.github.io/animate-anyone/},
  year={2023}
}

@article{mimicmotion2024,
  title={MimicMotion: High-Quality Human Motion Video Generation with Confidence-aware Pose Guidance},
  author={Yuang Zhang and Jiaxi Gu and Li-Wen Wang and Han Wang and Junqi Cheng and Yuefeng Zhu and Fangyuan Zou},
  journal={arXiv preprint arXiv:2406.19680},
  year={2024}
}

@inproceedings{
wang2022latent,
title={Latent Image Animator: Learning to Animate Images via Latent Space Navigation},
author={Yaohui Wang and Di Yang and Francois Bremond and Antitza Dantcheva},
booktitle={International Conference on Learning Representations},
year={2022}
}

@inproceedings{song2025x,
  title={X-UniMotion: Animating Human Images with Expressive, Unified and Identity-Agnostic Motion Latents},
  author={Song, Guoxian and Xu, Hongyi and Zhao, Xiaochen and Xie, You and Gu, Tianpei and Li, Zenan and Zhang, Chenxu and Luo, Linjie},
  booktitle={Proceedings of the SIGGRAPH Asia 2025 Conference Papers},
  pages={1--11},
  year={2025}
}

@article{wang2024lia,
  title={Lia: Latent image animator},
  author={Wang, Yaohui and Yang, Di and Bremond, Francois and Dantcheva, Antitza},
  journal={IEEE Transactions on Pattern Analysis and Machine Intelligence},
  volume={46},
  number={12},
  pages={10829--10844},
  year={2024},
  publisher={IEEE}
}

@article{luo2026dreamactor,
  title={DreamActor-M2: Universal Character Image Animation via Spatiotemporal In-Context Learning},
  author={Luo, Mingshuang and Liang, Shuang and Rong, Zhengkun and Luo, Yuxuan and Hu, Tianshu and Hou, Ruibing and Chang, Hong and Li, Yong and Zhang, Yuan and Gao, Mingyuan},
  journal={arXiv preprint arXiv:2601.21716},
  year={2026}
}

@misc{unrealengine,
  author       = {{Epic Games}},
  title        = {Unreal Engine},
  howpublished = {\url{https://www.unrealengine.com/}},
}

@misc{dreamina,
  author       = {{ByteDance}},
  title        = {Dreamina},
  howpublished = {\url{https://dreaminai.org/}},
}

@misc{kling,
  author       = {{KuaiShou}},
  title        = {Kling},
  howpublished = {\url{https://kling.ai/}},
}

@article{huang2026self,
  title={Self forcing: Bridging the train-test gap in autoregressive video diffusion},
  author={Huang, Xun and Li, Zhengqi and He, Guande and Zhou, Mingyuan and Shechtman, Eli},
  journal={Advances in Neural Information Processing Systems},
  volume={38},
  pages={167283--167308},
  year={2026}
}

@article{li2025stable,
  title={Stable video infinity: Infinite-length video generation with error recycling},
  author={Li, Wuyang and Pan, Wentao and Luan, Po-Chien and Gao, Yang and Alahi, Alexandre},
  journal={arXiv preprint arXiv:2510.09212},
  year={2025}
}

@article{longlive_2.0,
  title={LongLive2.0: An NVFP4 Parallel Infrastructure for Long Video Generation},
  author={Chen, Yukang and Wang, Luozhou and Huang, Wei and Yang, Shuai and Zhang, Bohan and Xiao, Yicheng and Chu, Ruihang and Mao, Weian and Hu, Qixin and Liu, Shaoteng and Zhao, Yuyang and Mao, Huizi and Chen, Ying-Cong and Xie, Enze and Qi, Xiaojuan and Han, Song},
  journal={arXiv preprint arXiv: 2605.18739},
  year={2026}
}

@inproceedings{pavlakos2019expressive,
  title={Expressive body capture: 3d hands, face, and body from a single image},
  author={Pavlakos, Georgios and Choutas, Vasileios and Ghorbani, Nima and Bolkart, Timo and Osman, Ahmed AA and Tzionas, Dimitrios and Black, Michael J},
  booktitle={Proceedings of the IEEE/CVF conference on computer vision and pattern recognition},
  pages={10975--10985},
  year={2019}
}

@inproceedings{cao2017realtime,
  title={Realtime multi-person 2d pose estimation using part affinity fields},
  author={Cao, Zhe and Simon, Tomas and Wei, Shih-En and Sheikh, Yaser},
  booktitle={Proceedings of the IEEE conference on computer vision and pattern recognition},
  pages={7291--7299},
  year={2017}
}

@article{xu2023vitpose++,
  title={Vitpose++: Vision transformer for generic body pose estimation},
  author={Xu, Yufei and Zhang, Jing and Zhang, Qiming and Tao, Dacheng},
  journal={IEEE Transactions on Pattern Analysis and Machine Intelligence},
  volume={46},
  number={2},
  pages={1212--1230},
  year={2023},
  publisher={IEEE}
}

@article{xu2022vitpose,
  title={Vitpose: Simple vision transformer baselines for human pose estimation},
  author={Xu, Yufei and Zhang, Jing and Zhang, Qiming and Tao, Dacheng},
  journal={Advances in neural information processing systems},
  volume={35},
  pages={38571--38584},
  year={2022}
}

@article{ma2023follow,
  title={Follow Your Pose: Pose-Guided Text-to-Video Generation using Pose-Free Videos},
  author={Ma, Yue and He, Yingqing and Cun, Xiaodong and Wang, Xintao and Shan, Ying and Li, Xiu and Chen, Qifeng},
  journal={arXiv preprint arXiv:2304.01186},
  year={2023}
}

@misc{karras2023dreamposefashionimagetovideosynthesis,
      title={DreamPose: Fashion Image-to-Video Synthesis via Stable Diffusion}, 
      author={Johanna Karras and Aleksander Holynski and Ting-Chun Wang and Ira Kemelmacher-Shlizerman},
      year={2023},
      eprint={2304.06025},
      archivePrefix={arXiv},
      primaryClass={cs.CV},
      url={https://arxiv.org/abs/2304.06025}, 
}

@article{wang2023disco,
  title={Disco: Disentangled control for realistic human dance generation},
  author={Wang, Tan and Li, Linjie and Lin, Kevin and Zhai, Yuanhao and Lin, Chung-Ching and Yang, Zhengyuan and Zhang, Hanwang and Liu, Zicheng and Wang, Lijuan},
  journal={arXiv preprint arXiv:2307.00040},
  year={2023}
}

@article{chang2023magicdance,
  title={MagicDance: Realistic Human Dance Video Generation with Motions \& Facial Expressions Transfer},
  author={Chang, Di and Shi, Yichun and Gao, Quankai and Fu, Jessica and Xu, Hongyi and Song, Guoxian and Yan, Qing and Yang, Xiao and Soleymani, Mohammad},
  journal={arXiv preprint arXiv:2311.12052},
  year={2023}
}

@inproceedings{xu2023magicanimate,
    author    = {Xu, Zhongcong and Zhang, Jianfeng and Liew, Jun Hao and Yan, Hanshu and Liu, Jia-Wei and Zhang, Chenxu and Feng, Jiashi and Shou, Mike Zheng},
    title     = {MagicAnimate: Temporally Consistent Human Image Animation using Diffusion Model},
    booktitle = {arXiv},
    year      = {2023}
}

@misc{kim2024tcananimatinghumanimages,
      title={TCAN: Animating Human Images with Temporally Consistent Pose Guidance using Diffusion Models}, 
      author={Jeongho Kim and Min-Jung Kim and Junsoo Lee and Jaegul Choo},
      year={2024},
      eprint={2407.09012},
      archivePrefix={arXiv},
      primaryClass={cs.CV},
      url={https://arxiv.org/abs/2407.09012}, 
}

@article{wang2024unianimate,
      title={UniAnimate: Taming Unified Video Diffusion Models for Consistent Human Image Animation},
      author={Wang, Xiang and Zhang, Shiwei and Gao, Changxin and Wang, Jiayu and Zhou, Xiaoqiang and Zhang, Yingya and Yan, Luxin and Sang, Nong},
      journal={arXiv preprint arXiv:2406.01188},
      year={2024}
}

@misc{yoon2025tpctesttimeprocrustescalibration,
      title={TPC: Test-time Procrustes Calibration for Diffusion-based Human Image Animation}, 
      author={Sunjae Yoon and Gwanhyeong Koo and Younghwan Lee and Chang D. Yoo},
      year={2025},
      eprint={2410.24037},
      archivePrefix={arXiv},
      primaryClass={cs.CV},
      url={https://arxiv.org/abs/2410.24037}, 
}

@misc{tan2024animatexuniversalcharacterimage,
      title={Animate-X: Universal Character Image Animation with Enhanced Motion Representation}, 
      author={Shuai Tan and Biao Gong and Xiang Wang and Shiwei Zhang and Dandan Zheng and Ruobing Zheng and Kecheng Zheng and Jingdong Chen and Ming Yang},
      year={2024},
      eprint={2410.10306},
      archivePrefix={arXiv},
      primaryClass={cs.CV},
      url={https://arxiv.org/abs/2410.10306}, 
}

@InProceedings{Siarohin_2019_NeurIPS,
  author={Siarohin, Aliaksandr and Lathuilière, Stéphane and Tulyakov, Sergey and Ricci, Elisa and Sebe, Nicu},
  title={First Order Motion Model for Image Animation},
  booktitle = {Conference on Neural Information Processing Systems (NeurIPS)},
  month = {December},
  year = {2019}
}

@article{ding2025mtvcrafter,
  title={Mtvcrafter: 4d motion tokenization for open-world human image animation},
  author={Ding, Yanbo and Hu, Xirui and Guo, Zhizhi and Zhang, Chi and Wang, Yali},
  journal={arXiv preprint arXiv:2505.10238},
  volume={3},
  year={2025}
}

@misc{yan2026scail2,
      title={SCAIL-2: Unifying Controlled Character Animation with End-to-end In-Context Conditioning}, 
      author={Wenhao Yan and Fengjia Guo and Zhuoyi Yang and Jie Tang},
      year={2026},
      eprint={2606.10804},
      archivePrefix={arXiv},
      primaryClass={cs.CV},
      url={https://arxiv.org/abs/2606.10804}, 
}
